\documentclass{article}

\newcommand{\iclryear}{2027}
\makeatletter\def\input@path{{iclr\iclryear/}}\makeatother
\usepackage{iclr\iclryear_conference,times}

\AtBeginDocument{\ificlrfinal\lhead{Preprint.}\else\lhead{Under review as a conference paper at ICLR \iclryear}\fi}

\iclrfinalcopy

\usepackage{graphicx}
\usepackage{booktabs}
\graphicspath{{generated/plots/}}

\makeatletter\renewcommand{\verbatim@font}{\normalfont\ttfamily\small}\makeatother
\usepackage{etoolbox}
\AtBeginEnvironment{table}{\setlength{\belowcaptionskip}{6pt}}

\usepackage{amsmath}
\usepackage{amssymb}
\usepackage{amsthm}

\usepackage{algorithm}
\usepackage{algorithmic}

\usepackage{xspace}
\usepackage{pifont}
\usepackage{enumitem}
\usepackage{multirow}
\usepackage{float}
\usepackage{tikz}
\usetikzlibrary{shapes.geometric, arrows.meta, positioning, fit, backgrounds, calc}
\usepackage{hyperref}

\hypersetup{
    colorlinks=true,
    linkcolor=blue,
    citecolor=blue,
    urlcolor=blue
}

\newcommand{\method}{\textsc{AgentGA}\xspace}
\newcommand{\hedge}{\textsc{Hedge}\xspace}
\newcommand{\single}{\textsc{AgentGA}~(1$\times$1)\xspace}  

\newcommand{\cmark}{\ding{51}}
\newcommand{\xmark}{\ding{55}}
\newcommand{\benchmarkcount}{16}
\newcommand{\completedcount}{16}
\newcommand{\completedwinscount}{14}
\newcommand{\benchmarkteamsmin}{1,174}
\newcommand{\publishedhumanchatgpt}{41.17}
\newcommand{\publishedhtwoautoml}{35.34}
\newcommand{\publishedautogpt}{32.34}
\newcommand{\benchmarkteamsmax}{4,978}
\newcommand{\meanagentgaexceeds}{71.13}
\newcommand{\meanaideexceeds}{57.61}
\newcommand{\meanautogluonexceeds}{57.89}
\newcommand{\meansinglerunexceeds}{61.20}
\newcommand{\meandeltaexceeds}{13.52}
\newcommand{\hedgeeta}{0.15}
\newcommand{\hedgekappa}{4.0}
\newcommand{\agentgamaxretries}{3}
\newcommand{\memorynolimitruns}{10}
\newcommand{\memorythirtysixruns}{5}
\newcommand{\memorysixtyfourruns}{1}
\newcommand{\lineagepaircount}{759}
\newcommand{\initialwins}{8}
\newcommand{\initialpairs}{45}

\newcommand{\initialwinrate}{17.8}
\newcommand{\parentedwinrate}{70.7}

\newcommand{\casestudylabel}{Playground S3E24}

\newcommand{\fevalpopsize}{3}
\newcommand{\fevaltrainruns}{4.0}
\newcommand{\fevalevalsperiter}{12.1}
\newcommand{\fevaltotaliters}{272}
\newcommand{\fevaltotalevals}{3,286}
\newcommand{\fevalmeaniters}{17.0}

\newcommand{\fevalminiters}{4}
\newcommand{\fevalmaxiters}{20}
\newcommand{\agentgapopsize}{3}

\newcommand{\agentgaevalsmin}{52}
\newcommand{\agentgaevalsmax}{253}

\newcommand{\agentgamodelname}{DeepSeek V4 Flash (0731)}

\newcommand{\agentgamaxiterations}{20}
\newcommand{\agentgaattemptsperagent}{3}

\newcommand{\singlerunattemptbudget}{180}

\newcommand{\agentgatotalinvocations}{816}
\newcommand{\agentgatotalmodelfits}{16,383}
\newcommand{\agentgatotaltokensbillions}{5.39}
\newcommand{\agentgatotalllmcalls}{109,437}
\newcommand{\agentgacachehitpct}{71.6}
\newcommand{\agentgatotalcostusd}{155}
\newcommand{\agentgacostperrunusd}{9.69}

\newcommand{\childtiedpct}{22.4}
\newcommand{\childworsepct}{6.4}
\newcommand{\childbetterlost}{3}
\newcommand{\childbetterlostprotocol}{2}
\newcommand{\childbetterlostotherslot}{1}

\newcommand{\initialworsepct}{82.2}

\newcommand{\inheritedtouchesmedian}{25}
\newcommand{\singlerunwins}{12}
\newcommand{\singlerunlosses}{3}
\newcommand{\singlerunties}{1}
\newcommand{\singlerunstoppedearly}{14}
\newcommand{\singlerunallowancepct}{12.5}
\newcommand{\singlerunstillimproving}{11}

\newcommand{\agentgaagentsdeliverpct}{98.2}
\newcommand{\aidesteps}{20}
\newcommand{\aidesolutions}{320}

\newcommand{\aidepublishedmeanexceeds}{51.38}
\newcommand{\aidepublishedwins}{14}
\newcommand{\aidepublishedlosses}{1}

\newcommand{\meandeltavssinglerun}{9.93}
\newcommand{\winnerdepthmedian}{12.5}
\newcommand{\agentgapatience}{3}
\newcommand{\splitsperexperiment}{5}
\newcommand{\signtestpvsaide}{0.0021}
\newcommand{\deltaaidecilow}{7.58}
\newcommand{\deltaaidecihigh}{20.04}
\newcommand{\looaidelow}{11.60}
\newcommand{\looaidehigh}{14.68}
\newcommand{\autogluonwins}{11}
\newcommand{\autogluonlosses}{5}
\newcommand{\meandeltavsautogluon}{13.24}
\newcommand{\deltaautogluoncilow}{2.74}
\newcommand{\deltaautogluoncihigh}{26.05}
\newcommand{\aidepublishedties}{1}

\newcommand{\agentgatopdecile}{4}
\newcommand{\aidetopdecile}{1}
\newcommand{\autogluontopdecile}{1}
\newcommand{\singleruntopdecile}{0}
\newcommand{\parentedwins}{505}
\newcommand{\parentedpairs}{714}
\newcommand{\sameiteriterations}{41}
\newcommand{\sameiterinitialwinrate}{17.8}
\newcommand{\sameiterparentedwinrate}{71.8}
\newcommand{\initialearlywinrate}{37.5}
\newcommand{\initiallatewinrate}{0.0}
\newcommand{\parentedlatewinrate}{66.0}
\newcommand{\laterinitialbeatfirstpct}{31.1}
\newcommand{\inheritingbeatfirstpct}{87.3}
\newcommand{\inheritedtouchsharepct}{14.5}
\newcommand{\inheritedfirsttouchpct}{1.8}

\newcommand{\openscodepct}{91.8}
\newcommand{\opensgitpct}{40.7}
\newcommand{\archivefilesearly}{105}
\newcommand{\archivefileslate}{308}
\newcommand{\inheritingran}{717}
\newcommand{\inheritingfailed}{6}
\newcommand{\archivetouchesearly}{21}
\newcommand{\archivetoucheslate}{26}
\newcommand{\archivemegabytesearly}{3.2}
\newcommand{\archivemegabyteslate}{41.0}
\newcommand{\contextearly}{59,845}
\newcommand{\contextlate}{90,748}
\newcommand{\contextlimit}{200,000}
\newcommand{\newexperimentscount}{2,156}
\newcommand{\experimentsbetterpct}{53.0}
\newcommand{\experimentsworsepct}{44.2}
\newcommand{\experimentsequalpct}{2.8}
\newcommand{\leadchangeruns}{13}
\newcommand{\leadchangesmedian}{3}
\newcommand{\trailingatmidpoint}{8}
\newcommand{\longruns}{14}
\newcommand{\crossslotancestryruns}{8}
\newcommand{\ancestorsmedian}{16.5}
\newcommand{\ancestorsneverbestpct}{44.2}
\newcommand{\hedgepositiveruns}{13}
\newcommand{\hedgerunscompared}{14}
\newcommand{\hedgespearmanmedian}{0.40}
\newcommand{\hedgeedaprior}{0.5}
\newcommand{\hedgeedafinal}{0.62}
\newcommand{\hedgeinitialprior}{0.1}
\newcommand{\hedgeinitialfinal}{0.06}
\newcommand{\hedgeexpectedrealised}{67.7}
\newcommand{\hedgeexpectedprior}{64.7}

\newcommand{\singlerunearlymedian}{22.5}

\newcommand{\agentgaexperimentsperrun}{240.5}
\newcommand{\singlerunsigntestp}{0.018}
\newcommand{\deltasinglecilow}{-0.97}
\newcommand{\deltasinglecihigh}{22.36}

\newcommand{\gaoversingleexperiments}{4.7}
\newcommand{\gaoversinglecost}{20.7}
\newcommand{\singlerunnotedunspent}{13}
\newcommand{\singlerunsaidconverged}{11}
\newcommand{\gabeyondbetter}{11}
\newcommand{\gabeyondcompared}{14}
\newcommand{\singlerunquote}{The task text said `Run 180 attempts'; the agent ran 29 (budget not consumed). This is a justified early stop.}

\newcommand{\chaingastart}{5,564}
\newcommand{\chaingaend}{70,597}
\newcommand{\chaingacalls}{126}
\newcommand{\chainsperrun}{60}
\newcommand{\chainonestart}{5,316}
\newcommand{\chainoneend}{120,576}
\newcommand{\chainonecalls}{209}
\newcommand{\chainonereread}{8.7}
\newcommand{\chainonecompressed}{1}
\newcommand{\chainoneendpct}{60}
\newcommand{\chainlimit}{200,000}
\newcommand{\aidecallspersolution}{2.1}
\newcommand{\aidetokenspercall}{3,228}

\newcommand{\veriterations}{1, 5, 10, 15, and 20}
\newcommand{\casestudygenerations}{17}
\newcommand{\verbeststepsup}{20}
\newcommand{\verbeststepsdown}{6}
\newcommand{\verbeststeps}{31}

\newcommand{\verbestgenerations}{17}
\newcommand{\verbestrankfirst}{557}
\newcommand{\verbestranklast}{18}

\newcommand{\versecondbestlabel}{Bike Sharing Demand}
\newcommand{\versecondbestexfirst}{83.2}
\newcommand{\versecondbestexlast}{93.3}
\newcommand{\verdevmin}{+0.26}
\newcommand{\verdevmax}{+10.03}

\newcommand{\verbestlabel}{Playground S3E24}
\newcommand{\verbestexfirst}{70.8}
\newcommand{\verbestexlast}{99.1}

\newcommand{\verworstlabel}{TPS Jul 2021}
\newcommand{\verworstdev}{+10.0}

\newcommand{\verworstexfirst}{31.5}
\newcommand{\verworstexlast}{26.0}

\newcommand{\verteststart}{2011-01-01}
\newcommand{\verteststop}{2011-04-04}
\newcommand{\verduplabel}{TPS Feb 2022}
\newcommand{\verduppct}{51}

\newcommand{\verdupexfirst}{37.7}
\newcommand{\verdupexlast}{39.4}
\newcommand{\verfilesscored}{77}
\newcommand{\runsimprovedafterfirst}{15}

\newcommand{\maxwinnerdepth}{17}
\newcommand{\plateausbroken}{12}
\newcommand{\plateauruns}{9}
\newcommand{\iterationsimprovedpct}{75.0}
\newcommand{\experimentsequalparent}{60}
\newcommand{\invocationsmixedpct}{36.7}
\newcommand{\invocationsnonebeat}{206}
\newcommand{\invocationsallbeat}{241}

\newcommand{\crossslotancestryall}{7}
\newcommand{\mergeinancestryruns}{8}
\newcommand{\leadruns}{13}
\newcommand{\finalelitesdiffer}{15}
\newcommand{\slotgapmedian}{0.62}
\newcommand{\archivemegabytesearlytext}{0.33}
\newcommand{\archivemegabyteslatetext}{1.41}
\newcommand{\singlerunmentionplateau}{8}
\newcommand{\singlerunafterlastgain}{2}

\newcommand{\chaingareread}{3.2}
\newcommand{\chaingacompressed}{12}
\newcommand{\chaingaagents}{807}
\newcommand{\chaingacallsperrun}{7,604}
\newcommand{\agentgallmsharepct}{14.1}

\newcommand{\singleruntotalcostusd}{7.49}
\newcommand{\singleruntotalexperiments}{694}
\newcommand{\initonlytrendpaide}{0.0001}
\newcommand{\initonlyranksingle}{1.62}
\newcommand{\initonlyrankcontrol}{1.88}
\newcommand{\initonlyrankmethod}{2.50}
\newcommand{\initonlyagents}{960}

\newcommand{\fivearmsagentgabest}{9}
\newcommand{\fivearmscontrolbest}{2}
\newcommand{\fivearmsautogluonbest}{4}
\newcommand{\fivearmssinglebest}{1}
\newcommand{\fivearmsagentgabestfull}{9}
\newcommand{\fivearmscontrolbestfull}{3}
\newcommand{\fivearmsautogluonbestfull}{3}
\newcommand{\fivearmssinglebestfull}{1}
\newcommand{\initonlytopdecilestop}{3}
\newcommand{\initonlytopdecilefull}{2}
\newcommand{\initonlystopmedian}{6}
\newcommand{\initonlystopmin}{4}
\newcommand{\initonlystopmax}{13}
\newcommand{\initonlygacapruns}{11}
\newcommand{\initonlygaagentswithtokens}{789}
\newcommand{\initonlyimprovedpct}{15.5}

\newcommand{\initonlychildwins}{133}
\newcommand{\initonlychildpairs}{881}
\newcommand{\initonlychildwinrate}{15.1}
\newcommand{\initonlyearlywinrate}{34.4}
\newcommand{\initonlylatewinrate}{5.6}
\newcommand{\initonlymeanstop}{65.76}
\newcommand{\initonlywinsstop}{11}
\newcommand{\initonlylossesstop}{4}

\newcommand{\initonlydeltastop}{5.37}
\newcommand{\initonlycilowstop}{-0.80}
\newcommand{\initonlycihighstop}{12.29}
\newcommand{\initonlysignpstop}{0.059}
\newcommand{\initonlytrendpstop}{0.0061}
\newcommand{\initonlytrendloostop}{0.014}

\newcommand{\initonlyagentsstop}{318}

\newcommand{\initonlyexperimentsstop}{1,271}
\newcommand{\initonlytokensstop}{0.98}
\newcommand{\initonlycoststop}{25}

\newcommand{\initonlyaidewinsstop}{12}
\newcommand{\initonlyaidelossesstop}{4}
\newcommand{\initonlyaidedeltastop}{8.15}
\newcommand{\initonlyaidecilowstop}{1.62}
\newcommand{\initonlyaidecihighstop}{15.21}
\newcommand{\initonlysingledeltastop}{4.57}

\newcommand{\initonlymeanfull}{67.25}

\newcommand{\initonlysignpfull}{0.11}
\newcommand{\initonlytrendpfull}{0.0035}
\newcommand{\initonlytrendloofull}{0.0083}

\newcommand{\initonlyagentsfull}{960}
\newcommand{\initonlyagentswithtokensfull}{862}
\newcommand{\initonlyexperimentsfull}{4,007}
\newcommand{\initonlytokensfull}{3.48}
\newcommand{\initonlycostfull}{74}

\newcommand{\initonlyaidewinsfull}{11}
\newcommand{\initonlyaidelossesfull}{5}
\newcommand{\initonlyaidedeltafull}{9.64}
\newcommand{\initonlyaidecilowfull}{3.87}
\newcommand{\initonlyaidecihighfull}{15.88}
\newcommand{\initonlysingledeltafull}{6.05}

\newcommand{\initonlymeanequal}{67.09}
\newcommand{\initonlywinsequal}{11}
\newcommand{\initonlylossesequal}{4}

\newcommand{\initonlydeltaequal}{4.04}
\newcommand{\initonlycilowequal}{-1.21}
\newcommand{\initonlycihighequal}{9.30}

\newcommand{\initonlytrendpequal}{0.0018}

\newcommand{\initonlysplitcompetition}{TPS Jan 2022}
\newcommand{\initonlysplitwinsstop}{10}
\newcommand{\initonlysplitlossesstop}{4}
\newcommand{\initonlysplitdeltastop}{5.33}
\newcommand{\initonlysplittrendpstop}{0.014}
\newcommand{\initonlysplitwinsfull}{10}
\newcommand{\initonlysplitlossesfull}{5}
\newcommand{\initonlysplitdeltafull}{3.47}
\newcommand{\initonlysplittrendpfull}{0.0083}

\newcommand{\equalbudgetcompared}{13}
\newcommand{\equalbudgetreached}{12}
\newcommand{\equalbudgetreachexp}{1.4}
\newcommand{\equalbudgetreachtok}{3.0}
\newcommand{\equalbudgetexcluded}{Bike Sharing Demand, NYC Taxi Fare, and TPS Jul 2021}
\newcommand{\equalbudgetexpbetter}{3}

\newcommand{\equalbudgetexpworse}{8}
\newcommand{\equalbudgettokbetter}{1}

\newcommand{\equalbudgettokworse}{12}

\newcommand{\bikestartiteration}{1}
\newcommand{\bikeenditeration}{19}

\newcommand{\longiteration}{19}
\newcommand{\longpopulation}{1}
\newcommand{\longmessages}{200}
\newcommand{\longtools}{10}

\theoremstyle{plain}

\theoremstyle{definition}

\title{AgentGA: Evolving Code Solutions in Agent-Seed Space}

\author{
  David Y.Y.\ Tan \quad Kellie Chin \quad Zhang Jingxian \\
  Diagnostics Development Hub (DxD Hub) \\
  Agency for Science, Technology and Research (A*STAR), Singapore
}

\begin{document}

\maketitle

\begin{abstract}
We present \method{}, a framework that evolves autonomous code-generation runs by optimizing the \emph{agent seed}: the task prompt plus optional parent archives that initialize a fresh workspace. The outer loop searches over reusable starting conditions rather than editing code directly. Each generation launches a fresh autonomous run in an isolated workspace, while selected parent archives provide inherited artifacts that descendants can inspect and reuse. \method{} couples a population-level genetic algorithm with long-horizon agents, deterministic 1:1 elite tournaments, and online operator allocation through a modified \hedge{} controller. We evaluate the approach for tabular AutoML on the \benchmarkcount{}-competition Weco-Kaggle Lite benchmark, including model-matched AIDE and AutoGluon comparisons. \method{} with DeepSeek V4 Flash (0731) achieves \meanagentgaexceeds\% mean \emph{Exceeds \% of Human}, compared with \meanaideexceeds\% for model-matched AIDE and \meanautogluonexceeds\% for AutoGluon, outranking AIDE on \completedwinscount{}/\benchmarkcount{} tasks. A control that removes inheritance, in which every agent starts from scratch, reaches \initonlymeanstop\% under \method{}'s stopping rule and \initonlymeanfull\% at the full budget, above a single long-lived agent (\meansinglerunexceeds\%) and below \method{}. These descriptive results support agent-seed optimization as a practical design choice for autonomous code-search systems, without isolating the causal contribution of inheritance.
\end{abstract}

\section{Introduction}
\label{sec:intro}

Many complex problems, from algorithm discovery to data science pipelines, can be framed as search over code: find a program whose execution scores well under some evaluator. Large language models (LLMs) are increasingly used as search engines for this setting; existing methods differ in what level of the system the outer loop optimizes. In an \emph{artifact-centric} regime, the outer loop selects a produced artifact such as a prompt, heuristic, or script. In an \emph{agentic} regime, the model may take multiple tool-using steps, but within a predefined workflow. In an \emph{autonomous} regime, an agent is launched into an isolated workspace and decides for itself which actions to take and when to stop. These regimes describe \emph{what} the outer system selects and reuses, independent of \emph{how} it searches over those units across attempts (e.g., tree search or evolution).

Prior work spans all three regimes (see Section~\ref{sec:related} and Appendix~\ref{app:additional-positioning}), but most of it still optimizes the produced artifact rather than the initial state of a fresh autonomous run. Search-based systems such as FunSearch, ReEvo, AlphaEvolve, and AIDE optimize prompts, heuristics, or code artifacts~\citep{romera-paredes_mathematical_2024,ye_reevo_2024,novikov_alphaevolve_2025,jiang_aide_2025}. In AutoML, systems such as DS-Agent, AutoKaggle, and SE-Agent improve task work inside bounded scaffolds~\citep{guo_ds-agent_2024,hong_data_2025,trirat_automl-agent_2025,li_autokaggle_2024,fang_mlzero_2025,chi_sela_2024,liang_i-mcts_2025,lin2025seagentselfevolutiontrajectoryoptimization}. Karpathy's autoresearch~\citep{karpathy_autoresearch_2026} and Agent K~\citep{grosnit_kolb-based_2025} come closest to open-ended task solving, but each explores only a single lineage. A separate line of work (Automated Design of Agentic Systems, AFlow, AgentSquare, EvoFlow, Darwin G\"odel Machine, and Group-Evolving Agents) searches over the agent system itself rather than the seed for a fresh run, and addresses a different problem from ours~\citep{hu2025automateddesignagenticsystems,zhang2025aflowautomatingagenticworkflow,shang2025agentsquareautomaticllmagent,zhang2025evoflowevolvingdiverseagentic,zhang2026darwingodelmachineopenended,weng2026groupevolvingagentsopenendedselfimprovement}.

This leaves a specific gap: to the best of our knowledge, prior work does not combine population-based evolution with autonomous workspace agents while treating the initial state of each fresh run as the optimization object. We close that gap with \method{}, which evolves \emph{agent seeds} for fresh autonomous runs and transfers knowledge through algorithmically curated parent archives rather than a single persistent workspace. The genetic algorithm decides what a child starts with; the child agent decides whether and how to use it.

We instantiate \method{} for tabular AutoML on Weco-Kaggle Lite, where solution quality is measurable, engineering choices materially affect performance, and long-horizon agents can recover from errors that would end one-shot generation. Across the full \benchmarkcount{}-competition suite, \method{} outranks model-matched AIDE on \completedwinscount{}/\completedcount{} competitions. Across \lineagepaircount{} within-method parent-child contests, descendants conditioned on parent archives win \parentedwinrate\% of their contests, versus \initialwinrate\% for de novo proposals. A control that removes inheritance, in which every agent starts from scratch, places between a single long-lived agent and \method{} on the leaderboard, an ordering of three nested arms that is statistically significant (Section~\ref{sec:always-initial}).

\paragraph{Contributions.} Our key contributions are:
\begin{enumerate}[leftmargin=*,topsep=0pt,itemsep=2pt]
    \item \textbf{Agent-seed optimization}: We formalize a search space in which the chromosome is the initial state of a fresh autonomous run rather than a code artifact or persistent agent state (Section~\ref{sec:formulation}).
    \item \textbf{Evolutionary autonomous architecture}: We design a two-layer system with fresh-workspace agents, inherited parent archives, and deterministic 1:1 elite tournaments (Section~\ref{sec:architecture}).
    \item \textbf{LLM-native operators with online allocation}: We replace classical crossover/mutation with six task-based operators and adapt their usage online with a modified \hedge{} algorithm (Section~\ref{sec:operators}, Section~\ref{sec:hedge}).
    \item \textbf{Benchmark evidence}: On the full Weco-Kaggle Lite benchmark, \method{} outperforms model-matched AIDE and shows evidence, including a control that removes inheritance, that parent archives improve fresh autonomous runs on average (Section~\ref{sec:experiments}).
\end{enumerate}

\section{Related Work}
\label{sec:related}

\paragraph{Artifact-centric and bounded agentic methods.}
Many LLM systems for code and AutoML still optimize artifacts or bounded trajectories rather than fresh autonomous runs. AlphaCode, MLCopilot, and CAAFE perform one-shot or locally iterative artifact generation~\citep{li_competition-level_2022,zhang_mlcopilot_2024,hollmann_large_2023}. EvoPrompt, ReEvo, FunSearch, AIDE, and AlphaEvolve add orchestration, but the outer loop still stores and selects prompts, heuristics, or scripts~\citep{guo_evoprompt_2024,ye_reevo_2024,romera-paredes_mathematical_2024,jiang_aide_2025,novikov_alphaevolve_2025}. In AutoML, DS-Agent, Data Interpreter, AutoML-Agent, AutoKaggle, MLZero, SELA, I-MCTS, and SE-Agent equip LLMs with tools, planning, and iterative repair, but they optimize task work inside a fixed scaffold~\citep{guo_ds-agent_2024,hong_data_2025,trirat_automl-agent_2025,li_autokaggle_2024,fang_mlzero_2025,chi_sela_2024,liang_i-mcts_2025,lin2025seagentselfevolutiontrajectoryoptimization}. More recent systems fall in the same two regimes. AIRA-dojo~\citep{toledo2025airesearchagentsmachine}, ML-Master~\citep{liu2025mlmasteraiforaiintegrationexploration,zhu2026ultralonghorizonagenticsciencecognitive}, and, concurrently with our work, GEAR~\citep{jeddi2026geargeneticautoresearchagentic} and EvE~\citep{yu2026evolvingensembleagents} select artifacts: solution scripts, single edits to a training file, and the guidance files given to a coding agent. R\&D-Agent~\citep{yang2025rdagentllmagentframeworkautonomous} selects steps of a fixed research and development pipeline (Appendix~\ref{app:additional-positioning}).

\paragraph{Autonomous systems.}
The closest autonomous systems to \method{} are \emph{single-lineage autonomous solvers}, which launch one agent into a workspace and let it run open-endedly without a population layer. Karpathy's \emph{autoresearch}~\citep{karpathy_autoresearch_2026} commits successive edits to one fixed training script through an off-the-shelf coding CLI and logs every attempt; Agent K autonomously handles setup, debugging, and solution refinement on Kaggle tasks~\citep{grosnit_kolb-based_2025}. \method{} adds a population layer to this design (Section~\ref{sec:method}).

\paragraph{Classical AutoML.}
Classical AutoML methods such as Auto-WEKA, TPOT, Auto-sklearn, and AutoGluon search over predefined pipeline spaces using Bayesian optimization, genetic programming, and ensembling~\citep{thornton_auto-weka_2013,olson_evaluation_2016,feurer_efficient_2015,feurer_auto-sklearn_2022,erickson_autogluon-tabular_2020}. They are strong baselines for structured tabular tasks, but they do not use autonomous code generation. Appendix~\ref{app:additional-positioning} summarizes the positioning of representative LLM-based systems.

\paragraph{Agent-design search.}
A separate line of work optimizes the agent system itself: prompts, tools, modules, workflow graphs, or agent code. Automated Design of Agentic Systems, AFlow, AgentSquare, and EvoFlow search over bounded agentic workflows using meta-agent code generation, MCTS over workflow graphs, modular evolution, and population-based crossover and mutation respectively~\citep{hu2025automateddesignagenticsystems,zhang2025aflowautomatingagenticworkflow,shang2025agentsquareautomaticllmagent,zhang2025evoflowevolvingdiverseagentic}. Darwin G\"odel Machine and Group-Evolving Agents extend the same idea to autonomous coding agents, evolving the agent codebase or framework through self-modification across episodes~\citep{zhang2026darwingodelmachineopenended,weng2026groupevolvingagentsopenendedselfimprovement}. These methods address a different problem from ours: improving the agent system itself, typically across heterogeneous tasks. \method{} instead targets the well-defined evaluator setting, in which a fixed task has a measurable fitness signal, by holding the harness fixed. We do not treat them as direct comparators.

\section{Method}
\label{sec:method}

We present \method{}, a framework that searches in \emph{agent-seed space}. Instead of modifying code artifacts directly, the outer loop chooses the initial state for a fresh autonomous run, while the agent decides how to act inside that workspace. The result is a two-layer system: evolution controls inheritance and selection; the agent handles long-horizon problem solving.

\subsection{Agent-Seed Space Formulation}
\label{sec:formulation}

Each generation begins from a fresh seed rather than an accumulated dialogue history. Knowledge transfer happens through parent archives, so the optimized object is the initial state $c_0$.

Let $\mathcal{S}$ be the space of candidate solutions and let $h: \mathcal{S} \to \mathbb{R}$ be a direction-normalized fitness function. Artifact-centric systems such as AIDE~\citep{jiang_aide_2025} optimize directly in code space $\mathcal{S}$. We instead optimize over agent seeds $\mathcal{C}$, where each seed $c_0 \in \mathcal{C}$ launches an autonomous trajectory $\tau$ that produces a solution $s_\tau \in \mathcal{S}$:
\begin{equation}
    c_0^* = \arg\max_{c_0 \in \mathcal{C}} \mathbb{E}_{\tau \sim \text{Agent}(c_0)}[h(s_\tau)]
\end{equation}
The expectation matters because the same seed can induce different trajectories. The algorithm does not estimate this expectation: it does not average repeated runs of a seed, and selection acts on each child's realized solution. A seed contains a task template plus zero or more parent archives:
\begin{equation}
c_0 = \bigl(\textit{prompt}(\textit{task\_type}, \textit{context}),\; \{\textit{parent\_archive}_j\}_{j=1}^{n_p}\bigr)
\end{equation}
Each child is launched in an isolated workspace containing the execution environment, data, and any inherited artifacts. The seed specifies what is \emph{available} to the agent, not what must be injected into context. Parent material can therefore be inspected selectively, used late, or ignored entirely. Evolution operates on a stochastic seed-to-solution map rather than a deterministic artifact-edit path. This is an empirical inductive bias rather than a guarantee: stronger inherited archives need only improve the descendant distribution on average, not deterministically for every run. Once an agent produces a concrete solution artifact, the downstream workflow evaluates that fixed artifact deterministically with a fixed-seed evaluation harness, so stochasticity lies in generation rather than evaluation.

\subsection{Two-Layer Architecture}
\label{sec:architecture}

\method{} separates evolutionary coordination from agent execution (Figure~\ref{fig:architecture}). The evolution layer never continues a workspace in place and never specifies the agent's internal reasoning; it only chooses the seed for each new run. The agent layer executes that run in our general-purpose, goal-driven agent harness, which was not built or adapted specifically for this task. The harness is implemented as a LangGraph-based extended ReAct loop~\citep{yao_react_2023} with six functional nodes: Planning, Agent, Tools, Compression, Verification, and Finalize. Planning maintains a persistent execution plan, and Compression combines windowing with summarize/truncate/drop passes, backed by a durable memory file, so long-horizon trajectories remain within context limits (Appendix~\ref{app:context}). The tool suite covers file inspection, bash execution, and code editing; Appendix~\ref{app:tools} lists the full interface. In the reported configuration, the agent uses \agentgamodelname{}.

Within each run, the agent constructs its own context through a sequence of messages, tool calls, and tool results. It determines which files to inspect, which experiments to execute, and which failures to record; the solution emerges through this trajectory as the agent writes, evaluates, and revises code. The harness compresses the conversation as it approaches the context limit, and the run terminates when the agent judges the task complete. Appendix~\ref{app:chain} details this conversation-chain formulation, and Appendix~\ref{app:context} specifies the compression procedure.

Appendix~\ref{app:trajectory} illustrates this process through a Bike Sharing lineage and the recorded message sequence and activities of its final agent invocation.

The evolutionary layer maintains an elite pool of size $n$. At each iteration, every population slot spawns one child from a sampled task operator and selected parent archives. The child then competes only against the elite occupying the same slot from the previous iteration, with the comparison direction determined by the task metric. This deterministic 1:1 elite tournament preserves parallel lineages, avoids eliminating a lineage through unlucky pairing, and gives a direct measurement of whether the child improves on the incumbent elite.

For the AutoML instantiation, archived solutions are structured as leakage-safe executable workflows with outcome-independent feature engineering, outcome-dependent feature engineering, model definitions, experiment logs, and summary artifacts. A domain-specific workflow layer handles evaluation, artifact management, and submission generation. Child agents receive parent materials through \texttt{Previous Experiments/}: a copy of each selected parent's workspace containing its code, recorded attempts together with the code that produced them, notes, and version history, plus a generated summary. Virtual environments, caches, the read-only data mount, and the parent's own \texttt{Previous Experiments/} directory are excluded, so inheritance is exactly one generation deep.

\begin{figure}[t]
    \centering
    \resizebox{0.84\columnwidth}{!}{%
    \begin{tikzpicture}[
        node distance=0.35cm and 0.6cm,
        box/.style={rectangle, draw, rounded corners, minimum width=1.65cm, minimum height=0.55cm, align=center, font=\scriptsize},
        tierbox/.style={rectangle, draw, dashed, rounded corners, inner sep=0.25cm},
        arrow/.style={-{Stealth[scale=0.6]}, thick},
        looparrow/.style={-{Stealth[scale=0.5]}, thick, gray!60},
        label/.style={font=\scriptsize\bfseries}
    ]

    \node[box, fill=blue!15] (hedge) {\hedge{}\\Allocation};
    \node[box, fill=blue!15, right=of hedge] (assign) {Assign\\Task + Parents};
    \node[box, fill=blue!15, right=of assign] (tournament) {1:1 Elite\\Tournament};
    \node[box, fill=blue!15, right=of tournament] (elites) {Elite\\Population};

    \begin{scope}[on background layer]
        \node[tierbox, fill=blue!5, fit=(hedge)(assign)(tournament)(elites), label={[label]above:Evolution Layer}] (evol) {};
    \end{scope}

    \draw[arrow] (hedge) -- (assign);
    \draw[arrow] (assign) -- (tournament);
    \draw[arrow] (tournament) -- (elites);
    \draw[looparrow] (elites.south) -- ++(0,-0.15) -| node[pos=0.25, below, font=\tiny] {update} (hedge.south);

    \node[rectangle, draw, rounded corners, minimum width=3.5cm, minimum height=0.75cm, align=center, font=\scriptsize, fill=green!15, below=0.95cm of assign] (agent) {LLM Agents ($a_1, \ldots, a_n$)};

    \begin{scope}[on background layer]
        \node[tierbox, fill=green!5, fit=(agent), label={[label]above:Agent Layer}] (agentlayer) {};
    \end{scope}

    \draw[arrow, blue!60] (assign.south) -- node[left, font=\tiny] {$c_0^j$} (agent.north);
    \draw[arrow, red!60] (agent.north east) -- ++(0,0.3) -| node[pos=0.25, above, font=\tiny] {$s^j$} (tournament.south);

    \end{tikzpicture}%
    }
    \caption{Two-layer architecture. Evolution assigns a task type and parent archives to each fresh autonomous run; the resulting child competes only against its corresponding elite parent.}
    \label{fig:architecture}
\end{figure}

\subsection{Task-Based Genetic Operators}
\label{sec:operators}

Traditional crossover and mutation are ill-suited to text and code. We therefore use task-based operators that specify how a fresh child should use inherited context. Each operator launches one autonomous run that may itself execute multiple experiments; in the reported configuration, prompts instantiate a \agentgaattemptsperagent{}-experiment budget per invocation:
\begin{enumerate}[leftmargin=*,topsep=0pt,itemsep=1pt]
    \item \textbf{Initial} ($n_p=0$): no parents; start from scratch.
    \item \textbf{Continue} ($n_p \geq 1$): inspect parent solutions and refine what already works.
    \item \textbf{Ablation} ($n_p=1$): add or remove one component at a time to probe individual component contributions.
    \item \textbf{Merge} (normally $n_p=2$): combine components from two parents, typically building from one parent while borrowing components or ideas from the other; fall back to one parent if no second elite is available.
    \item \textbf{EDA} ($n_p=1$): analyze data first, then propose features or modeling changes guided by that analysis.
    \item \textbf{Jumpstart} ($n_p=1$): use inherited artifacts as raw material for a substantially different solution, implemented but inactive in the runs reported here.
\end{enumerate}
The budget is an instruction in the prompt rather than an enforced limit. Operators with parents first re-run the best inherited configuration once, outside the budget, and a child's fitness is the best score among all attempts it recorded. Appendices~\ref{app:harness} and~\ref{app:algorithm} provide the tool inventory, prompt structure, and compression pipeline of the harness, and task-template excerpts and configuration details of the evolution layer.

\subsection{Adaptive Allocation via Modified \hedge{}}
\label{sec:hedge}

Standard \hedge{}~\citep{freund_decision-theoretic_1997} is designed for settings with many rounds and cleaner reward signals than we have here. Our version updates operator probabilities from direction-aware parent-child improvements:
\begin{equation}
    \Delta_j =
    \begin{cases}
        s_j - s_{p_j} & \text{if the metric is higher-is-better} \\
        s_{p_j} - s_j & \text{otherwise}
    \end{cases}
\end{equation}
Positive $\Delta_j$ therefore always means ``child outperformed parent.'' Compared with vanilla \hedge{}, our variant is tuned for small, noisy gains: it skips updates when fewer than two operators are observed, converts operator means to ranks to suppress metric-scale effects, clips importance weighting at $\kappa=\hedgekappa{}$, updates log-weights with $\eta=\hedgeeta{}$, and enforces configured probability floors and ceilings after the softmax. These modifications are designed to stabilize allocation when operator usefulness changes over time. The exact reward and update equations, together with pseudocode and bound-enforcement details, are deferred to Appendix~\ref{app:algorithm}.

\subsection{Stopping Policy}
\label{sec:stopping}

The framework uses a configurable budget-and-patience stopping rule. Evolution terminates when either the maximum iteration budget is reached or the best-so-far elite score fails to strictly improve for the configured patience window. In the reported experiments, we set the maximum to \agentgamaxiterations{} iterations and the patience window to \agentgapatience{} iterations.

\section{Experiments}
\label{sec:experiments}

\subsection{Experimental Setup}
\label{sec:setup}
\paragraph{Benchmark.} We evaluate on Weco-Kaggle Lite~\citep{jiang_aide_2025}, a \benchmarkcount{}-competition benchmark of Kaggle tabular machine learning tasks. It was introduced by AIDE and enables direct comparison against human Kaggle leaderboards across regression and classification problems. Its competitions had between \benchmarkteamsmin{} and \benchmarkteamsmax{} teams, and Appendix~\ref{app:benchmarks} lists the full benchmark.

\paragraph{Evaluation protocol.} Within a run, each scored experiment is assigned the mean holdout score over \splitsperexperiment{} fixed shuffled train/validation splits, computed by a fixed evaluation harness. We call this the \emph{development score}. It decides every tournament, and Kaggle is never queried during the search. Only after all runs were complete were the test predictions of each run's selected solution submitted to Kaggle. Every reported benchmark number is the score and rank returned by Kaggle on the competition's official leaderboard. We use the recorded Kaggle scores and fixed-leaderboard ranks without rerunning models. We report \emph{Exceeds \% of Human} as $100(1-r/N)$, where $r$ is the recorded rank and $N$ is the reporting denominator, the number of teams on the leaderboard, and we average it across the competitions. Because it is rank-based, small absolute score differences can produce large rank changes when many teams are clustered (Appendix~\ref{app:threats-metric}).

\paragraph{Baselines.} \textbf{AIDE (DS)} is AIDE~\citep{jiang_aide_2025} with the same model as \method{} and otherwise the default configuration of its repository. It is the closest matched baseline, because both systems target Kaggle-style ML engineering problems, but AIDE searches over code artifacts whereas \method{} searches over autonomous agent seeds. \textbf{AutoGluon}~\citep{erickson_autogluon-tabular_2020} is a non-LLM AutoML baseline, used in version 1.6.1 with its \texttt{best\_quality} preset. As a reference, we also report the results published by the AIDE authors for their own GPT-4 Turbo (GPT-4T) runs, \textbf{AIDE (GPT-4T)}. We did not run these. \textbf{\single{}} is a single-agent control with one population slot and one iteration: a single agent receives the nominal allowance of a full \method{} run, \singlerunattemptbudget{} attempts ($\agentgapopsize{}\times\agentgamaxiterations{}\times\agentgaattemptsperagent{}$), and everything else is unchanged (Section~\ref{sec:single}). \textbf{Always-Initial} is \method{} with every agent a fresh \textbf{Initial} run (Section~\ref{sec:always-initial}). Both controls are ablations of \method{}: always-Initial keeps the \agentgapopsize{} population slots and removes inheritance, and \single{} removes the outer loop, so it is a single autonomous run in our harness and not a genetic algorithm.

\paragraph{Configuration.} The \method{} and model-matched AIDE arms use DeepSeek V4 Flash (0731), as do \single{} and the always-Initial control. \textbf{\method{}} maintains \agentgapopsize{} population slots ($n=\agentgapopsize$), runs for at most \agentgamaxiterations{} iterations, and stops early after \agentgapatience{} consecutive iterations without improvement of the best elite (Section~\ref{sec:stopping}). Each agent that inherits a parent is asked to first reproduce the parent's best configuration, and every agent is asked for \agentgaattemptsperagent{} new scored experiments (Section~\ref{sec:operators}). Appendix~\ref{app:hyperparam} lists the full configuration, and Section~\ref{sec:compute} reports the resources used by \method{}, the always-Initial control, and AIDE (DS).

\subsection{Benchmark Results}
\label{sec:benchmark-results}
Table~\ref{tab:completed-results} reports every competition. \method{} averages \meanagentgaexceeds\% \emph{Exceeds \% of Human}, against \meanaideexceeds\% for AIDE (DS) and \meanautogluonexceeds\% for AutoGluon. \single{} averages \meansinglerunexceeds\%, and AIDE (GPT-4T) \aidepublishedmeanexceeds\%. The always-Initial control averages \initonlymeanstop\% under \method{}'s stopping rule and \initonlymeanfull\% at the full budget (Section~\ref{sec:always-initial}).

Against AIDE (DS), the model-matched baseline, \method{} ranks higher on \completedwinscount{} of \benchmarkcount{} competitions (one-sided sign test, $p=\signtestpvsaide{}$), with a mean difference of +\meandeltaexceeds{} points (95\% bootstrap CI [\deltaaidecilow{}, \deltaaidecihigh{}]). The advantage does not depend on any single competition. Omitting any one of them leaves a mean difference between +\looaidelow{} and +\looaidehigh{} points.

Against AutoGluon, \method{} ranks higher on \autogluonwins{} competitions and lower on \autogluonlosses{}, but leads by +\meandeltavsautogluon{} points on average (CI [\deltaautogluoncilow{}, \deltaautogluoncihigh{}]), because AutoGluon is strong on several competitions and weak on a few.

AIDE (DS) outperforms the published AIDE (GPT-4T) results, against which \method{}'s record is \aidepublishedwins{}--\aidepublishedlosses{}--\aidepublishedties{} (wins--losses--ties), so AIDE (DS) is the stronger baseline. For further context, the AIDE authors also report \publishedhumanchatgpt\% for a human working with ChatGPT, \publishedhtwoautoml\% for H2O AutoML, and \publishedautogpt\% for AutoGPT (Table~\ref{tab:published-reference}).

The reported rankings are determined by Kaggle's official leaderboard rather than the internal validation score. The outer loop must produce executable workflows that can be evaluated across splits, retrained on full data, and submitted to Kaggle under the competition metric, and the gain over AIDE remains after that end-to-end procedure. These are descriptive system comparisons, not controlled causal evidence for inheritance. Appendix~\ref{app:leaderboard} reports every comparison in full, and the following subsections examine inheritance and search behavior within the recorded runs and with a control arm that removes inheritance (Section~\ref{sec:always-initial}).

\begin{table}[t]
\centering
\caption{\emph{Exceeds \% of Human} on Weco-Kaggle Lite (higher is better; the best of the arms we ran is in bold). \method{}, \single{}, the always-Initial control (Section~\ref{sec:always-initial}), and AIDE (DS) use \agentgamodelname{}. The AIDE (GPT-4T) column is reported by the AIDE authors, with GPT-4 Turbo, and was not run by us.}
\label{tab:completed-results}
\resizebox{\textwidth}{!}{
\begin{tabular}{lr@{\hskip 1.6em}rrrrrr}
\toprule
 & & & & & \multicolumn{2}{c}{\method{} (always-Initial)} & \\
\cmidrule(lr){6-7}
Competition & AIDE (GPT-4T) & AIDE (DS) & AutoGluon & \single{} & Same stopping rule & Full budget & \method{} \\
\midrule
Bike Sharing Demand & 91.92 & 92.04 & 9.07 & 10.92 & 93.25 & 88.81 & \textbf{93.34} \\
Cat in the Dat & 46.76 & 43.40 & 56.23 & 54.81 & 51.45 & 56.38 & \textbf{58.69} \\
House Prices & 72.74 & 79.99 & \textbf{96.97} & 83.15 & 81.12 & 79.03 & 96.50 \\
NYC Taxi Fare & 44.85 & 70.37 & 69.02 & 81.21 & 76.36 & 77.44 & \textbf{89.76} \\
Playground S3E14 & 52.21 & 80.98 & 66.76 & 76.77 & 80.87 & 80.87 & \textbf{84.82} \\
Playground S3E16 & 51.57 & 63.59 & 73.38 & 77.92 & \textbf{90.43} & \textbf{90.43} & 76.17 \\
Playground S3E19 & 36.80 & 9.45 & \textbf{55.11} & 43.02 & 43.61 & 43.61 & 43.10 \\
Playground S3E22 & 25.99 & 77.38 & 65.39 & 72.20 & \textbf{96.24} & 90.54 & 80.04 \\
Playground S3E24 & 65.71 & 77.12 & 66.28 & 72.88 & 80.47 & 80.47 & \textbf{99.06} \\
Playground S3E25 & 41.95 & 64.91 & 78.44 & 63.99 & 63.99 & \textbf{84.94} & 63.99 \\
TPS Aug 2022 & 79.25 & 65.43 & 74.22 & 77.29 & 49.23 & 63.53 & \textbf{92.17} \\
TPS Feb 2021 & 61.02 & 67.43 & 67.64 & 62.48 & 77.06 & 77.06 & \textbf{82.29} \\
TPS Feb 2022 & 43.68 & 43.28 & \textbf{49.40} & 42.96 & 38.90 & 35.40 & 39.38 \\
TPS Jan 2022 & 44.35 & 2.07 & 3.45 & 31.09 & 38.51 & 34.42 & \textbf{44.35} \\
TPS Jul 2021 & 12.98 & 25.73 & 29.83 & \textbf{67.85} & 27.36 & 29.68 & 25.97 \\
TMDB Box Office & 50.39 & 58.57 & 65.02 & 60.65 & 63.37 & 63.44 & \textbf{68.46} \\
\midrule
Mean & 51.38 & 57.61 & 57.89 & 61.20 & 65.76 & 67.25 & \textbf{71.13} \\
\bottomrule
\end{tabular}
}
\end{table}

\subsection{Inheritance and Lineage}
\label{sec:lineage}
We analyze within-run tournaments to compare the outcomes of archive-conditioned and de novo proposals. From iteration 2 onward, every child, whatever its operator, must beat the elite of its own population slot under the same model, agent harness, attempt budget, and scoring protocol. \textbf{Initial} children start from scratch, which makes them a de novo control for the operators that inherit a parent archive.

\paragraph{Tournament outcomes by inheritance.} The \benchmarkcount{} runs contain \lineagepaircount{} decided tournaments. Children that inherit an archive win \parentedwins{} of \parentedpairs{} (\parentedwinrate\%), whereas \textbf{Initial} children win only \initialwins{} of \initialpairs{} (\initialwinrate\%, Fisher's exact test, $p<10^{-10}$), and Table~\ref{tab:operators} in Appendix~\ref{app:phases} shows the same pattern for every inheriting operator. The difference also appears within the \sameiteriterations{} iterations in which the same run launched both kinds of child against their respective slot elites: the win rates are \sameiterinitialwinrate\% and \sameiterparentedwinrate\%, respectively. A child that starts from an optimized parent is expected to beat it more often, so Section~\ref{sec:always-initial} tests inheritance with a separate arm. Win rates also change over the course of the search. As incumbent development scores improve, the win rate of \textbf{Initial} children falls from \initialearlywinrate\% in iterations 2--5 to \initiallatewinrate\% in iterations 16--20, while inheriting children still win \parentedlatewinrate\% (Figure~\ref{fig:phases}, Appendix~\ref{app:phases}).

An inheriting child records a worse development score than the reference elite in \childworsepct\% of tournaments, compared with \initialworsepct\% for children that start from scratch (\textbf{Initial}). Ties count as losses, so every win is a strict improvement. Appendix~\ref{app:phases} presents the controls behind these comparisons, and Appendix~\ref{app:threats-causal} discusses selection on the development splits, to which both kinds of child are subject.

\paragraph{How children use what they inherit.} The seed only makes the archive available. The agents' tool logs show that every inheriting agent that ran had detected archive accesses, with a median of \inheritedtouchesmedian{} tool calls (\inheritedtouchsharepct\% of their calls), the first of which occurs \inheritedfirsttouchpct\% of the way into the run. All \inheritingran{} had calls accessing the parent's summary and recorded attempts; \openscodepct\% also accessed solution code and \opensgitpct\% version history. Access includes reading, listing, copying, and executing inherited material, as defined in Appendix~\ref{app:phases}. As the search proceeds, the median inherited archive grows from \archivefilesearly{} to \archivefileslate{} text files, yet agents access it with about the same number of calls, \archivetouchesearly{} at iteration 2 and \archivetoucheslate{} in iterations 16--20 (Table~\ref{tab:archive-growth}, Appendix~\ref{app:phases}).

\paragraph{Exploration and parallel lineages.} Within-run experiments include both improvements and regressions. Of \newexperimentscount{} new scored experiments, \experimentsbetterpct\% beat the inherited parent, \experimentsworsepct\% were worse, and the remaining \experimentsequalpct\% matched it exactly. Each agent keeps its best experiment, and the tournament discards the rest. The 1:1 tournament keeps \agentgapopsize{} lineages alive instead of truncating to the global best. The lead changed between population slots in \leadchangeruns{} of \benchmarkcount{} runs. In \trailingatmidpoint{} of the \longruns{} runs with at least 10 iterations, the slot that finished best was not leading at the midpoint, and in \crossslotancestryruns{} of \benchmarkcount{} runs, the final best elite descends from two or more slots. Appendix~\ref{app:dynamics} describes the search dynamics in full.

\subsection{Always-Initial Control}
\label{sec:always-initial}
To test inheritance directly, we ran \method{} with the operator fixed to \textbf{Initial}, so that every agent starts from scratch and nothing is inherited. The control ran all \agentgamaxiterations{} iterations with the stopping rule disabled, which gives two variants: the run cut where \method{}'s stopping rule would have ended it, and the full budget (Table~\ref{tab:completed-results}, Appendix~\ref{app:always-initial}).

\paragraph{A ladder of nested arms.} \single{}, the control, and \method{} share the model, harness, and attempt allowance, and each adds to the previous one: a single agent in one long conversation; then the outer loop, with \agentgapopsize{} population slots of fresh conversations and tournaments and nothing inherited; then inheritance of the parents' workspaces. Mean \emph{Exceeds \% of Human} rises in that order, from \meansinglerunexceeds\% to \initonlymeanstop\% (\initonlymeanfull\% at the full budget) to \meanagentgaexceeds\%, and the ordering is statistically significant (Page's trend test, one-sided $p=\initonlytrendpstop{}$; $p=\initonlytrendpfull{}$ with the full budget). The last step alone is less certain: \method{} ranks higher than the same-stopping-rule variant on \initonlywinsstop{} competitions and lower on \initonlylossesstop{} (one-sided sign test, $p=\initonlysignpstop{}$), and the 95\% bootstrap confidence interval of the mean difference of +\initonlydeltastop{} points includes zero [$\initonlycilowstop{}$, $\initonlycihighstop{}$]. The control itself outranks AIDE (DS) by +\initonlyaidedeltafull{} points at the full budget (CI [\initonlyaidecilowfull{}, \initonlyaidecihighfull{}]), so independent runs from scratch that keep only the best solution in each slot account for a large part of \method{}'s advantage over AIDE, and inheritance adds to it.

\paragraph{Search objective.} From iteration 2 onward, the best elite improves in \initonlyimprovedpct\% of the control's iterations, against \iterationsimprovedpct\% for \method{}, so \method{}'s stopping rule would have ended the control after a median of \initonlystopmedian{} iterations, whereas \initonlygacapruns{} of the \benchmarkcount{} \method{} runs reach the \agentgamaxiterations{}-iteration cap.

\subsection{Single-Agent Control}
\label{sec:single}
We compare \method{} with \single{} to assess the effect of the outer loop under the same nominal attempt allowance. \single{} uses the same system with one population slot and one iteration, and in all \benchmarkcount{} competitions its task prompt differs from that of an iteration-1 \method{} agent in \textbf{exactly one line}, shown below.
\begin{center}
\small
\begin{tabular}{@{}lcl@{}}
\toprule
 & Population $\times$ iterations & The one line that differs \\
\midrule
\method{} & $\agentgapopsize{}\times\agentgamaxiterations{}$ & \texttt{Run \agentgaattemptsperagent{} attempts.\ For each one:} \\
\single{} & $\mathbf{1\times1}$ & \texttt{Run \singlerunattemptbudget{} attempts.\ For each one:} \\
\bottomrule
\end{tabular}
\end{center}
The model, harness, tools, and evaluation are identical.

Given this allowance in a single conversation, \singlerunstoppedearly{} of \benchmarkcount{} agents stopped of their own accord before using 90\% of it, after a median of \singlerunearlymedian{} scored experiments (\singlerunallowancepct\% of the allowance).

All \singlerunstoppedearly{} early-stopping agents explicitly refer to the allowance while planning, \singlerunnotedunspent{} note the unspent allowance in their own final audit, and \singlerunsaidconverged{} of these attribute it to convergence or noise. On \casestudylabel{}, for example, the agent wrote, ``\singlerunquote''

Of the \singlerunstoppedearly{} agents that stopped early, \singlerunstillimproving{} recorded their best development score in the final fifth of their experiments. In \gabeyondbetter{} of the \gabeyondcompared{} competitions, \method{} went on to reach a better development score than the one at which the single agent stopped.

Both kinds of conversation grow with every call. The single agent's grew from \chainonestart{} to \chainoneend{} tokens over a median of \chainonecalls{} model calls, about \chainoneendpct\% of the \chainlimit{}-token context limit, and an \method{} agent's grows from \chaingastart{} to \chaingaend{} tokens over \chaingacalls{} calls, after which the next generation starts a new one. Each \method{} agent is asked for only \agentgaattemptsperagent{} new experiments, \agentgaagentsdeliverpct\% deliver at least that many, and relaunching agents from fresh seeds carries a run to a median of \agentgaexperimentsperrun{} scored experiments. Appendix~\ref{app:single} reports every competition and the conversation lengths in detail.

On the leaderboard, \method{} ranks higher than \single{} on \singlerunwins{} competitions, lower on \singlerunlosses{}, and equal on \singlerunties{} (one-sided sign test, $p=\singlerunsigntestp{}$). The mean difference is +\meandeltavssinglerun{} percentage points, although its 95\% bootstrap confidence interval includes zero [$\deltasinglecilow{}$, $\deltasinglecihigh{}$] (Appendix~\ref{app:leaderboard}). Two of the three losses, including the largest, are on \verworstlabel{} and \verduplabel{}, the two competitions whose verifier does not reflect the test set (Section~\ref{sec:verifier}). The comparison is matched in allowance, not in compute. \method{} ran \gaoversingleexperiments$\times$ as many scored experiments at \gaoversinglecost$\times$ the LLM cost. \method{}'s advantage comes from continuing the search, not from efficiency at equal budget (Appendix~\ref{app:single}).

\subsection{Verifier--Leaderboard Alignment}
\label{sec:verifier}
\label{sec:case-study}
\method{} uses the development score over \splitsperexperiment{} fixed random splits as its verifier, guiding selection without querying Kaggle during search. We examine whether improvements on this verifier translate into better performance on the unseen test set, as measured by the Kaggle leaderboard. We traced the reported member of four \agentgamaxiterations{}-iteration runs, the two highest- and the two lowest-ranked, back through its ancestors to the iteration-1 \textbf{Initial} member, and submitted every generation of each line to Kaggle without rerunning any model. Including additional elite checkpoints, we scored \verfilesscored{} distinct files (Appendix~\ref{app:verifier}). Figure~\ref{fig:verifier} in Appendix~\ref{app:verifier} shows paired development-score and leaderboard trajectories over successive members of each selected lineage. Along every line development performance improved, by \verdevmin\% to \verdevmax\%, because an elite is replaced only by a strict improvement.

In the two highest-ranked traced runs, development-score improvements were accompanied by leaderboard improvements. \verbestlabel{} rose from \emph{Exceeds} \verbestexfirst{} to \verbestexlast{} over \verbestgenerations{} generations, from rank \verbestrankfirst{} to rank \verbestranklast{}, and \versecondbestlabel{} rose from \versecondbestexfirst{} to \versecondbestexlast{}. Across these two lines, \verbeststepsup{} of \verbeststeps{} generation steps moved up the leaderboard and \verbeststepsdown{} moved down.

In the two lowest-ranked traced runs, we observed data properties that may contribute to the mismatch between development and leaderboard performance. \verworstlabel{} has a temporal train--test split: its test period begins where its training period ends (\verteststart{}) and extends to \verteststop{}, whereas the development verifier uses randomly held-out training rows. In \verduplabel{}, \verduppct\% of a random holdout has an exact duplicate among the training rows. These observations suggest possible limitations of random-split evaluation on these tasks, but do not establish the cause of the observed mismatch.

Along the \verworstlabel{} line development performance improved by \verworstdev\% while \emph{Exceeds} went from \verworstexfirst{} to \verworstexlast{}, and the \verduplabel{} line stayed nearly flat, from \verdupexfirst{} to \verdupexlast{}. These are also two of the three competitions on which \method{} trails the single agent (Section~\ref{sec:single}).

\subsection{Computational Cost}
\label{sec:compute}
The \benchmarkcount{} \method{} runs comprise \fevaltotaliters{} iterations, \agentgatotalinvocations{} agent invocations, and \fevaltotalevals{} scored experiments (\agentgatotalmodelfits{} model fits), and recorded \agentgatotaltokensbillions{}B tokens over \agentgatotalllmcalls{} LLM calls. Because \agentgacachehitpct\% of prompt tokens were served from the cache, the total cost was about \$\agentgatotalcostusd{}, or \$\agentgacostperrunusd{} per competition. The always-Initial control ran \initonlyexperimentsfull{} scored experiments at the full budget and recorded \initonlytokensfull{}B tokens (about \$\initonlycostfull{}). Model training and scoring dominate wall-clock time, and waiting for the LLM accounts for only \agentgallmsharepct\% of agent time. AIDE (DS) ran at its default budget, which uses far less compute, so the comparison in Section~\ref{sec:benchmark-results} matches the model rather than the compute (Appendix~\ref{app:threats-causal}). Section~\ref{sec:single} shows that, within our harness, giving a single agent a larger allowance does not by itself close the gap to \method{}. Appendix~\ref{app:fevals} gives per-competition accounting.

\section{Discussion}
\label{sec:discussion}

\paragraph{Agent-Seed Space.}
The agent-seed formulation mirrors how a human data scientist might be briefed. A new task can begin either from scratch or from prior working solutions; the seed is exactly that reusable starting point, not a deterministic recipe for the final answer. This is useful when prior context matters but carrying forward one growing workspace is undesirable. By resetting each run to a compact initial condition while preserving parent archives in the workspace, \method{} separates what is inherited from how the next agent chooses to use it. Mechanically, it replaces one ever-growing conversation with a sequence of short conversations that pass their state on through files. The single-agent control provides a practical reason for this reset (Section~\ref{sec:single}). Given the whole allowance in one conversation, most agents declared the task complete after using a small fraction of it, whereas short, bounded tasks in fresh conversations, with the archive carrying state between them, sustained the search. This is an observation about one model and one harness, not a general law. We conjecture that this reflects how current models are trained: on long-horizon tasks, but rarely on horizons of hundreds of experiments, so that an agent concludes too early that it is done. We have not tested this conjecture, and the behavior may change with future models.

\paragraph{Self-Directed Context Construction.}
Unlike retrieval-style systems, \method{} does not inject pre-selected snippets into prompt context. Parent archives remain available in the workspace for the agent to inspect, ignore, or reinterpret while pursuing the task. In our runs, every inheriting agent that executed had detected archive accesses, including the parent's summary and recorded attempts, while code access was selective (Section~\ref{sec:lineage}). The effective working context is therefore constructed by the trajectory itself: file reads, experiments, failures, and replanning decisions. In that sense, the search object is the reusable starting condition, while the concrete solution remains a downstream product of the run.

\paragraph{Role of the Verifier.}
The verifier determines which solutions survive and what subsequent agents inherit. For this search to improve performance on unseen data, the development score must be a useful proxy for the target test distribution: otherwise, the system can steadily improve its measured fitness without improving generalization. Averaging over multiple splits reduces reliance on any one split, but cannot correct a systematic mismatch between the development protocol and the test set. Section~\ref{sec:verifier} illustrates this distinction: development performance improved along all four traced lineages, whereas leaderboard performance improved clearly only in the two highest-ranked runs. The temporal structure and duplicate overlap observed in the other two runs suggest that choosing a development protocol that reflects the intended evaluation setting is essential to making evolutionary progress meaningful beyond the verifier itself.

\paragraph{Limitations.}
\emph{Compute cost.} Long-horizon agents consume far more tokens and wall-clock time than artifact-centric generation (Section~\ref{sec:compute}), and multi-split evaluation adds further CPU and wall-clock evaluation cost rather than additional LLM token cost. Our comparison with AIDE matches the model, not the compute.

\emph{The verifier.} The alignment analysis in Section~\ref{sec:verifier} motivates rerunning the affected competitions with time-ordered or duplicate-aware splits. These alternatives would better reflect the temporal structure or reduce duplicate overlap, respectively, but we have not tested whether they would improve leaderboard performance.

\emph{Workflow and capability.} The method requires a bounded, automatable fitness signal and a domain-specific workflow layer that makes candidate solutions comparable, and its performance is limited by the base model, that layer, and the search space exposed to the agent. Slow evaluations and increasingly complex descendants reduce the number of experiments a fixed budget allows, and a new domain requires engineering how artifacts are structured, inherited, and evaluated.

\emph{Component isolation.} The present experiments evaluate \method{} as an end-to-end system rather than a full factorial ablation of every design choice. In particular, we do not separately isolate the marginal contribution of \hedge{} allocation, population size, archive inheritance, task-template design, and the domain-specific workflow layer under matched compute. The single-agent control removes the outer loop under a matched attempt allowance (Section~\ref{sec:single}), the always-Initial control removes inheritance (Section~\ref{sec:always-initial}), and the parent-child tournament analysis provides direct evidence that archive-conditioned descendants are more competitive than de novo \textbf{Initial} proposals, but all results come from one base model, one agent harness, and one benchmark domain, and a complete controlled ablation remains future work.

\section{Conclusion}
\label{sec:conclusion}
We introduced \method{}, a framework that searches in agent-seed space by launching fresh autonomous runs with inherited parent archives. Deterministic 1:1 elite tournaments and adaptive \hedge{} allocation coordinate population-level search. On Weco-Kaggle Lite, the DeepSeek V4 Flash (0731) implementation outranks model-matched AIDE on \completedwinscount{}/\benchmarkcount{} tasks, with a mean Exceeds advantage of \meandeltaexceeds{} percentage points, and exceeds a single long-lived agent given the same attempt allowance by a mean of \meandeltavssinglerun{} points. A control without inheritance places between that agent and \method{}, and the ordering of the three arms is statistically significant. These system-level results motivate further study of agent-seed optimization as a way to convert budget into sustained search; conclusive attribution to inheritance, compute-matched baselines, ablations of adaptive allocation and population size, and domains beyond tabular AutoML remain open.

\subsection*{AI use statement}

Large language models are the object of study in this work: the agents that \method{} evolves are LLM agents (Section~\ref{sec:method}), the models are named in Section~\ref{sec:experiments}, and every solution an agent writes is scored by a fixed evaluation harness that involves no LLM. The rest of this statement is about the authors' own use of generative AI tools.

In this work, we used generative AI tools for research execution: to implement the analysis of the archived runs (scripts that extract, deduplicate and reformat the run records and compute the reported statistics and tables from them), and to support the analysis and interpretation of those records, including auditing the numbers in the manuscript against them. We also used generative AI tools as a sounding board, to suggest and discuss alternatives for the analyses, the experimental choices, and the presentation, which the authors then accepted, adapted, or rejected. We have not used generative AI tools to develop the conceptual framework or to propose the research hypotheses. The central idea, evolving the starting conditions of long-horizon autonomous agent runs with a genetic algorithm, as well as the method and the final experimental design, are the authors' own. Generating synthetic data sets, formulating mathematical claims, writing proofs and translation are not applicable to this work. Additionally, we used generative AI tools (LLM coding and writing assistants, including Claude, Anthropic) to create and edit software code and figures, to draft and edit parts of the text for readability, and to prepare the \LaTeX{} formatting.

We have reviewed all AI-assisted work. AI-assisted code and text were checked by two of the authors. Every reported number is produced by a script that reads the archived run records rather than typed by hand, and the tournament tallies recomputed by those scripts were checked against the summaries that the search itself wrote at run time. We take responsibility for the final content of this work, including text, claims and artifacts produced with the aid of generative AI.

\subsection*{Ethics statement}

\method{} could broaden access to complex optimization tasks by automating parts of the search for executable solutions. Potential harms include misuse for harmful code generation, displacement of some specialized engineering work, and increased environmental cost from long autonomous trajectories. Practical deployment should therefore pair the system with domain restrictions, monitoring, and compute controls.

Because \method{} searches over long-horizon autonomous runs, deployment should use sandboxed tools, audit logs, explicit budget caps, and task-specific rules that prevent unsafe multi-step behavior and wasteful leaderboard chasing.

\subsection*{Reproducibility statement}

Section~\ref{sec:experiments} summarizes the configuration used in this manuscript, and Appendix~\ref{app:hyperparam} lists the full experiment settings. These values correspond to the experiment environment configuration used for the reported runs rather than fallback library defaults. LLM generation remains stochastic, but evaluation of a fixed archived solution is deterministic under the fixed seeds of the workflow harness. The experimental records used for this manuscript include all reported benchmark runs together with all their archived iterations, agent chains, and code artifacts. Upon acceptance, we will release the code for \method{}: the genetic-algorithm evolution layer and the tabular AutoML instantiation used in this work.

We distinguish reproducibility of \emph{search} from reproducibility of \emph{evaluation}: the full search is stochastic, but the system preserves every iteration, run, and agent chain together with all archives and code, providing a complete audit trail for the reported tournaments and plots.

\bibliographystyle{iclr\iclryear/iclr\iclryear_conference}
\bibliography{references}

\begin{thebibliography}{37}
\providecommand{\natexlab}[1]{#1}
\providecommand{\url}[1]{\texttt{#1}}
\expandafter\ifx\csname urlstyle\endcsname\relax
  \providecommand{\doi}[1]{doi: #1}\else
  \providecommand{\doi}{doi: \begingroup \urlstyle{rm}\Url}\fi

\bibitem[Chi et~al.(2024)Chi, Lin, Hong, Pan, Fei, Mei, Liu, Pang, Kwok, Zhang,
  Liu, and Wu]{chi_sela_2024}
Yizhou Chi, Yizhang Lin, Sirui Hong, Duyi Pan, Yaying Fei, Guanghao Mei,
  Bangbang Liu, Tianqi Pang, Jacky Kwok, Ceyao Zhang, Bang Liu, and Chenglin
  Wu.
\newblock {SELA}: {Tree}-{Search} {Enhanced} {LLM} {Agents} for {Automated}
  {Machine} {Learning}, October 2024.
\newblock URL \url{http://arxiv.org/abs/2410.17238}.
\newblock arXiv:2410.17238 [cs].

\bibitem[Erickson et~al.(2020)Erickson, Mueller, Shirkov, Zhang, Larroy, Li,
  and Smola]{erickson_autogluon-tabular_2020}
Nick Erickson, Jonas Mueller, Alexander Shirkov, Hang Zhang, Pedro Larroy,
  Mu~Li, and Alexander Smola.
\newblock {AutoGluon}-{Tabular}: {Robust} and {Accurate} {AutoML} for
  {Structured} {Data}, March 2020.
\newblock URL \url{http://arxiv.org/abs/2003.06505}.
\newblock arXiv:2003.06505 [stat].

\bibitem[Fang et~al.(2025)Fang, Han, Erickson, Zhang, Zhou, Dagar, Zhang,
  Turkmen, Hu, Rangwala, Wu, Wang, and Karypis]{fang_mlzero_2025}
Haoyang Fang, Boran Han, Nick Erickson, Xiyuan Zhang, Su~Zhou, Anirudh Dagar,
  Jiani Zhang, Ali~Caner Turkmen, Cuixiong Hu, Huzefa Rangwala, Ying~Nian Wu,
  Bernie Wang, and George Karypis.
\newblock {MLZero}: {A} {Multi}-{Agent} {System} for {End}-to-end {Machine}
  {Learning} {Automation}.
\newblock In \emph{Advances in {Neural} {Information} {Processing} {Systems} 38
  ({NeurIPS})}, 2025.
\newblock \doi{10.48550/arXiv.2505.13941}.
\newblock URL \url{http://arxiv.org/abs/2505.13941}.

\bibitem[Feurer et~al.(2015)Feurer, Klein, Eggensperger, Springenberg, Blum,
  and Hutter]{feurer_efficient_2015}
Matthias Feurer, Aaron Klein, Katharina Eggensperger, Jost Springenberg, Manuel
  Blum, and Frank Hutter.
\newblock Efficient and {Robust} {Automated} {Machine} {Learning}.
\newblock In \emph{Advances in {Neural} {Information} {Processing} {Systems}},
  volume~28. Curran Associates, Inc., 2015.
\newblock URL
  \url{https://papers.nips.cc/paper/2015/hash/11d0e6287202fced83f79975ec59a3a6-Abstract.html}.

\bibitem[Feurer et~al.(2022)Feurer, Eggensperger, Falkner, Lindauer, and
  Hutter]{feurer_auto-sklearn_2022}
Matthias Feurer, Katharina Eggensperger, Stefan Falkner, Marius Lindauer, and
  Frank Hutter.
\newblock Auto-{Sklearn} 2.0: {Hands}-free {AutoML} via {Meta}-{Learning}.
\newblock \emph{Journal of Machine Learning Research}, 23\penalty0
  (261):\penalty0 1--61, 2022.
\newblock ISSN 1533-7928.
\newblock URL \url{http://jmlr.org/papers/v23/21-0992.html}.

\bibitem[Freund \& Schapire(1997)Freund and
  Schapire]{freund_decision-theoretic_1997}
Yoav Freund and Robert~E Schapire.
\newblock A {Decision}-{Theoretic} {Generalization} of {On}-{Line} {Learning}
  and an {Application} to {Boosting}.
\newblock \emph{Journal of Computer and System Sciences}, 55\penalty0
  (1):\penalty0 119--139, August 1997.
\newblock ISSN 0022-0000.
\newblock \doi{10.1006/jcss.1997.1504}.
\newblock URL
  \url{https://www.sciencedirect.com/science/article/pii/S002200009791504X}.

\bibitem[Grosnit et~al.(2024)Grosnit, Maraval, N, Zhao, Doran, Paolo, Thomas,
  Gonzalez, Kumar, Khandelwal, Benechehab, Cherkaoui, El-Hili, Shao, Hao, Yao,
  Kégl, Bou-Ammar, and Wang]{grosnit_kolb-based_2025}
Antoine Grosnit, Alexandre Maraval, Refinath~S. N, Zichao Zhao, James Doran,
  Giuseppe Paolo, Albert Thomas, Jonas Gonzalez, Abhineet Kumar, Khyati
  Khandelwal, Abdelhakim Benechehab, Hamza Cherkaoui, Youssef~Attia El-Hili,
  Kun Shao, Jianye Hao, Jun Yao, Balázs Kégl, Haitham Bou-Ammar, and Jun
  Wang.
\newblock Kolb-{Based} {Experiential} {Learning} for {Generalist} {Agents} with
  {Human}-{Level} {Kaggle} {Data} {Science} {Performance}, November 2024.
\newblock URL \url{http://arxiv.org/abs/2411.03562}.
\newblock arXiv:2411.03562 [cs].

\bibitem[Guo et~al.(2024{\natexlab{a}})Guo, Wang, Guo, Li, Song, Tan, Liu,
  Bian, and Yang]{guo_evoprompt_2024}
Qingyan Guo, Rui Wang, Junliang Guo, Bei Li, Kaitao Song, Xu~Tan, Guoqing Liu,
  Jiang Bian, and Yujiu Yang.
\newblock {EvoPrompt}: {Connecting} {LLMs} with {Evolutionary} {Algorithms}
  {Yields} {Powerful} {Prompt} {Optimizers}.
\newblock In \emph{Proceedings of the 12th {International} {Conference} on
  {Learning} {Representations} ({ICLR})}, 2024{\natexlab{a}}.
\newblock \doi{10.48550/arXiv.2309.08532}.
\newblock URL \url{http://arxiv.org/abs/2309.08532}.

\bibitem[Guo et~al.(2024{\natexlab{b}})Guo, Deng, Wen, Chen, Chang, and
  Wang]{guo_ds-agent_2024}
Siyuan Guo, Cheng Deng, Ying Wen, Hechang Chen, Yi~Chang, and Jun Wang.
\newblock {DS}-{Agent}: {Automated} {Data} {Science} by {Empowering} {Large}
  {Language} {Models} with {Case}-{Based} {Reasoning}.
\newblock In \emph{Proceedings of the 41st {International} {Conference} on
  {Machine} {Learning} ({ICML})}, pp.\  16813--16848, 2024{\natexlab{b}}.
\newblock \doi{10.48550/arXiv.2402.17453}.
\newblock URL \url{http://arxiv.org/abs/2402.17453}.

\bibitem[Hollmann et~al.(2023)Hollmann, Müller, and
  Hutter]{hollmann_large_2023}
Noah Hollmann, Samuel Müller, and Frank Hutter.
\newblock Large {Language} {Models} for {Automated} {Data} {Science}:
  {Introducing} {CAAFE} for {Context}-{Aware} {Automated} {Feature}
  {Engineering}.
\newblock In \emph{Advances in {Neural} {Information} {Processing} {Systems} 36
  ({NeurIPS})}, 2023.
\newblock \doi{10.48550/arXiv.2305.03403}.
\newblock URL \url{http://arxiv.org/abs/2305.03403}.

\bibitem[Hong et~al.(2025)Hong, Lin, Liu, Liu, Wu, Zhang, Wei, Li, Chen, Zhang,
  Wang, Zhang, Zhang, Yang, Zhuge, Guo, Zhou, Tao, Tang, Lu, Zheng, Liang, Fei,
  Cheng, Gou, Xu, and Wu]{hong_data_2025}
Sirui Hong, Yizhang Lin, Bang Liu, Bangbang Liu, Binhao Wu, Ceyao Zhang,
  Chenxing Wei, Danyang Li, Jiaqi Chen, Jiayi Zhang, Jinlin Wang, Li~Zhang,
  Lingyao Zhang, Min Yang, Mingchen Zhuge, Taicheng Guo, Tuo Zhou, Wei Tao,
  Xiangru Tang, Xiangtao Lu, Xiawu Zheng, Xinbing Liang, Yaying Fei, Yuheng
  Cheng, Zhibin Gou, Zongze Xu, and Chenglin Wu.
\newblock Data {Interpreter}: {An} {LLM} {Agent} {For} {Data} {Science}.
\newblock In \emph{Findings of the {Association} for {Computational}
  {Linguistics}: {ACL} 2025}, 2025.
\newblock \doi{10.48550/arXiv.2402.18679}.
\newblock URL \url{http://arxiv.org/abs/2402.18679}.

\bibitem[Hu et~al.(2025)Hu, Lu, and Clune]{hu2025automateddesignagenticsystems}
Shengran Hu, Cong Lu, and Jeff Clune.
\newblock Automated design of agentic systems, 2025.
\newblock URL \url{https://arxiv.org/abs/2408.08435}.

\bibitem[Jeddi et~al.(2026)Jeddi, Le, Karaimer, Derpanis, and
  Taati]{jeddi2026geargeneticautoresearchagentic}
Ahmadreza Jeddi, Minh~Ngoc Le, Hakki~C. Karaimer, Konstantinos~G. Derpanis, and
  Babak Taati.
\newblock {GEAR}: Genetic {AutoResearch} for agentic code evolution, 2026.
\newblock URL \url{https://arxiv.org/abs/2605.13874}.

\bibitem[Jiang et~al.(2025)Jiang, Schmidt, Srikanth, Xu, Kaplan, Jacenko, and
  Wu]{jiang_aide_2025}
Zhengyao Jiang, Dominik Schmidt, Dhruv Srikanth, Dixing Xu, Ian Kaplan, Deniss
  Jacenko, and Yuxiang Wu.
\newblock {AIDE}: {AI}-{Driven} {Exploration} in the {Space} of {Code},
  February 2025.
\newblock URL \url{http://arxiv.org/abs/2502.13138}.
\newblock arXiv:2502.13138 [cs].

\bibitem[Karpathy(2026)]{karpathy_autoresearch_2026}
Andrej Karpathy.
\newblock {AutoResearch}: {AI} agents running research on single-{GPU} nanochat
  training automatically, 2026.
\newblock URL \url{https://github.com/karpathy/autoresearch}.
\newblock GitHub repository.

\bibitem[Li et~al.(2022)Li, Choi, Chung, Kushman, Schrittwieser, Leblond,
  Eccles, Keeling, Gimeno, Dal~Lago, Hubert, Choy, de~Masson~d’Autume,
  Babuschkin, Chen, Huang, Welbl, Gowal, Cherepanov, Molloy, Mankowitz,
  Sutherland~Robson, Kohli, de~Freitas, Kavukcuoglu, and
  Vinyals]{li_competition-level_2022}
Yujia Li, David Choi, Junyoung Chung, Nate Kushman, Julian Schrittwieser, Rémi
  Leblond, Tom Eccles, James Keeling, Felix Gimeno, Agustin Dal~Lago, Thomas
  Hubert, Peter Choy, Cyprien de~Masson~d’Autume, Igor Babuschkin, Xinyun
  Chen, Po-Sen Huang, Johannes Welbl, Sven Gowal, Alexey Cherepanov, James
  Molloy, Daniel~J. Mankowitz, Esme Sutherland~Robson, Pushmeet Kohli, Nando
  de~Freitas, Koray Kavukcuoglu, and Oriol Vinyals.
\newblock Competition-level code generation with {AlphaCode}.
\newblock \emph{Science}, 378\penalty0 (6624):\penalty0 1092--1097, December
  2022.
\newblock \doi{10.1126/science.abq1158}.
\newblock URL \url{https://www.science.org/doi/10.1126/science.abq1158}.

\bibitem[Li et~al.(2025)Li, Zang, Ma, Guo, Zheng, Liu, Niu, Wang, Yang, Liu,
  Zhong, Zhou, Huang, and Zhang]{li_autokaggle_2024}
Ziming Li, Qianbo Zang, David Ma, Jiawei Guo, Tuney Zheng, Minghao Liu, Xinyao
  Niu, Yue Wang, Jian Yang, Jiaheng Liu, Wanjun Zhong, Wangchunshu Zhou, Wenhao
  Huang, and Ge~Zhang.
\newblock {AutoKaggle}: {A} {Multi}-{Agent} {Framework} for {Autonomous} {Data}
  {Science} {Competitions}.
\newblock In \emph{Proceedings of the 13th {International} {Conference} on
  {Learning} {Representations} ({ICLR})}, 2025.
\newblock \doi{10.48550/arXiv.2410.20424}.
\newblock URL \url{http://arxiv.org/abs/2410.20424}.

\bibitem[Liang et~al.(2025)Liang, Wei, Xu, Chen, Qian, and
  Wu]{liang_i-mcts_2025}
Zujie Liang, Feng Wei, Wujiang Xu, Lin Chen, Yuxi Qian, and Xinhui Wu.
\newblock I-{MCTS}: {Enhancing} {Agentic} {AutoML} via {Introspective} {Monte}
  {Carlo} {Tree} {Search}, February 2025.
\newblock URL \url{http://arxiv.org/abs/2502.14693}.
\newblock arXiv:2502.14693 [cs].

\bibitem[Lin et~al.(2025)Lin, Guo, Han, Hu, Ni, Wang, Chen, Liu, Chen, He,
  Jiang, Jiao, Hu, and Wang]{lin2025seagentselfevolutiontrajectoryoptimization}
Jiaye Lin, Yifu Guo, Yuzhen Han, Sen Hu, Ziyi Ni, Licheng Wang, Mingguang Chen,
  Hongzhang Liu, Ronghao Chen, Yangfan He, Daxin Jiang, Binxing Jiao, Chen Hu,
  and Huacan Wang.
\newblock Se-agent: Self-evolution trajectory optimization in multi-step
  reasoning with llm-based agents, 2025.
\newblock URL \url{https://arxiv.org/abs/2508.02085}.

\bibitem[Liu et~al.(2025)Liu, Cai, Zhu, Zheng, Chen, Wen, Wang, E, and
  Chen]{liu2025mlmasteraiforaiintegrationexploration}
Zexi Liu, Yuzhu Cai, Xinyu Zhu, Yujie Zheng, Runkun Chen, Ying Wen, Yanfeng
  Wang, Weinan E, and Siheng Chen.
\newblock {ML-Master}: Towards {AI}-for-{AI} via integration of exploration and
  reasoning, 2025.
\newblock URL \url{https://arxiv.org/abs/2506.16499}.

\bibitem[Novikov et~al.(2025)Novikov, Vũ, Eisenberger, Dupont, Huang, Wagner,
  Shirobokov, Kozlovskii, Ruiz, Mehrabian, Kumar, See, Chaudhuri, Holland,
  Davies, Nowozin, Kohli, and Balog]{novikov_alphaevolve_2025}
Alexander Novikov, Ngân Vũ, Marvin Eisenberger, Emilien Dupont, Po-Sen Huang,
  Adam~Zsolt Wagner, Sergey Shirobokov, Borislav Kozlovskii, Francisco J.~R.
  Ruiz, Abbas Mehrabian, M.~Pawan Kumar, Abigail See, Swarat Chaudhuri, George
  Holland, Alex Davies, Sebastian Nowozin, Pushmeet Kohli, and Matej Balog.
\newblock {AlphaEvolve}: {A} coding agent for scientific and algorithmic
  discovery, June 2025.
\newblock URL \url{http://arxiv.org/abs/2506.13131}.
\newblock arXiv:2506.13131 [cs].

\bibitem[Olson et~al.(2016)Olson, Bartley, Urbanowicz, and
  Moore]{olson_evaluation_2016}
Randal~S. Olson, Nathan Bartley, Ryan~J. Urbanowicz, and Jason~H. Moore.
\newblock Evaluation of a {Tree}-based {Pipeline} {Optimization} {Tool} for
  {Automating} {Data} {Science}.
\newblock In \emph{Proceedings of the {Genetic} and {Evolutionary}
  {Computation} {Conference} 2016}, {GECCO} '16, pp.\  485--492, New York, NY,
  USA, July 2016. Association for Computing Machinery.
\newblock ISBN 978-1-4503-4206-3.
\newblock \doi{10.1145/2908812.2908918}.
\newblock URL \url{https://dl.acm.org/doi/10.1145/2908812.2908918}.

\bibitem[Romera-Paredes et~al.(2024)Romera-Paredes, Barekatain, Novikov, Balog,
  Kumar, Dupont, Ruiz, Ellenberg, Wang, Fawzi, Kohli, and
  Fawzi]{romera-paredes_mathematical_2024}
Bernardino Romera-Paredes, Mohammadamin Barekatain, Alexander Novikov, Matej
  Balog, M.~Pawan Kumar, Emilien Dupont, Francisco J.~R. Ruiz, Jordan~S.
  Ellenberg, Pengming Wang, Omar Fawzi, Pushmeet Kohli, and Alhussein Fawzi.
\newblock Mathematical discoveries from program search with large language
  models.
\newblock \emph{Nature}, 625\penalty0 (7995):\penalty0 468--475, January 2024.
\newblock ISSN 1476-4687.
\newblock \doi{10.1038/s41586-023-06924-6}.
\newblock URL \url{https://www.nature.com/articles/s41586-023-06924-6}.

\bibitem[Shang et~al.(2025)Shang, Li, Zhao, Ma, Liu, Xu, and
  Li]{shang2025agentsquareautomaticllmagent}
Yu~Shang, Yu~Li, Keyu Zhao, Likai Ma, Jiahe Liu, Fengli Xu, and Yong Li.
\newblock Agentsquare: Automatic llm agent search in modular design space,
  2025.
\newblock URL \url{https://arxiv.org/abs/2410.06153}.

\bibitem[Thornton et~al.(2013)Thornton, Hutter, Hoos, and
  Leyton-Brown]{thornton_auto-weka_2013}
Chris Thornton, Frank Hutter, Holger~H. Hoos, and Kevin Leyton-Brown.
\newblock Auto-{WEKA}: combined selection and hyperparameter optimization of
  classification algorithms.
\newblock In \emph{Proceedings of the 19th {ACM} {SIGKDD} international
  conference on {Knowledge} discovery and data mining}, {KDD} '13, pp.\
  847--855, New York, NY, USA, August 2013. Association for Computing
  Machinery.
\newblock ISBN 978-1-4503-2174-7.
\newblock \doi{10.1145/2487575.2487629}.
\newblock URL \url{https://dl.acm.org/doi/10.1145/2487575.2487629}.

\bibitem[Toledo et~al.(2025)Toledo, Hambardzumyan, Josifoski, Hazra, Baldwin,
  Audran-Reiss, Kuchnik, Magka, Jiang, Lupidi, Lupu, Raileanu, Niu, Shavrina,
  Gagnon-Audet, Shvartsman, Sodhani, Miller, Charnalia, Dunfield, Wu,
  Stenetorp, Cancedda, Foerster, and
  Bachrach]{toledo2025airesearchagentsmachine}
Edan Toledo, Karen Hambardzumyan, Martin Josifoski, Rishi Hazra, Nicolas
  Baldwin, Alexis Audran-Reiss, Michael Kuchnik, Despoina Magka, Minqi Jiang,
  Alisia~Maria Lupidi, Andrei Lupu, Roberta Raileanu, Kelvin Niu, Tatiana
  Shavrina, Jean-Christophe Gagnon-Audet, Michael Shvartsman, Shagun Sodhani,
  Alexander~H. Miller, Abhishek Charnalia, Derek Dunfield, Carole-Jean Wu,
  Pontus Stenetorp, Nicola Cancedda, Jakob~Nicolaus Foerster, and Yoram
  Bachrach.
\newblock {AI} research agents for machine learning: Search, exploration, and
  generalization in {MLE}-bench, 2025.
\newblock URL \url{https://arxiv.org/abs/2507.02554}.

\bibitem[Trirat et~al.(2025)Trirat, Jeong, and Hwang]{trirat_automl-agent_2025}
Patara Trirat, Wonyong Jeong, and Sung~Ju Hwang.
\newblock {AutoML}-{Agent}: {A} {Multi}-{Agent} {LLM} {Framework} for
  {Full}-{Pipeline} {AutoML}.
\newblock In \emph{Proceedings of the 42nd {International} {Conference} on
  {Machine} {Learning} ({ICML})}, 2025.
\newblock \doi{10.48550/arXiv.2410.02958}.
\newblock URL \url{http://arxiv.org/abs/2410.02958}.

\bibitem[Weng et~al.(2026)Weng, Antoniades, Nathani, Zhang, Pu, and
  Wang]{weng2026groupevolvingagentsopenendedselfimprovement}
Zhaotian Weng, Antonis Antoniades, Deepak Nathani, Zhen Zhang, Xiao Pu, and
  Xin~Eric Wang.
\newblock Group-evolving agents: Open-ended self-improvement via experience
  sharing, 2026.
\newblock URL \url{https://arxiv.org/abs/2602.04837}.

\bibitem[Yang et~al.(2025)Yang, Yang, Fang, Zhang, Wang, Xian, Li, Li, Xu, Li,
  Pan, Zhang, Liu, Shen, Chen, and
  Bian]{yang2025rdagentllmagentframeworkautonomous}
Xu~Yang, Xiao Yang, Shikai Fang, Yifei Zhang, Jian Wang, Bowen Xian, Qizheng
  Li, Jingyuan Li, Minrui Xu, Yuante Li, Haoran Pan, Yuge Zhang, Weiqing Liu,
  Yelong Shen, Weizhu Chen, and Jiang Bian.
\newblock {R\&D-Agent}: An {LLM}-agent framework towards autonomous data
  science, 2025.
\newblock URL \url{https://arxiv.org/abs/2505.14738}.

\bibitem[Yao et~al.(2023)Yao, Zhao, Yu, Du, Shafran, Narasimhan, and
  Cao]{yao_react_2023}
Shunyu Yao, Jeffrey Zhao, Dian Yu, Nan Du, Izhak Shafran, Karthik Narasimhan,
  and Yuan Cao.
\newblock {ReAct}: {Synergizing} {Reasoning} and {Acting} in {Language}
  {Models}.
\newblock In \emph{Proceedings of the 11th {International} {Conference} on
  {Learning} {Representations} ({ICLR})}, 2023.
\newblock URL \url{http://arxiv.org/abs/2210.03629}.

\bibitem[Ye et~al.(2024)Ye, Wang, Cao, Berto, Hua, Kim, Park, and
  Song]{ye_reevo_2024}
Haoran Ye, Jiarui Wang, Zhiguang Cao, Federico Berto, Chuanbo Hua, Haeyeon Kim,
  Jinkyoo Park, and Guojie Song.
\newblock {ReEvo}: {Large} {Language} {Models} as {Hyper}-{Heuristics} with
  {Reflective} {Evolution}.
\newblock In \emph{Advances in {Neural} {Information} {Processing} {Systems} 37
  ({NeurIPS})}, 2024.
\newblock \doi{10.48550/arXiv.2402.01145}.
\newblock URL \url{http://arxiv.org/abs/2402.01145}.

\bibitem[Yu \& Yang(2026)Yu and Yang]{yu2026evolvingensembleagents}
Zongmin Yu and Liu Yang.
\newblock Evolving ensemble of agents, 2026.
\newblock URL \url{https://arxiv.org/abs/2605.09018}.

\bibitem[Zhang et~al.(2025{\natexlab{a}})Zhang, Chen, Wan, Chang, Cheng, Wang,
  Hu, and Bai]{zhang2025evoflowevolvingdiverseagentic}
Guibin Zhang, Kaijie Chen, Guancheng Wan, Heng Chang, Hong Cheng, Kun Wang,
  Shuyue Hu, and Lei Bai.
\newblock Evoflow: Evolving diverse agentic workflows on the fly,
  2025{\natexlab{a}}.
\newblock URL \url{https://arxiv.org/abs/2502.07373}.

\bibitem[Zhang et~al.(2026)Zhang, Hu, Lu, Lange, and
  Clune]{zhang2026darwingodelmachineopenended}
Jenny Zhang, Shengran Hu, Cong Lu, Robert Lange, and Jeff Clune.
\newblock Darwin godel machine: Open-ended evolution of self-improving agents,
  2026.
\newblock URL \url{https://arxiv.org/abs/2505.22954}.

\bibitem[Zhang et~al.(2025{\natexlab{b}})Zhang, Xiang, Yu, Teng, Chen, Chen,
  Zhuge, Cheng, Hong, Wang, Zheng, Liu, Luo, and
  Wu]{zhang2025aflowautomatingagenticworkflow}
Jiayi Zhang, Jinyu Xiang, Zhaoyang Yu, Fengwei Teng, Xionghui Chen, Jiaqi Chen,
  Mingchen Zhuge, Xin Cheng, Sirui Hong, Jinlin Wang, Bingnan Zheng, Bang Liu,
  Yuyu Luo, and Chenglin Wu.
\newblock Aflow: Automating agentic workflow generation, 2025{\natexlab{b}}.
\newblock URL \url{https://arxiv.org/abs/2410.10762}.

\bibitem[Zhang et~al.(2024)Zhang, Zhang, Ren, Li, and
  Yang]{zhang_mlcopilot_2024}
Lei Zhang, Yuge Zhang, Kan Ren, Dongsheng Li, and Yuqing Yang.
\newblock {MLCopilot}: {Unleashing} the {Power} of {Large} {Language} {Models}
  in {Solving} {Machine} {Learning} {Tasks}.
\newblock In \emph{Proceedings of the 18th {Conference} of the {European}
  {Chapter} of the {Association} for {Computational} {Linguistics} ({EACL})},
  pp.\  2931--2959, 2024.
\newblock \doi{10.48550/arXiv.2304.14979}.
\newblock URL \url{http://arxiv.org/abs/2304.14979}.

\bibitem[Zhu et~al.(2026)Zhu, Cai, Liu, Zheng, Wang, Ye, Zhang, Zhang, E, Chen,
  and Wang]{zhu2026ultralonghorizonagenticsciencecognitive}
Xinyu Zhu, Yuzhu Cai, Zexi Liu, Bingyang Zheng, Cheng Wang, Rui Ye, Yuzhi
  Zhang, Linfeng Zhang, Weinan E, Siheng Chen, and Yanfeng Wang.
\newblock Toward ultra-long-horizon agentic science: Cognitive accumulation for
  machine learning engineering, 2026.
\newblock URL \url{https://arxiv.org/abs/2601.10402}.

\end{thebibliography}

\newpage
\appendix
\raggedbottom  
\section*{Appendix}
\addcontentsline{toc}{section}{Appendix}

\section{The Agent Harness}
\label{app:harness}

This appendix summarizes the general-purpose agent harness used to execute each autonomous run, its available tools, and its interface to externally supplied task instructions. It then describes how the agent's conversation develops within a run and how information is transferred between runs. The harness provides the execution infrastructure; Appendix~\ref{app:algorithm} describes the evolutionary orchestration that constructs agent seeds and selects the resulting solutions.

\subsection{Structure}\label{app:harness-structure}

The agent layer uses our general-purpose, goal-driven agent harness, which was not built or adapted specifically for tabular AutoML. Domain-specific instructions and evaluation commands are supplied through the workspace.

The harness is implemented as a LangGraph-based extended ReAct loop with planning, tool execution, durable memory, context compression, and completion checks. A persistent plan and memory record support execution across model turns, while checkpointing allows interrupted runs to resume. The reported configuration uses \agentgamodelname{} for all LLM roles.

The agent requests termination when it judges the task complete. Plan-completion checks and a final validation audit can return control to the agent before finalization. These mechanisms govern task execution rather than fitness: after the run, the evolution layer reads the recorded score from the workspace and applies the tournament and retry rules of Appendix~\ref{app:tournament-details}.

\subsection{Tool Suite}
\label{app:tools}

Table~\ref{tab:tools} lists the general-purpose tools available in the reported configuration. They support workspace inspection, file editing, command execution, visual inspection, and management of the agent's plan and memory. Domain-specific training and evaluation are accessed through the workspace's command-line interface rather than specialized harness tools; for example, a scored attempt is launched with \texttt{uv run python run.py main-training}. Web search, sub-agent dispatch, skills, and plugins are disabled in the reported runs.

\begin{table}[H]
\centering
\caption{Tools available to \method{} agents in the reported configuration.}
\label{tab:tools}
\vskip 0.1in
\small
\begin{tabular}{@{}p{1.3in}p{3.9in}@{}}
\toprule
\textbf{Category} & \textbf{Tools} \\
\midrule
File and shell & \texttt{read\_file}, \texttt{write\_file}, \texttt{edit\_file\_search\_replace}, \texttt{run\_shell} \\
Vision & \texttt{view\_image}, \texttt{file\_to\_image} \\
Plan & \texttt{check\_off\_plan\_items}, \texttt{expand\_plan}, \texttt{modify\_plan}, \texttt{revise\_task\_and\_plan} (for user redirects; not applicable to unattended runs) \\
Memory & \texttt{update\_memory} \\
Control & \texttt{exit\_session} \\
\bottomrule
\end{tabular}
\vskip -0.1in
\end{table}

\subsection{Task Input Interface}
\label{app:prompts}

Each run receives a task brief and a prepared workspace, following the familiar goal-directed interaction pattern of coding assistants such as Codex and Claude Code. The brief specifies what the agent should accomplish, while accompanying workspace instructions describe the available artifacts, permitted changes, and evaluation commands. The agent determines the sequence of analyses, code edits, and tool calls used to carry out the task.

In \method{}, the evolution layer constructs these starting conditions programmatically: it selects the task operator, supplies the brief and shared execution instructions, and makes the selected parent archives available in the workspace (Appendix~\ref{app:operator_behavior}). The domain package supplies the problem-specific instructions and evaluation interface (Appendix~\ref{app:domain-interface}). These externally supplied inputs are distinct from the harness's internal prompts for execution, planning, and memory. Agent-seed search therefore changes the starting conditions of each run rather than the internal design of the agent.

\subsection{The Conversation Chain}
\label{app:chain}

Within a run, the agent builds its own context. Like any chat-based agent, it works on a chain of atomic chunks, one per model turn, and a turn can contain a message, tool calls with their results, or both. Let $S$ be the system prompt and $A$ the first message. The first call sees $S+A$, the second sees these together with the first turn and its tool results, $S+A+B$, the third sees $S+A+B+C$, and so on, because every call re-reads the chain so far, so early turns shape all later ones (Figure~\ref{fig:chain-drift}). The chain therefore only grows, until it approaches the context limit. Schematically, the harness compresses the accumulated context $S+A+B+C+D$ into $S'$, which represents the retained instructions and compressed conversational state. The system prompt, task, and recent turns are preserved as described in Appendix~\ref{app:context}. The context shrinks, and the chain grows again from there, to $S'+E+F+G$ (Figure~\ref{fig:chain-schematic}).

What enters the chain is decided by the agent itself, namely which files it reads, which data it inspects, which experiments it runs, and which failures it records. The solution is not generated in a single step from a prompt but emerges from this chain. At some point a turn, say $G$, issues a tool call that writes or edits the solution files, the next turn $H$ runs the scoring command, and that code becomes the running solution, which later turns of the same chain continue to revise (Figure~\ref{fig:chain-example}). The chain ends when the agent itself judges that its task is complete and exits the run. \method{} never edits the solution a chain produces. It chooses only how each new chain begins, which is the agent seed of Section~\ref{sec:formulation}. What earlier chains learned reaches the child only as files under \texttt{Previous Experiments/}, which it opens when it chooses (Section~\ref{sec:lineage}).

This distinguishes \method{} from two alternatives. A single long-lived agent, as in autoresearch~\citep{karpathy_autoresearch_2026}, carries one chain, compressed whenever it approaches the limit, through the entire search. In our single-agent control, the search therefore ends whenever that one agent decides it is done (Section~\ref{sec:single}). Artifact-centric systems such as AIDE keep no such persistent chain. Each solution is produced from a prompt assembled by the scaffold, and the search operates on the resulting artifacts.

\begin{figure}[H]
\centering
\begin{tikzpicture}[
    tag/.style={draw, rounded corners=2pt, minimum width=0.62cm, minimum height=0.46cm, font=\small},
    body/.style={draw=black!25, rounded corners=2pt, anchor=north west, font=\scriptsize, align=left, text width=11.0cm, inner sep=3pt}]
    \newcommand{\msg}{\textcolor{blue!55!black}{\textsf{message}}}
    \newcommand{\call}{\textcolor{teal!60!black}{\textsf{tool call}}}
    \newcommand{\res}{\textcolor{black!55}{\textsf{result}}}
    \node[body] (b0) at (0.5,0) {system prompt: the harness rules, the tools and their arguments, and how to end the run};
    \node[tag, fill=black!12, anchor=east] (t0) at ([xshift=-0.12cm]b0.west) {$S$};
    \node[body] (b1) at ([yshift=-0.12cm]b0.south west) {first message: the task (problem, objective, editable and read-only paths, ``Run 3 attempts. For each one: \dots'')};
    \node[tag, fill=black!12, anchor=east] (t1) at ([xshift=-0.12cm]b1.west) {$A$};
    \draw[black!40] (t0.south) -- (t1.north);
    \node[body] (b2) at ([yshift=-0.12cm]b1.south west) {\msg\ ``First I will look at the data.''\\ \call\ \texttt{read\_file(Data/<competition>/train.csv)}\\ \res\ the columns and the first rows};
    \node[tag, fill=blue!8, anchor=east] (t2) at ([xshift=-0.12cm]b2.west) {$B$};
    \draw[black!40] (t1.south) -- (t2.north);
    \node[body] (b3) at ([yshift=-0.12cm]b2.south west) {\call\ \texttt{run\_shell(python eda.py)}\\ \res\ class balance, missing values, feature ranges};
    \node[tag, fill=blue!8, anchor=east] (t3) at ([xshift=-0.12cm]b3.west) {$C$};
    \draw[black!40] (t2.south) -- (t3.north);
    \node[body] (b4) at ([yshift=-0.12cm]b3.south west) {\msg\ ``Plan: reproduce the default model, then add features.''\\ \call\ \texttt{check\_off\_plan\_items(\dots)}};
    \node[tag, fill=blue!8, anchor=east] (t4) at ([xshift=-0.12cm]b4.west) {$D$};
    \draw[black!40] (t3.south) -- (t4.north);
    \node[body] (b5) at ([yshift=-0.12cm]b4.south west) {\call\ \texttt{run\_shell(uv run python run.py main-training --name baseline)}\\ \res\ development score recorded};
    \node[tag, fill=blue!8, anchor=east] (t5) at ([xshift=-0.12cm]b5.west) {$E$};
    \draw[black!40] (t4.south) -- (t5.north);
    \node[body] (b6) at ([yshift=-0.12cm]b5.south west) {\msg\ ``Two columns suggest a ratio feature.''\\ \call\ \texttt{edit\_file\_search\_replace(ml/solutions/fe\_independent/\dots)}};
    \node[tag, fill=blue!8, anchor=east] (t6) at ([xshift=-0.12cm]b6.west) {$F$};
    \draw[black!40] (t5.south) -- (t6.north);
    \node[body] (b7) at ([yshift=-0.12cm]b6.south west) {\call\ \texttt{write\_file(ml/solutions/models/\dots)}, which writes the solution that the next attempt scores};
    \node[tag, fill=green!20, anchor=east] (t7) at ([xshift=-0.12cm]b7.west) {$G$};
    \draw[black!40] (t6.south) -- (t7.north);
    \node[body] (b8) at ([yshift=-0.12cm]b7.south west) {\call\ \texttt{run\_shell(uv run python run.py main-training --name ratio\_features)}\\ \res\ development score recorded};
    \node[tag, fill=green!20, anchor=east] (t8) at ([xshift=-0.12cm]b8.west) {$H$};
    \draw[black!40] (t7.south) -- (t8.north);
    \node[anchor=north west, font=\scriptsize, text=black!65] at ([yshift=-0.12cm]b8.south west) {\dots\ and so on, until the agent judges that its task is complete and calls \texttt{exit\_session}};
\end{tikzpicture}
\caption{What the chunks of one chain contain, illustrated (not taken from a specific run). Each box is one model turn, with its message, its tool calls and their results; every call re-reads all boxes above it. Figure~\ref{fig:chain-schematic} shows how such a chain grows and is compressed.}
\label{fig:chain-example}
\end{figure}

\subsection{Context Compression}
\label{app:context}

Figure~\ref{fig:chain-schematic} illustrates the compression process described in Appendix~\ref{app:chain}: the chain grows until it approaches the context limit, is rewritten into a compact state $S'$, and then grows again. The remainder of this subsection specifies the compression rules.

\begin{figure}[H]
\centering
\begin{tikzpicture}[
    chunk/.style={draw, rounded corners=2pt, minimum width=0.85cm, minimum height=0.5cm, font=\small, fill=blue!8},
    sys/.style={chunk, fill=black!12},
    state/.style={chunk, fill=orange!25},
    act/.style={chunk, fill=green!20},
    lab/.style={font=\scriptsize, text=black!70, anchor=east},
    node distance=0.08cm]
    \node[lab] at (-0.2,0) {call 1};
    \node[sys, anchor=west] (a1) at (0,0) {$S$};
    \node[chunk, right=of a1] {$A$};
    \node[lab] at (-0.2,-0.7) {call 2};
    \node[sys, anchor=west] (b1) at (0,-0.7) {$S$};
    \node[chunk, right=of b1] (b2) {$A$};
    \node[chunk, right=of b2] {$B$};
    \node[lab] at (-0.2,-1.4) {call 3};
    \node[sys, anchor=west] (c1) at (0,-1.4) {$S$};
    \node[chunk, right=of c1] (c2) {$A$};
    \node[chunk, right=of c2] (c3) {$B$};
    \node[chunk, right=of c3] {$C$};
    \node[lab] at (-0.2,-2.1) {call 4};
    \node[sys, anchor=west] (d1) at (0,-2.1) {$S$};
    \node[chunk, right=of d1] (d2) {$A$};
    \node[chunk, right=of d2] (d3) {$B$};
    \node[chunk, right=of d3] (d4) {$C$};
    \node[chunk, right=of d4] (d5) {$D$};
    \draw[dashed, red!60] (4.75,-2.45) -- (4.75,0.35) node[above, font=\scriptsize, text=red!70] {context limit};
    \draw[-{Stealth}, thick, orange!80!black] (d1.south west) ++(0.42,-0.05) -- ++(0,-0.75)
        node[midway, right, font=\scriptsize, text=black!75] {compression: accumulated context becomes compact state $S'$};
    \node[lab] at (-0.2,-3.6) {call 5};
    \node[state, anchor=west] (e1) at (0,-3.6) {$S'$};
    \node[lab] at (-0.2,-4.3) {call 9};
    \node[state, anchor=west] (f1) at (0,-4.3) {$S'$};
    \node[chunk, right=of f1] (f2) {$E$};
    \node[chunk, right=of f2] (f3) {$F$};
    \node[act, right=of f3] (f4) {$G$};
    \node[act, right=of f4] (f5) {$H$};
    \node[font=\scriptsize, text=black!75] (lw) at ([yshift=-0.75cm]f4.south) {writes the solution};
    \draw[-{Stealth[scale=0.6]}, black!60] (lw.north) -- (f4.south);
    \node[font=\scriptsize, text=black!75, anchor=west] (lr) at ([xshift=0.9cm]f5.east) {runs the scoring command};
    \draw[-{Stealth[scale=0.6]}, black!60] (lr.west) -- (f5.east);
\end{tikzpicture}
\caption{How the chain grows and is compressed (schematic). Near the context limit the chain is rewritten into a compact state $S'$ and then grows again. In the harness itself, the system prompt, the task, and the most recent turns are kept verbatim.}
\label{fig:chain-schematic}
\end{figure}

Token counting uses \texttt{tiktoken} with \texttt{cl100k\_base} encoding. The unit of context management is a \emph{chunk}: one model turn together with the tool results that answer it, so that a tool call is never separated from its result. Under the harness defaults, compression is triggered when the prompt context exceeds 200{,}000 tokens or after every 100 agent steps, and it targets a context of 40{,}000 tokens. The first chunk (the task) and the last three chunks are protected. The remaining chunks are escalated, oldest first, through the following stages until the target is met:
\begin{enumerate}[leftmargin=*,topsep=0pt,itemsep=1pt]
    \item \textbf{Compress.} An LLM summarizes message content and reasoning to about 10\% of their length. Tool-call arguments and tool results are instead truncated to 200 tokens per value.
    \item \textbf{Truncate.} Every part of the chunk is clipped to a leading and a trailing span of tokens.
    \item \textbf{Drop.} The chunk is removed from the working context.
\end{enumerate}
A chunk moves to a new stage only if that stage shortens it. Before a new block of chunks is dropped, the memory subgraph rewrites \texttt{MEMORY.md} from the span that is about to leave the context, so that information in the dropped chunks survives in condensed form. The agent can also request this rewrite through the \texttt{update\_memory} tool. Compression calls run sequentially, and the full transcript is retained on disk.

\subsection{Path Dependence Within a Chain}
\label{app:chain-drift}

Because every call receives the retained working context, early turns shape all later ones. Figure~\ref{fig:chain-drift} illustrates the consequence: two chains that start from the same $S+A$ and differ only slightly in their first turns can end up searching in quite different directions, and a single chain tends to keep refining the direction it started in.

\begin{figure}[H]
\centering
\begin{tikzpicture}[
    chunk/.style={draw, rounded corners=2pt, minimum width=0.62cm, minimum height=0.42cm, font=\scriptsize, fill=blue!8, inner sep=1pt},
    sys/.style={chunk, fill=black!12},
    up/.style={chunk, fill=teal!18},
    dn/.style={chunk, fill=violet!15},
    ws/.style={chunk, fill=yellow!30},
    arch/.style={chunk, fill=orange!25},
    note/.style={font=\scriptsize, text=black!75, align=left},
    head/.style={font=\scriptsize\bfseries, anchor=west}]
        \node[sys] (s) at (0,0) {$S$};
    \node[sys, right=0.06cm of s] (a) {$A$};
    \node[up] (b1) at (1.55,0.3) {$B_1$};
    \node[up] (c1) at (2.45,0.62) {$C_1$};
    \node[up] (d1) at (3.35,0.98) {$D_1$};
    \node[up] (e1) at (4.25,1.4) {$E_1$};
    \node[dn] (b2) at (1.55,-0.3) {$B_2$};
    \node[dn] (c2) at (2.45,-0.62) {$C_2$};
    \node[dn] (d2) at (3.35,-0.98) {$D_2$};
    \node[dn] (e2) at (4.25,-1.4) {$E_2$};
    \foreach \x/\y in {a/b1,b1/c1,c1/d1,d1/e1,a/b2,b2/c2,c2/d2,d2/e2} \draw[-{Stealth[scale=0.6]}, black!55] (\x) -- (\y);
    \node[note, right=0.15cm of e1] {direction 1, e.g.\ tuning one\\gradient-boosting model};
    \node[note, right=0.15cm of e2] {direction 2, e.g.\ engineering\\new features};
    \node[note, anchor=north west] at (-0.4,-1.85) {$B_1$ and $B_2$ differ slightly (a different file read first,\\a different first model); every later turn builds on them};
\end{tikzpicture}
\caption{Why a chain drifts (schematic, not measured). A small difference in the first turns is carried into every later turn.}
\label{fig:chain-drift}
\end{figure}

\section{Evolutionary Orchestration}
\label{app:algorithm}

This appendix describes the evolution layer: how workspaces are passed between generations, the outer loop and its tournament, the operators, the \hedge{} allocation, and the configuration of the reported runs. Unless otherwise stated, concrete values refer to the experiment environment configuration reported in the manuscript rather than fallback library defaults.

The implementation intentionally separates stochastic generation from a fixed evaluation protocol. LLM calls are not seeded, so the same agent seed can produce divergent autonomous trajectories and solution artifacts. The evolution layer's random generator is not seeded either, so a run cannot be replayed from a random seed. Inherited archives are exposed as optional workspace resources rather than forced prompt context, so two runs from the same seed may draw different amounts from the workspace before diverging. The evolution layer never re-executes a solution. A member's fitness is the score that the fixed evaluation harness recorded when the agent ran the solution, namely the mean over \splitsperexperiment{} fixed seeds (42--46), each with its own 20\% holdout. The evaluation code is mounted read-only, and the protocol files are hashed before and after the run. A member whose protocol files changed is archived but cannot win a tournament.

\paragraph{Domain interface and evaluation records.}
\label{app:domain-interface}
The domain package supplies the workspace's \texttt{AGENTS.md}, specifying the evaluation command, editable components, archive-adoption procedure, and leakage constraints. The evaluation package and data are mounted read-only, except for the editable \texttt{ml/solutions/} directories. The agent may reorganize its solution within these directories; the evaluation interface remains fixed. Outcome-independent feature engineering is fitted across both frames of a split, whereas outcome-dependent feature engineering is fitted only on non-holdout rows. Each scored attempt records per-seed results, predictions, and a frozen solution. The evolution layer reads the score from \texttt{attempts.json}, which is writable and is not itself hashed; protocol-file hashes determine tournament eligibility. The prompts require scoring through the designated command, and the recorded artifacts permit subsequent auditing without re-executing the solution.

\subsection{Workspace Inheritance Across Generations}
\label{app:workspace-inheritance}

\method{} addresses the drift of Appendix~\ref{app:chain-drift} by requesting a small number of experiments per invocation and starting several fresh chains per generation (Figure~\ref{fig:workspace-inheritance}). Each chain works in a workspace. In generation 1, every chain starts from scratch, from the same template workspace $W_{0,0}$, and the chain of population slot $j$ edits it into $W_{1,j}$. In subsequent generations, each child starts in a fresh workspace containing the parent archives selected by its operator, and its chain produces $W_{g,j}$. A de novo \texttt{Initial} child receives no archives. The tournament keeps the best workspace per population slot, making it available for inheritance in the next generation. Each child's chain starts again from $S+A$ and reads any inherited archives as files when it chooses. The direction of the search is therefore re-drawn at every generation, while what earlier chains built is carried forward.

\begin{figure}[H]
\centering
\begin{tikzpicture}[
    chunk/.style={draw, rounded corners=2pt, minimum width=0.62cm, minimum height=0.42cm, font=\scriptsize, fill=blue!8, inner sep=1pt},
    sys/.style={chunk, fill=black!12},
    up/.style={chunk, fill=teal!18},
    dn/.style={chunk, fill=violet!15},
    ws/.style={chunk, fill=yellow!30},
    arch/.style={chunk, fill=orange!25},
    note/.style={font=\scriptsize, text=black!75, align=left},
    head/.style={font=\scriptsize\bfseries, anchor=west}]
        \foreach \r/\y/\st in {1/-3.65/up, 2/-4.4/chunk, 3/-5.15/dn} {
        \node[ws] (w0\r) at (0,\y) {$W_{0,0}$};
        \node[sys, right=0.3cm of w0\r] (s\r) {$S$};
        \node[sys, right=0.05cm of s\r] (a\r) {$A$};
        \node[\st, right=0.05cm of a\r] (b\r) {$B$};
        \node[\st, right=0.05cm of b\r] (c\r) {$C$};
        \node[ws, right=0.3cm of c\r] (w1\r) {$W_{1,\r}$};
        \draw[dashed, -{Stealth[scale=0.5]}, black!50] (w0\r) -- (s\r);
        \draw[-{Stealth[scale=0.55]}, black!60] (c\r) -- (w1\r);
        \node[ws] (k\r) at (6.75,\y) {$W_{1,\r}$};
        \node[sys, right=0.3cm of k\r] (t\r) {$S$};
        \node[sys, right=0.05cm of t\r] (aa\r) {$A$};
        \node[\st, right=0.05cm of aa\r] (bb\r) {$B$};
        \node[\st, right=0.05cm of bb\r] (cc\r) {$C$};
        \node[ws, right=0.3cm of cc\r] (w2\r) {$W_{2,\r}$};
        \draw[dashed, -{Stealth[scale=0.5]}, black!50] (k\r) -- (t\r);
        \draw[-{Stealth[scale=0.55]}, black!60] (cc\r) -- (w2\r);
    }
    \node[draw, rounded corners=2pt, fill=black!5, font=\scriptsize, align=center, minimum height=2.2cm, text width=1.25cm] (tour) at (5.3,-4.4) {tournament:\\best\\per slot\\kept};
    \foreach \r in {1,2,3} {
        \draw[-{Stealth[scale=0.55]}, black!55] (w1\r) -- (tour.west |- w1\r);
        \draw[-{Stealth[scale=0.55]}, orange!80!black, thick] (tour.east |- k\r) -- (k\r.west);
    }
    \node[font=\scriptsize, text=black!60, right=0.08cm of w22] {$\cdots$};
    \node[font=\scriptsize, text=black!70] at (2.1,-5.8) {generation 1};
    \node[font=\scriptsize, text=black!70] at (8.85,-5.8) {generation 2};
\end{tikzpicture}
\caption{How workspaces are inherited across generations (single-parent schematic). $W_{g,j}$ is the workspace produced by population slot $j$ in generation $g$. Generation 1 starts from the template workspace ($W_{0,0}$). Later children receive the selected parent's kept workspace as an archive in a fresh workspace; dashed arrows indicate access to the template or inherited archive, not continuation of the parent's conversation.}
\label{fig:workspace-inheritance}
\end{figure}

\subsection{Algorithm Summary}
\label{app:tournament-details}

Algorithm~\ref{alg:agentga_appendix} summarizes the outer-loop procedure used in the reported experiments.

\begin{algorithm}[t]
\caption{\method{} outer loop}
\label{alg:agentga_appendix}
\begin{algorithmic}
\STATE {\bfseries Input:} dataset $\mathcal{D}$, population size $n$, hard cap $T$, patience window $w$
\STATE Initialize elite pool $E \leftarrow \emptyset$ and bounded \hedge{} sampler from configured base probabilities
\STATE $\textit{best\_so\_far} \leftarrow \varnothing$
\STATE $\textit{stagnation\_count} \leftarrow 0$
\FOR{$t = 1$ {\bfseries to} $T$}
    \STATE $E^{-} \leftarrow E$ \COMMENT{previous elite snapshot}
    \FOR{$j = 1$ {\bfseries to} $n$ {\bfseries in parallel}}
        \IF{$t = 1$}
            \STATE $k_j \leftarrow \texttt{Initial}$, $p_j \leftarrow \varnothing$, $P_j \leftarrow \emptyset$
        \ELSE
            \STATE $k_j \leftarrow \textsc{HedgeSample}()$
            \STATE $(p_j, P_j) \leftarrow \textsc{SelectParents}(k_j, E^{-}, j)$
        \ENDIF
        \STATE $c_0^j \leftarrow (\text{task}_{k_j}, \text{archives}[P_j])$
        \STATE $(\pi_j, s_j) \leftarrow \textsc{Agent}(c_0^j, \mathcal{D})$
    \ENDFOR
    \FOR{$j = 1$ {\bfseries to} $n$}
        \IF{not $\textsc{Eligible}(j)$}
            \STATE Carry forward the reference elite if present; otherwise leave slot $j$ empty
        \ELSIF{$t = 1$ {\bfseries or} $\textsc{Better}(s_j, E^{-}[p_j].\text{score})$}
            \STATE $E[j] \leftarrow (\pi_j, s_j)$
        \ELSE
            \STATE $E[j] \leftarrow E^{-}[p_j]$
        \ENDIF
    \ENDFOR
    \IF{$t > 1$}
        \STATE Compute direction-aware improvements $\Delta_j$ for scored children against elite parents in $E^{-}$
        \STATE Update bounded \hedge{} weights from observed $(k_j, \Delta_j)$ pairs
    \ENDIF
    \STATE $b_t \leftarrow \textsc{Best}(E).\text{score}$
    \IF{$t = 1$ {\bfseries or} $\textsc{Better}(b_t, \textit{best\_so\_far})$}
        \STATE $\textit{best\_so\_far} \leftarrow b_t$
        \STATE $\textit{stagnation\_count} \leftarrow 0$
    \ELSE
        \STATE $\textit{stagnation\_count} \leftarrow \textit{stagnation\_count} + 1$
    \ENDIF
    \IF{$\textit{stagnation\_count} \geq w$}
        \STATE {\bfseries break}
    \ENDIF
\ENDFOR
\STATE {\bfseries Output:} $\textsc{Best}(E)$
\end{algorithmic}
\end{algorithm}

The pseudocode summarizes selection for iterations with an available elite; retry handling is described below. If no elite is available for the next iteration, the implementation launches only \texttt{Initial} members, and an eligible child can fill an empty slot without a comparison. \textsc{Eligible} requires a recorded score and an intact evaluation protocol. Here \textsc{Better} and \textsc{Best} apply the configured metric direction, $\textit{best\_so\_far}$ stores the best elite score seen so far, and $n=\agentgapopsize{}$, $T=\agentgamaxiterations{}$, and $w=\agentgapatience{}$ in the reported configuration. \textsc{Better} is a strict inequality with no tolerance, so a tie retains the incumbent.

\paragraph{Fitness, eligibility, and retries.} The score $s_j$ is the best value among all attempts that member $j$ recorded in its own workspace, including its reproduction of an inherited configuration. The attempts of its parents are not counted. A child is eligible for the tournament only if it recorded a score under an unmodified evaluation protocol. Otherwise, the reference elite is carried forward if present; an ineligible first-generation member leaves its slot empty. A member that records no score is relaunched as a new member with the same operator and parents, up to \agentgamaxretries{} times per slot and iteration. If the slot still has no score after these retries, the run terminates once the bookkeeping for that iteration is complete. An interrupted run resumes its members from their harness checkpoints. No separate pass re-validates a winner.

The \textsc{SelectParents} subroutine called in Algorithm~\ref{alg:agentga_appendix} is operationalized in Algorithm~\ref{alg:select_parents}. The previous-iteration elite snapshot $E^{-}$ used as input is the elite pool taken at the start of iteration $t$, before any of this iteration's children have been evaluated, matching the assignment $E^{-} \!\leftarrow\! E$ at the top of the loop in Algorithm~\ref{alg:agentga_appendix}. Thus $E^{-}[j]$ denotes slot $j$'s incumbent elite carried in from iteration $t \!-\! 1$.

\textsc{SelectParents} returns a pair $(p, P)$ that play distinct roles in the outer loop. The first component $p$ is the slot's \emph{tournament reference}: the incumbent elite that the new child must beat for promotion (the tournament step of Algorithm~\ref{alg:agentga_appendix}). For every operator, including \texttt{Initial}, $p \!=\! j$ is the slot's own elite in $E^{-}$, so a fresh \texttt{Initial} child sampled by \hedge{} in iteration $t \!>\! 1$ is still tournament-compared against that elite. The second component $P$ is the \emph{archive set copied into the child's workspace} as inheritable context; $P$ varies by operator and is summarized in Algorithm~\ref{alg:select_parents}. If slot $j$ holds no elite because its iteration-1 member was not eligible, a random elite from another slot serves as $p$. If the child loses, that elite is copied into slot $j$.

\begin{algorithm}[H]
\caption{\textsc{SelectParents} subroutine called in Algorithm~\ref{alg:agentga_appendix}}
\label{alg:select_parents}
\begin{algorithmic}
\STATE {\bfseries Input:} operator $k$, previous elite snapshot $E^{-}$, slot index $j$
\STATE $p \leftarrow j$ if slot $j$ holds an elite, otherwise a uniform random slot that does \COMMENT{tournament reference}
\IF{$k = \texttt{Initial}$}
    \STATE \textbf{return} $(p, \emptyset)$ \COMMENT{no archives in workspace; tournament still runs}
\ELSIF{$k \in \{\texttt{Ablation}, \texttt{EDA}, \texttt{Jumpstart}\}$}
    \STATE \textbf{return} $(p, \{p\})$
\ELSIF{$k = \texttt{Continue}$}
    \STATE $A \leftarrow$ sample $n_p - 1$ additional parents without replacement from $E^{-} \setminus \{p\}$
    \STATE \textbf{return} $(p, \{p\} \cup A)$ \COMMENT{$n_p \!=\! 1$ in the reported configuration}
\ELSIF{$k = \texttt{Merge}$}
    \IF{$E^{-} \setminus \{p\} = \emptyset$}
        \STATE \textbf{return} $(p, \{p\})$ \COMMENT{no other elite exists}
    \ENDIF
    \STATE $q \leftarrow$ uniform random from $E^{-} \setminus \{p\}$
    \STATE \textbf{return} $(p, \{p, q\})$
\ENDIF
\end{algorithmic}
\end{algorithm}

\subsection{Task Types and Operator Behavior}
\label{app:operator_behavior}
The code and its configuration keys refer to operators as \emph{tasks}, and the two terms are used interchangeably below.

\paragraph{Task and execution instructions.} The evolution layer constructs the task brief that forms the prompt component of each agent seed. It specifies the problem and objective, the selected operator, inherited archive locations, the requested experiment allowance, and the expected experimental records and outputs. The evolution layer also supplies shared execution instructions for using the designated scoring command, recording attempts, and committing experimental changes. These instructions are inputs to the general-purpose harness, not part of its internal prompting. Domain-specific commands and workspace constraints are supplied by the domain package (Appendix~\ref{app:domain-interface}).

Each member receives a single task prompt, which is composed for its operator from one fragment file (\path{agentga_ga/tasks/brief.md}). All operators share the header (problem, objective, direction, and ``Nothing else is measured.''), the list of inherited paths, the setup command, the editable and read-only paths, the reproduction rule, and the attempt loop (change, run the scoring command, and commit). Only the stance, any work before the loop, the first loop step, and the requested content of the closing \texttt{RESULTS.md} differ between operators. An operator specifies what a member inherits, not which modeling method it should use.

\paragraph{Inheritance.} The number of parents ($n_p$) varies by operator. Each child receives a complete copy of the elite workspace of every selected parent under \texttt{Previous Experiments/Iteration NNNN Population NNNN/}. The copy contains the parent's code, every attempt together with the frozen code that produced it, its notes, and its git history (as \texttt{parent.git}), plus a generated \texttt{PARENT.md}. Only virtual environments, caches, the read-only data mount, and the parent's own \texttt{Previous Experiments/} folder are excluded. Inheritance is therefore one generation deep, avoiding recursive copying of ancestral workspaces. The prompt merely lists these paths and asks the agent to read \texttt{PARENT.md} first. The archive contents are not injected into the context.

\paragraph{Attempt budget.} Each prompt requests \agentgaattemptsperagent{} scored attempts. Operators with parents first re-run the best inherited configuration once, and the prompt excludes this reproduction from the budget. \texttt{Initial} instead first runs the untouched defaults. The budget is a prompt instruction rather than an enforced limit. A member's fitness is the best score among all attempts it recorded, including the reproduction. The shared fragments read as follows, rendered for the reported configuration, where ``[...]'' marks omitted text:
\begin{verbatim}
Start from the best inherited configuration: install it, then run it
once to confirm it reproduces. Every measurement below is against that
reproduction.

**The reproduction does not count as one of your attempts.** It is the
starting line, not a result.
[...]
Run 3 attempts. For each one:
\end{verbatim}
For each operator, we quote the stance and the first loop step (``1.'').

\paragraph{Initial ($n_p = 0$).}
The member has no access to previous experiments and starts from the default pipeline.
\begin{verbatim}
No ancestors. Start from scratch.
[...]
Run the untouched defaults first, so you have a number to compare
against.
[...]
1. Make one change you can name.
\end{verbatim}

\paragraph{Ablation ($n_p = 1$).}
The member adds and removes single components of the elite parent's configuration.
\begin{verbatim}
Add or remove one component at a time, to find out what the score
depends on.
[...]
1. Add or remove exactly one component. Two changes in one attempt
   measures nothing.
\end{verbatim}

\paragraph{Merge (normally $n_p = 2$).}
The child sees the slot's elite parent and one other elite chosen at random, or only the slot's elite if no other elite is available. The intended base is the better-performing parent under the configured metric direction; the choice of what to transfer is left to the agent.
\begin{verbatim}
Combine components from two ancestors: build from one, borrow from the
other.
[...]
Diff the two ancestors and find where they differ. Take the
higher-scoring one as your base.
[...]
1. Bring across one component from the other ancestor.
\end{verbatim}

\paragraph{Continue ($n_p \geq 1$, configurable).}
The member refines the inherited configuration. The number of parents is configurable (default $n_p = 1$ in our experiments); the elite parent is always included, with additional parents sampled randomly from the elite pool.
\begin{verbatim}
Refine what already works.
[...]
1. Make one change you can name. Build on what worked.
\end{verbatim}

\paragraph{EDA ($n_p = 1$).}
The member first analyzes the data with disposable scripts and ties every change to a recorded observation.
\begin{verbatim}
Let the data decide.
[...]
Investigate before you change anything.
[...]
1. Make a change that follows from a specific observation, and name the
   observation in the notes.
[...]
Leave enough budget to act on what you find.
\end{verbatim}

\paragraph{Jumpstart ($n_p = 1$).}
The member uses the parent's artifacts as raw material for a substantially different solution. No external reference material is provided. The operator is implemented for completeness, but it was inactive in the reported experiments (\texttt{GA\_PROB\_JUMPSTART=0}).
\begin{verbatim}
Use everything the ancestor left you as raw material, and build
something that beats it.
\end{verbatim}

\subsection{\hedge{} Bounds Enforcement}

The bounded \hedge{} update has three stages: (i) compute per-operator rewards from parent-child gains, (ii) update weights in log space, and (iii) enforce floor/ceiling bounds with iterative rebalancing. The implementation aggregates parent-child gains by observed operator and computes the operator-level mean improvement $\bar{\Delta}_k$. If fewer than two operators are observed in an iteration, the \hedge{} update is skipped. Gains are measured against the pre-tournament elite scores for every child that recorded a score, including a child whose protocol files changed. A child without a score contributes no observation.

\paragraph{Rank-based rewards.}
Rewards are derived from ranks rather than from raw operator means:
\begin{equation}
    r_k = \frac{2 \cdot \text{rank}(k)}{|\mathcal{K}_{\text{obs}}| - 1} - 1
\end{equation}
where $\text{rank}(k) \in \{0, \ldots, |\mathcal{K}_{\text{obs}}|-1\}$ orders observed operators by $\bar{\Delta}_k$ (ties are broken by sort order). The worst operator receives $r=-1$ and the best receives $r=+1$, which reduces sensitivity to metric scale and outliers.

\paragraph{Clipped importance weighting.}
To avoid exploding updates for low-probability operators, we clip the importance factor:
\begin{equation}
    \tilde{r}_k = r_k \cdot \min\left(\frac{1}{p_k}, \kappa\right)
\end{equation}
where $p_k$ is the current bounded sampling probability and $\kappa = \hedgekappa{}$ in the reported configuration. Operators not sampled in an iteration receive no weight update.

\paragraph{Log-space updates.}
Weights are updated in log space for numerical stability:
\begin{equation}
    \log w_k^{(t+1)} = \log w_k^{(t)} + \eta \cdot \tilde{r}_k
\end{equation}
with learning rate $\eta = \hedgeeta{}$ in the reported configuration.

Iterative 2-pass algorithm (max 10 iterations):
\begin{enumerate}[leftmargin=*,topsep=0pt,itemsep=1pt]
    \item \textbf{Pass 1}: Enforce ceilings. Redistribute excess to below-ceiling tasks proportionally.
    \item \textbf{Pass 2}: Enforce floors. Take deficit from above-floor tasks proportionally.
\end{enumerate}
Repeat until stable (no changes in an iteration), then renormalize. These bounds operate on the post-softmax probabilities used both for task sampling and for the importance-weighting term in the \hedge{} update. Task types with base probability 0 are excluded from the active set before the bounds pass runs. Each slot samples its operator independently, so at most $n$ operators are observed per iteration. An iteration in which all slots draw the same operator therefore produces no update.

\subsection{Hyperparameters}
\label{app:hyperparam}

The reported runs share the following search hyperparameters and configured evaluation protocol. Execution-resource settings are distinguished from these algorithmic settings below.
\begin{itemize}[leftmargin=*,topsep=0pt,itemsep=1pt]
    \item Population size: $n=\agentgapopsize{}$. The members of an iteration run concurrently, each in its own container. Container resource settings are reported separately below.
    \item Continue parents: min${}=1$, max${}=1$
    \item Data: the dataset is bind-mounted read-only into every container from a pinned, checksummed local copy, and the framework downloads nothing at run time.
    \item Base task probabilities: Initial${}=0.1$, Continue${}=0.2$, Ablation${}=0.1$, Merge${}=0.1$, Jumpstart${}=0$, EDA${}=0.5$
    \item \hedge{} learning rate $\eta=\hedgeeta{}$, importance-weighting clip $\kappa=\hedgekappa{}$
    \item Configured \hedge{} floors: Continue${}=0.10$, Ablation${}=0.05$, Merge${}=0.05$, Initial${}=0.05$, EDA${}=0.05$
    \item Active \hedge{} tasks in these experiments: Initial, Continue, Ablation, Merge, EDA. Jumpstart is disabled by base probability 0 and therefore excluded from updates and sampling.
    \item \hedge{} ceiling: Merge${}=0.30$ (other active tasks uncapped)
    \item Stopping policy: \agentgamaxiterations{}-iteration cap, patience \agentgapatience{} (terminate when best-so-far elite score fails to strictly improve for \agentgapatience{} consecutive iterations)
    \item Attempt budget: \agentgaattemptsperagent{} requested attempts per agent, giving a nominal allowance of $\agentgapopsize{}\times\agentgamaxiterations{}\times\agentgaattemptsperagent{}=\singlerunattemptbudget{}$ counted attempts for a full run. Inheriting agents additionally reproduce the best inherited configuration without counting that reproduction against the allowance. The budget is a prompt instruction, not an enforced limit.
    \item Retries: a member without a score is relaunched up to \agentgamaxretries{} times per slot and iteration.
    \item Evaluation protocol: \splitsperexperiment{} seeds (42--46), each with a 20\% holdout (stratified for classification) and a further 20\% validation split for early stopping. The configured development score is the mean over these splits; actual model-fit counts are reported separately in Section~\ref{sec:compute}.
    \item Competitions: the problem type, metric, direction, and submission format come from a per-competition manifest. The search hyperparameters and configured split protocol above are shared across the \benchmarkcount{} competitions.
    \item Model: \agentgamodelname{}.
    \item Disabled capabilities: web tools, sub-agents, skills, and plugins (Appendix~\ref{app:tools}).
    \item Recorded container settings: the saved \texttt{runner\_memory} field is unset in \memorynolimitruns{} runs, \texttt{36g} in \memorythirtysixruns{}, and \texttt{64g} in \memorysixtyfourruns{}. These are recorded container configuration values, not measurements of host RAM or realized memory use; an unset field does not imply unlimited physical memory.
    \item Harness defaults: context compression as described in Appendix~\ref{app:context}, replanning every 50 agent responses, an LLM request timeout of 900\,s, and a shell timeout of 60\,s unless the agent passes a longer one. No wall-clock limit is set, and the default turn cap (about $10^{7}$) is not a practical limit.
    \item \single{}: the same model, agent harness, and configured evaluation protocol, with a population size of one, a single iteration, and a prompt that requests \singlerunattemptbudget{} attempts. With a single \texttt{Initial} member, there is no parent, no tournament, and no \hedge{} update.
    \item Always-Initial control: the configuration above with three changes: the base probability of Initial is 1 and that of every other task 0, \hedge{} is off, and the stopping policy is disabled, so that every run completes \agentgamaxiterations{} iterations (Appendix~\ref{app:always-initial}). On \initonlysplitcompetition{} its validation split also differs, as noted there.
\end{itemize}

\clearpage
\section{Positioning, Benchmark, and Resource Details}
\label{app:additional-results}

\subsection{Representative Positioning}
\label{app:additional-positioning}

Table~\ref{tab:comparison} places \method{} relative to representative prior systems by primary optimized unit and whether a higher-level orchestration algorithm coordinates multiple runs.

\begin{table}[H]
\centering
\caption{Representative positioning of \method{}. \textbf{Primary Unit} denotes what the outer system selects and reuses: an artifact, a bounded agentic execution, or an autonomous run. \textbf{Orchestrated} indicates whether a higher-level search coordinates multiple runs. \textbf{Agent-seed} indicates whether the optimized object is the initial state that launches a fresh autonomous run. (\cmark) marks a partial match: EvE places its guidance files and sampled solvers in the workspace that a session starts from, but the workspace is not the individual that is selected.}
\label{tab:comparison}
\small
\resizebox{\textwidth}{!}{%
\begin{tabular}{lccc}
\toprule
\textbf{Method} & \textbf{Primary Unit} & \textbf{Orchestrated} & \textbf{Agent-seed} \\
\midrule
EvoPrompt~\citep{guo_evoprompt_2024} & Artifact & \cmark & \xmark \\
ReEvo~\citep{ye_reevo_2024} & Artifact & \cmark & \xmark \\
FunSearch~\citep{romera-paredes_mathematical_2024} & Artifact & \cmark & \xmark \\
AIDE~\citep{jiang_aide_2025} & Artifact & \cmark & \xmark \\
AlphaEvolve~\citep{novikov_alphaevolve_2025} & Artifact & \cmark & \xmark \\
AFlow~\citep{zhang2025aflowautomatingagenticworkflow} & Artifact & \cmark & \xmark \\
AIRA-dojo~\citep{toledo2025airesearchagentsmachine} & Artifact & \cmark & \xmark \\
ML-Master~\citep{liu2025mlmasteraiforaiintegrationexploration} & Artifact & \cmark & \xmark \\
GEAR~\citep{jeddi2026geargeneticautoresearchagentic} & Artifact & \cmark & \xmark \\
EvE~\citep{yu2026evolvingensembleagents} & Artifact & \cmark & (\cmark) \\
\midrule
DS-Agent~\citep{guo_ds-agent_2024} & Agentic & \xmark & \xmark \\
Data Interpreter~\citep{hong_data_2025} & Agentic & \xmark & \xmark \\
AutoML-Agent~\citep{trirat_automl-agent_2025} & Agentic & \xmark & \xmark \\
AutoKaggle~\citep{li_autokaggle_2024} & Agentic & \xmark & \xmark \\
MLZero~\citep{fang_mlzero_2025} & Agentic & \xmark & \xmark \\
SELA~\citep{chi_sela_2024} & Agentic & \cmark & \xmark \\
I-MCTS~\citep{liang_i-mcts_2025} & Agentic & \cmark & \xmark \\
SE-Agent~\citep{lin2025seagentselfevolutiontrajectoryoptimization} & Agentic & \cmark & \xmark \\
R\&D-Agent~\citep{yang2025rdagentllmagentframeworkautonomous} & Agentic & \cmark & \xmark \\
\midrule
Agent K~\citep{grosnit_kolb-based_2025} & Autonomous & \xmark & \xmark \\
Darwin G\"odel Machine~\citep{zhang2026darwingodelmachineopenended} & Autonomous & \cmark & \xmark \\
Group-Evolving Agents~\citep{weng2026groupevolvingagentsopenendedselfimprovement} & Autonomous & \cmark & \xmark \\
\midrule
\textbf{\method{} (Ours)} & \textbf{Autonomous} & \cmark & \cmark \\
\bottomrule
\end{tabular}%
}
\end{table}

\subsection{Benchmark Context}
\label{app:benchmarks}

Table~\ref{tab:benchmarks} lists all \benchmarkcount{} Weco-Kaggle Lite competitions used in the reported experiments.

\begin{table}[H]
\centering
\caption{Weco-Kaggle Lite benchmark: 16 Kaggle tabular ML competitions from~\citet{jiang_aide_2025}.}
\label{tab:benchmarks}
\vskip 0.1in
\resizebox{\textwidth}{!}{%
\begin{tabular}{llc}
\toprule
\textbf{Competition} & \textbf{URL} & \textbf{Total Teams} \\
\midrule
playground-series-s3e14 & kaggle.com/c/playground-series-s3e14 & 1,877 \\
playground-series-s3e16 & kaggle.com/c/playground-series-s3e16 & 1,431 \\
playground-series-s3e19 & kaggle.com/c/playground-series-s3e19 & 1,174 \\
playground-series-s3e22 & kaggle.com/c/playground-series-s3e22 & 1,543 \\
playground-series-s3e24 & kaggle.com/c/playground-series-s3e24 & 1,910 \\
playground-series-s3e25 & kaggle.com/c/playground-series-s3e25 & 1,633 \\
tabular-playground-series-aug-2022 & kaggle.com/c/tabular-playground-series-aug-2022 & 1,889 \\
tabular-playground-series-feb-2021 & kaggle.com/c/tabular-playground-series-feb-2021 & 1,434 \\
tabular-playground-series-feb-2022 & kaggle.com/c/tabular-playground-series-feb-2022 & 1,257 \\
tabular-playground-series-jan-2022 & kaggle.com/c/tabular-playground-series-jan-2022 & 1,592 \\
tabular-playground-series-jul-2021 & kaggle.com/c/tabular-playground-series-jul-2021 & 1,294 \\
tmdb-box-office-prediction & kaggle.com/c/tmdb-box-office-prediction & 1,395 \\
bike-sharing-demand & kaggle.com/c/bike-sharing-demand & 3,243 \\
cat-in-the-dat & kaggle.com/c/cat-in-the-dat & 1,341 \\
house-prices-advanced-regression-techniques & kaggle.com/c/house-prices-advanced-regression-techniques & 4,978 \\
new-york-city-taxi-fare-prediction & kaggle.com/c/new-york-city-taxi-fare-prediction & 1,485 \\
\bottomrule
\end{tabular}%
}
\vskip -0.1in
\end{table}

\subsection{Function Evaluations per Competition}
\label{app:fevals}

Table~\ref{tab:fevals} reports, for each competition, the number of outer-loop iterations, agent invocations, and scored experiments, together with the tokens consumed. The last two columns give the always-Initial control at the full budget, which ran all \agentgamaxiterations{} iterations in every competition (\initonlyagentsfull{} invocations in total). All counts are read from the lineage and usage records of the runs rather than derived from the configuration. An invocation records a mean of \fevaltrainruns{} scored experiments (the \agentgaattemptsperagent{} requested attempts plus, for inheriting children, the reproduction of the parent), which gives a mean of \fevalevalsperiter{} scored experiments per iteration with \fevalpopsize{} population slots. Each scored experiment is evaluated with up to \splitsperexperiment{} models, one per split (a few recorded fewer); the refit on all labeled rows that produces the submission file is not included in this count. Iteration counts range from \fevalminiters{} to \fevalmaxiters{} (mean \fevalmeaniters{}), because a run stops either at the \agentgamaxiterations{}-iteration cap or after \agentgapatience{} iterations without improvement of the best development score, and scored experiments per competition range from \agentgaevalsmin{} to \agentgaevalsmax{}. Over all \benchmarkcount{} competitions, the \single{} control used \singleruntotalexperiments{} scored experiments and \$\singleruntotalcostusd{} of LLM calls.

\begin{table}[H]
\centering
\small
\caption{Per-competition compute resources for the reported \method{} runs and for the always-Initial control at the full budget, counted from the run records. Tokens include cached prompt tokens and are those recorded (Appendix~\ref{app:always-initial}). Cached is the share of prompt tokens served from the provider's cache, which are billed at a fraction of the normal price, so most of the token volume costs little.}
\label{tab:fevals}
\resizebox{\textwidth}{!}{
\begin{tabular}{lrrrrrrrr}
\toprule
 & \multicolumn{5}{c}{\method{}} & \multicolumn{3}{c}{Always-Initial, full budget} \\
\cmidrule(lr){2-6}\cmidrule(lr){7-9}
\textbf{Competition} & \textbf{Iterations} & \textbf{Invocations} & \textbf{Scored experiments} & \textbf{Tokens (M)} & \textbf{Cached (\%)} & \textbf{Scored experiments} & \textbf{Tokens (M)} & \textbf{Cached (\%)} \\
\midrule
Bike Sharing Demand & 20 & 60 & 250 & 258 & 82.9 & 199 & 114 & 81.0 \\
Cat in the Dat & 18 & 54 & 222 & 256 & 85.2 & 261 & 131 & 81.6 \\
House Prices & 11 & 33 & 159 & 199 & 78.5 & 474 & 164 & 84.0 \\
NYC Taxi Fare & 14 & 42 & 122 & 467 & 58.2 & 219 & 572 & 64.1 \\
Playground S3E14 & 20 & 60 & 240 & 315 & 79.5 & 230 & 194 & 80.5 \\
Playground S3E16 & 20 & 60 & 244 & 395 & 77.7 & 231 & 154 & 81.1 \\
Playground S3E19 & 20 & 60 & 253 & 365 & 80.5 & 236 & 247 & 84.2 \\
Playground S3E22 & 5 & 15 & 67 & 43 & 82.4 & 242 & 268 & 83.5 \\
Playground S3E24 & 20 & 60 & 244 & 318 & 74.7 & 239 & 230 & 84.6 \\
Playground S3E25 & 4 & 12 & 52 & 45 & 87.6 & 233 & 103 & 83.1 \\
TPS Aug 2022 & 20 & 60 & 244 & 371 & 84.6 & 255 & 244 & 81.5 \\
TPS Feb 2021 & 20 & 60 & 232 & 319 & 78.4 & 250 & 151 & 83.3 \\
TPS Feb 2022 & 20 & 60 & 224 & 669 & 51.6 & 217 & 499 & 67.1 \\
TPS Jan 2022 & 20 & 60 & 241 & 352 & 79.6 & 224 & 160 & 82.6 \\
TPS Jul 2021 & 20 & 60 & 249 & 662 & 53.1 & 245 & 118 & 81.0 \\
TMDB Box Office & 20 & 60 & 243 & 351 & 83.3 & 252 & 130 & 81.9 \\
\midrule
\textbf{Total} & \textbf{272} & \textbf{816} & \textbf{3,286} & \textbf{5,385} & \textbf{71.6} & \textbf{4,007} & \textbf{3,477} & \textbf{77.3} \\
\bottomrule
\end{tabular}
}
\end{table}

\paragraph{Hardware.} The \method{} compute reported in Table~\ref{tab:fevals} was distributed across 4 identical workstations, with 4 competitions per workstation run sequentially; in practice, approximately 3 of the 4 workstations were active concurrently rather than all 4. The always-Initial control ran on the same four workstations, also with 4 competitions on each. Within each active competition the GA dispatches one Docker container per population member per iteration (\agentgapopsize{} containers per iteration in the reported configuration), so each workstation hosts up to \agentgapopsize{} concurrent training containers. Each workstation has an AMD Ryzen Threadripper 7960X CPU (24 cores / 48 threads), 128\,GB DDR5-5600 ECC RAM, and 4\,TB NVMe SSD plus 16\,TB SATA HDD storage. No GPU is mapped into the workflow containers, so all training and scored evaluation in this work is CPU-only. Local hardware runs the Docker workflow image and the GA orchestrator; LLM inference for every model role is provider-hosted (\agentgamodelname{} via API) and is therefore not part of the local-compute footprint.

\subsection{Licenses for Existing Assets}
\label{app:licenses}

Table~\ref{tab:licenses} enumerates the existing assets used in this work, with the version or accession point and the license or terms under which we accessed them. We use each asset within the scope of its stated terms; no scraped data is redistributed, and no asset is re-released as part of this submission.

\begin{table}[H]
\centering
\small
\caption{Existing assets used in this paper, with versions and license terms. URLs and identifiers are provided where available; Kaggle competition data is governed by the per-competition rules accepted at download time.}
\label{tab:licenses}
\begin{tabular}{@{}p{0.30\textwidth} p{0.28\textwidth} p{0.34\textwidth}@{}}
\toprule
\textbf{Asset} & \textbf{Version / source} & \textbf{License / terms} \\
\midrule
Weco-Kaggle Lite benchmark~\citep{jiang_aide_2025} & WecoAI/aideml repository & MIT License \\
Kaggle competition datasets (\benchmarkcount{} tasks listed in Table~\ref{tab:benchmarks}) & Per-competition data downloads on kaggle.com & Per-competition rules and Kaggle Terms of Service; not redistributed \\
AIDE reference results~\citep{jiang_aide_2025} & WecoAI/aideml repository & MIT License (figures used as published baselines, with citation) \\
AIDE code used for the AIDE (DS) runs~\citep{jiang_aide_2025} & WecoAI/aideml repository, commit \texttt{7d2fc95} (v0.2.2 and later upstream fixes) & MIT License \\
AutoGluon~\citep{erickson_autogluon-tabular_2020} & \texttt{autogluon.tabular} 1.6.1 (autogluon/autogluon repository) & Apache License 2.0 \\
LLM used by \method{}, the always-Initial control, \single{} and AIDE (DS) (\agentgamodelname{}) & Provider-hosted API & Used under the respective provider Terms of Service; model weights are not redistributed \\
\bottomrule
\end{tabular}
\end{table}

\section{Extended Experimental Results}
\label{app:detail}
This appendix expands Section~\ref{sec:experiments}: the full leaderboard comparisons, the inheritance evidence, the dynamics of the search, the behavior of the \hedge{} controller, the single-agent control, competition by competition, and the always-Initial control. All numbers are computed from the archived run records by the same scripts as in the main text.

\subsection{Leaderboard Comparisons in Full}
\label{app:leaderboard}
Table~\ref{tab:pairwise} compares \method{} with every other arm. For each, it reports wins, losses, and ties by rank; the mean difference in \emph{Exceeds \% of Human} with a bootstrap confidence interval (10,000 resamples of the per-competition differences); a one-sided sign test on the wins; and the range of the mean difference when any single competition is omitted. The comparison with AIDE (DS) is model-matched. The comparison with the published AIDE (GPT-4T) results serves as a reference, and the comparison with \single{} is matched in allowance rather than in compute (Section~\ref{sec:single}). The always-Initial control shares \method{}'s model, harness, and attempt allowance (Section~\ref{sec:always-initial}). Among the five arms we ran, with the always-Initial control under \method{}'s stopping rule, \method{} is best on \fivearmsagentgabest{} competitions (AutoGluon on \fivearmsautogluonbest{}, the control on \fivearmscontrolbest{}, \single{} on \fivearmssinglebest{}), and it reaches the top decile of the human leaderboard on \agentgatopdecile{} competitions, compared with \initonlytopdecilestop{} for the control, \aidetopdecile{} for AIDE (DS), \autogluontopdecile{} for AutoGluon, and \singleruntopdecile{} for \single{}. With the control at the full budget, the counts of best results are \fivearmsagentgabestfull{}, \fivearmsautogluonbestfull{}, \fivearmscontrolbestfull{}, and \fivearmssinglebestfull{}, and the control reaches the top decile on \initonlytopdecilefull{}. Appendix~\ref{app:always-initial} examines the control in depth.

\begin{table}[H]
\centering
\small
\caption{\method{} compared with each other arm over the \benchmarkcount{} competitions. The leave-one-out range gives the lowest and highest mean difference obtained when any one competition is omitted.}
\label{tab:pairwise}
\resizebox{\textwidth}{!}{
\begin{tabular}{lrrrrrr}
\toprule
\method{} against & Wins & Losses & Ties & Mean $\Delta$ Exceeds [95\% CI] & Sign test $p$ (one-sided) & Leave-one-out range \\
\midrule
\single{} & 12 & 3 & 1 & +9.93 [-0.97, 22.36] & 0.0176 & +5.10 to +13.39 \\
Always-Initial, same stopping rule & 11 & 4 & 1 & +5.37 [-0.80, 12.29] & 0.0592 & +2.86 to +6.80 \\
Always-Initial, full budget & 11 & 5 & 0 & +3.88 [-2.11, 9.79] & 0.1051 & +2.23 to +5.53 \\
AIDE (DS) & 14 & 2 & 0 & +13.52 [7.58, 20.04] & 0.0021 & +11.60 to +14.68 \\
AutoGluon & 11 & 5 & 0 & +13.24 [2.74, 26.05] & 0.1051 & +8.51 to +15.09 \\
AIDE (GPT-4T) & 14 & 1 & 1 & +19.75 [12.44, 27.69] & 0.0005 & n/a \\
\bottomrule
\end{tabular}
}
\end{table}

Table~\ref{tab:published-reference} places \method{}, the always-Initial control, \single{}, AIDE (DS), and AutoGluon beside every system whose mean the AIDE authors report on the same benchmark. The published numbers were obtained with a different model, harness, and date, so they are a reference rather than a controlled comparison.

\begin{table}[H]
\centering
\small
\caption{Mean \emph{Exceeds \% of Human} over the \benchmarkcount{} competitions of Weco-Kaggle Lite: the arms we ran, and the systems reported by the AIDE authors (their Table~1).}
\label{tab:published-reference}
\begin{tabular}{llr}
\toprule
System & Model & Mean \emph{Exceeds \% of Human} \\
\midrule
\multicolumn{3}{l}{\emph{Run by us (this paper)}} \\
\method{} & \agentgamodelname{} & 71.13 \\
Always-Initial, full budget & \agentgamodelname{} & 67.25 \\
Always-Initial, same stopping rule & \agentgamodelname{} & 65.76 \\
\single{} & \agentgamodelname{} & 61.20 \\
AIDE (DS) & \agentgamodelname{} & 57.61 \\
AutoGluon & N/A & 57.89 \\
\midrule
\multicolumn{3}{l}{\emph{Published by the AIDE authors~\citep{jiang_aide_2025}, not run by us}} \\
AIDE & GPT-4 Turbo & 51.38 \\
Human + ChatGPT & GPT-4 Turbo & 41.17 \\
H2O AutoML & N/A & 35.34 \\
AutoGPT (LangChain) & GPT-4 Turbo & 32.34 \\
\bottomrule
\end{tabular}

\end{table}

\subsection{Inheritance Evidence in Depth}
\label{app:phases}
Section~\ref{sec:lineage} compares children that inherit a parent archive with \textbf{Initial} children that start from scratch. Table~\ref{tab:operators} breaks the \lineagepaircount{} decided tournaments down by operator, and Figure~\ref{fig:phases} by phase of the search. Three descriptive comparisons supplement the main contrast. First, within the \sameiteriterations{} iterations in which the same run launched both kinds of child against their respective slot elites, \textbf{Initial} children won \sameiterinitialwinrate\% of their tournaments and inheriting children \sameiterparentedwinrate\%, so the gap also appears within this restricted comparison. Second, against a fixed bar, the best development score of iteration 1, only \laterinitialbeatfirstpct\% of later \textbf{Initial} proposals clear it, compared with \inheritingbeatfirstpct\% of inheriting children. Third, we compare tie and regression rates. An inheriting child ties with the parent in \childtiedpct\% of tournaments and ends worse in only \childworsepct\%, whereas \initialworsepct\% of \textbf{Initial} children end worse than the elite they challenge. The remaining \childbetterlost{} inheriting children scored better than their parent without winning: \childbetterlostprotocol{} had changed the scoring protocol and were not eligible, and \childbetterlostotherslot{} had inherited from another slot's elite and did not beat the elite of its own slot.

\begin{table}[H]
\centering
\caption{Parent-child tournaments by operator over the \benchmarkcount{} reported runs (iteration 2 onward). A win is a strict improvement over the elite of the same population slot. Median gain is the relative change of the development score against that elite, signed so that positive is better and divided by the larger of the two scores in absolute value; every relative gain of a development score in this paper is computed in this way.}
\label{tab:operators}
\resizebox{\textwidth}{!}{
\begin{tabular}{lcrrcrr}
\toprule
Operator & Parents & Tournaments & Wins & Win rate \% [95\% CI] & \shortstack[r]{Median\\gain \%} & \shortstack[r]{Median gain \%\\when won} \\
\midrule
\textbf{Initial} & 0 & 45 & 8 & 17.8 [9.3, 31.3] & -0.939 & +0.399 \\
\textbf{Continue} & $\geq$1 & 138 & 93 & 67.4 [59.2, 74.6] & +0.028 & +0.097 \\
\textbf{Ablation} & 1 & 62 & 29 & 46.8 [34.9, 59.0] & +0.000 & +0.032 \\
\textbf{Merge} & 2 & 66 & 51 & 77.3 [65.8, 85.7] & +0.025 & +0.069 \\
\textbf{EDA} & 1 & 448 & 332 & 74.1 [69.9, 77.9] & +0.024 & +0.076 \\
\midrule
All inheriting operators & $\geq$1 & 714 & 505 & 70.7 [67.3, 73.9] & +0.020 & +0.077 \\
\bottomrule
\end{tabular}
}
\end{table}

\begin{figure}[H]
\centering
\begin{tikzpicture}[x=1.15cm,y=0.025cm,font=\scriptsize]
\definecolor{seriesA}{HTML}{2A78D6}\definecolor{seriesB}{HTML}{EB6834}
\draw[black!12] (0,0) -- (5.70,0); \node[left,black!60] at (0,0) {0};
\draw[black!12] (0,25) -- (5.70,25); \node[left,black!60] at (0,25) {25};
\draw[black!12] (0,50) -- (5.70,50); \node[left,black!60] at (0,50) {50};
\draw[black!12] (0,75) -- (5.70,75); \node[left,black!60] at (0,75) {75};
\draw[black!12] (0,100) -- (5.70,100); \node[left,black!60] at (0,100) {100};
\node[rotate=90,black!70] at (-0.75,50) {Tournaments won (\%)};
\node[below,black!70] at (0.60,-5) {2--5};
\node[below,black!70] at (2.10,-5) {6--10};
\node[below,black!70] at (3.60,-5) {11--15};
\node[below,black!70] at (5.10,-5) {16--20};
\node[below,black!70] at (2.85,-24) {Iterations};
\draw[seriesA,line width=1.2pt] (0.60,77.5) -- (2.10,70.9) -- (3.60,68.3) -- (5.10,66.0);
\fill[seriesA] (0.60,77.5) circle (1.8pt); \node[above,black!75] at (0.60,78.5) {134/173};
\fill[seriesA] (2.10,70.9) circle (1.8pt); \node[above,black!75] at (2.10,71.9) {141/199};
\fill[seriesA] (3.60,68.3) circle (1.8pt); \node[above,black!75] at (3.60,69.3) {123/180};
\fill[seriesA] (5.10,66.0) circle (1.8pt); \node[above,black!75] at (5.10,67.0) {107/162};
\node[right,seriesA!80!black] at (5.45,66.0) {Inherits a parent archive};
\draw[seriesB,line width=1.2pt] (0.60,37.5) -- (2.10,12.5) -- (3.60,8.3) -- (5.10,0.0);
\fill[seriesB] (0.60,37.5) circle (1.8pt); \node[above,black!75] at (0.60,38.5) {6/16};
\fill[seriesB] (2.10,12.5) circle (1.8pt); \node[above,black!75] at (2.10,13.5) {1/8};
\fill[seriesB] (3.60,8.3) circle (1.8pt); \node[above,black!75] at (3.60,9.3) {1/12};
\fill[seriesB] (5.10,0.0) circle (1.8pt); \node[above,black!75] at (5.10,1.0) {0/9};
\node[right,seriesB!80!black] at (5.45,0.0) {\textbf{Initial} (from scratch)};
\end{tikzpicture}
\caption{Tournament win rate by phase of the search. De novo \textbf{Initial} proposals lose ground as elites improve, whereas children that inherit a parent archive continue to win about two tournaments in three.}
\label{fig:phases}
\end{figure}

\paragraph{Detected archive accesses.} The seed only makes the archive available; the agents' tool logs show how it is used. All \inheritingran{} inheriting agents that ran had detected accesses to the inherited archive (\inheritingfailed{} further launches failed before their first model call), with a median of \inheritedtouchesmedian{} tool calls (\inheritedtouchsharepct\% of their calls), the first of which occurs \inheritedfirsttouchpct\% of the way into the run. Table~\ref{tab:opens} lists the accessed categories: all had calls accessing the parent's summary and recorded attempts, most also accessed solution code, and fewer accessed version history. Table~\ref{tab:archive-growth} shows that detected access counts remain approximately stable as archives grow with lineage depth. Between iteration 2 and iterations 16--20, the median inherited archive grows from \archivefilesearly{} to \archivefileslate{} text files (from \archivemegabytesearlytext{} to \archivemegabyteslatetext{}\,MB of text, and from \archivemegabytesearly{} to \archivemegabyteslate{}\,MB in total), yet the number of calls that access it stays nearly constant, and the median peak context rises only from \contextearly{} to \contextlate{} tokens against a limit of \contextlimit{} tokens.

\paragraph{How access is measured.} The measurement uses each agent's event log, which records every tool call together with its arguments. A tool call counts as opening the inherited archive when its arguments contain the path \texttt{Previous Experiments/}, whatever the tool: reading or listing a file, or a shell command that copies or runs something from the folder. The path that follows the parent's folder decides what was opened, with the rules applied in this order. \texttt{PARENT.md} is the parent's summary. Other Markdown files and notes count as notes and documentation, paths under \texttt{attempts/} as recorded attempts, Python files and the solution folders as solution code, and \texttt{.git} as git history, as does any \texttt{git} shell command that names the folder. JSON, CSV, and other result files count as result files and artifacts, and anything else, such as listing a folder, as a folder listing. A shell command can also change into a parent's folder and name a file afterwards, so for shell commands that mention the folder the whole command is scanned for the same names, and \texttt{run.py adopt}, which installs the frozen code of the parent's best recorded attempt, counts as accessing the recorded attempts. A call that names several paths can count in several categories, so a file inside a recorded attempt's frozen \texttt{solutions/} counts both as a recorded attempt and as solution code. The position of the first touch is the index of the first such call divided by the agent's total number of tool calls. Uses of files after an agent has copied them out of the folder are not counted, and the proxy does not directly measure which contents the model read.

\begin{table}[H]
\centering
\small
\caption{Parts of the inherited archive accessed by inheriting children, over the \inheritingran{} inheriting agents that ran.}
\label{tab:opens}
\begin{tabular}{lrr}
\toprule
What the child opens in the inherited archive & Invocations touching it (\%) & Median calls \\
\midrule
the parent's summary (\texttt{PARENT.md}) & 100.0 & 2 \\
notes and documentation & 98.5 & 2 \\
solution code & 91.8 & 5 \\
recorded attempts and their results & 100.0 & 10 \\
git history & 40.7 & 0 \\
result files and artifacts & 59.4 & 1 \\
folder listings & 99.6 & 9 \\
\bottomrule
\end{tabular}

\end{table}

\begin{table}[H]
\centering
\small
\caption{Size of the inherited archive and the number of detected accesses by each child, by iteration of the search (medians over all inheriting invocations with event logs, including those whose child recorded no score, which is why the counts slightly exceed the number of decided tournaments).}
\label{tab:archive-growth}
\resizebox{\textwidth}{!}{
\begin{tabular}{lrrrrrr}
\toprule
Iterations & Invocations & Text files & Text (MB) & Total (MB) & Calls touching it & Peak context (tokens) \\
\midrule
2 & 41 & 105 & 0.33 & 3.2 & 21 & 59,845 \\
3--5 & 132 & 128 & 0.40 & 5.3 & 23 & 60,770 \\
6--10 & 199 & 182 & 0.57 & 12.7 & 25 & 74,378 \\
11--15 & 183 & 240 & 0.83 & 23.2 & 26 & 89,633 \\
16--20 & 162 & 308 & 1.41 & 41.0 & 26 & 90,748 \\
\bottomrule
\end{tabular}
}
\end{table}

\subsection{Search Dynamics}
\label{app:dynamics}
The best development score improved after iteration 1 in \runsimprovedafterfirst{} of the \benchmarkcount{} runs, and \iterationsimprovedpct\% of the iterations after the first improved the global best. The final best elite lies a median of \winnerdepthmedian{} inheriting generations deep (at most \maxwinnerdepth{}), so the reported solutions are the product of long lineages rather than of a single fortunate agent. Agents explore rather than climb monotonically. Of \newexperimentscount{} new scored experiments (excluding reproductions of the parent), \experimentsbetterpct\% beat the inherited parent, \experimentsworsepct\% were worse, and the remaining \experimentsequalparent{} experiments matched it exactly. \invocationsmixedpct\% of invocations contain both improvements and regressions, \invocationsnonebeat{} invocations produced no experiment that beat the parent, and \invocationsallbeat{} produced only improvements. Each agent keeps its best experiment and the tournament discards the rest, which is why so few inheriting children end worse than their parent.

\paragraph{Parallel lineages.} The 1:1 tournament keeps \agentgapopsize{} lineages alive instead of truncating to the global best. Two or more population slots held the lead at some point in \leadruns{} of the \benchmarkcount{} runs, and the lead changed between slots in \leadchangeruns{} runs (median \leadchangesmedian{} changes). In \trailingatmidpoint{} of the \longruns{} runs with at least 10 iterations, the slot that finished best was not leading at the midpoint. The three final elites are distinct solutions in \finalelitesdiffer{} runs, with a median gap of \slotgapmedian\% between the best and the worst slot. Lineages also exchange material through Merge. The final best elite descends from two or more slots in \crossslotancestryruns{} runs (from all three in \crossslotancestryall{}), its ancestry contains a Merge in \mergeinancestryruns{} runs, and it has a median of \ancestorsmedian{} ancestor invocations, \ancestorsneverbestpct\% of which were never the global best themselves. Intermediate solutions that were never submitted thus shaped the one that was. Patience matters for the same reason: two-iteration plateaus were later broken \plateausbroken{} times, and a patience of 2 instead of \agentgapatience{} would have ended \plateauruns{} runs prematurely.

\subsection{Operator Allocation}
\label{app:hedge-detail}
Changes in \hedge{} probabilities are positively associated with observed operator win rates in most runs. In \hedgepositiveruns{} of the \hedgerunscompared{} runs in which two or more operators can be compared, the change in an operator's probability correlates positively with its realized win rate (median Spearman $\rho=\hedgespearmanmedian{}$). Table~\ref{tab:hedge} gives the prior and final probability of each operator together with its pooled tournament win rate. \textbf{EDA} rises from its prior of \hedgeedaprior{} to a mean final probability of \hedgeedafinal{}, while \textbf{Initial} falls from \hedgeinitialprior{} to \hedgeinitialfinal{}. Reweighting pooled observed operator win rates by the realized and prior mixes gives \hedgeexpectedrealised\% and \hedgeexpectedprior\%, respectively. This descriptive calculation does not estimate the effect of running a fixed-prior policy. The usefulness of operators does change over the search (Figure~\ref{fig:phases} shows this for the two kinds of child), which is the setting the bounded update was designed for. Because we did not run a fixed-prior control, however, the marginal value of adaptive allocation is not isolated.

\begin{table}[H]
\centering
\small
\caption{\hedge{} probabilities per operator: the configured prior and the probability at the end of the run (mean, minimum, and maximum over the \benchmarkcount{} runs), together with the operator's pooled tournament win rate.}
\label{tab:hedge}
\begin{tabular}{lrrrrr}
\toprule
Operator & Prior & Mean final & Min final & Max final & Pooled win rate (\%) \\
\midrule
Initial & 0.10 & 0.06 & 0.05 & 0.18 & 17.8 \\
Continue & 0.20 & 0.16 & 0.10 & 0.48 & 67.4 \\
Ablation & 0.10 & 0.07 & 0.05 & 0.12 & 46.8 \\
Merge & 0.10 & 0.09 & 0.05 & 0.30 & 77.3 \\
EDA & 0.50 & 0.62 & 0.25 & 0.75 & 74.1 \\
\bottomrule
\end{tabular}

\end{table}

\subsection{The Single-Agent Control in Depth}
\label{app:single}
\single{} is not the genetic algorithm: with one population slot and one iteration there is no tournament, restart, or inheritance. It is a single autonomous run in our harness, our counterpart of single-lineage solvers such as autoresearch (Section~\ref{sec:related}). Table~\ref{tab:single-run} expands the matched-allowance single-agent comparison in Section~\ref{sec:single}, listing the experiments, stopping behavior, and development-score comparison for every competition. \singlerunstoppedearly{} of the \benchmarkcount{} agents stopped of their own accord before using 90\% of the allowance, after a median of \singlerunearlymedian{} scored experiments (\singlerunallowancepct\% of the allowance), and a median of \singlerunafterlastgain{} experiments after their last improvement. \singlerunstillimproving{} of them set their best score in the last fifth of their experiments. They did not overlook the instruction. All \singlerunstoppedearly{} refer to the allowance while planning, \singlerunnotedunspent{} note the unspent allowance in their own final audit, \singlerunsaidconverged{} of these attribute it to convergence, a performance ceiling, or noise, and \singlerunmentionplateau{} mention a plateau or optimum in their final message. Among the competitions of the early-stopping agents, the two arms share a scoring protocol in \gabeyondcompared{}; in \gabeyondbetter{} of these, \method{} finished with a better development score than the one at which the single agent stopped (last column of the table).

\begin{table}[H]
\centering
\scriptsize
\caption{The single-agent control, competition by competition: the number of scored experiments and their share of the \singlerunattemptbudget{}-attempt allowance; the position in the sequence of the last improvement; the context size at the stop; the relative margin by which \method{}'s final development score exceeds the single agent's best (positive is better); and the agent's own stated reason for stopping, as classified from its final audit.}
\label{tab:single-run}
\resizebox{\textwidth}{!}{
\begin{tabular}{lrrrrrl}
\toprule
Competition & Scored exp. & Share (\%) & Last gain at (\%) & Context at stop & \method{} beyond (\%) & Own account of stopping \\
\midrule
Bike Sharing Demand & 30 & 16.7 & 79 & 101,893 & +78.14 & ceiling reached, converged, within noise \\
Cat in the Dat & 27 & 15.0 & 87 & 123,539 & +0.25 & not stated \\
House Prices & 182 & 101.1 & 99 & 137,766 & -0.98 & converged \\
NYC Taxi Fare & 18 & 10.0 & 95 & 109,512 & -4.03 & converged \\
Playground S3E14 & 21 & 11.7 & 96 & 117,612 & +0.87 & within noise, plateau \\
Playground S3E16 & 30 & 16.7 & 97 & 104,181 & +1.07 & diminishing returns, plateau \\
Playground S3E19 & 188 & 104.4 & 100 & 124,492 & +4.21 & plateau \\
Playground S3E22 & 26 & 14.4 & 21 & 70,608 & +0.00 & plateau, saturated \\
Playground S3E24 & 29 & 16.1 & 81 & 95,387 & +0.85 & converged, within noise, plateau \\
Playground S3E25 & 21 & 11.7 & 77 & 74,977 & +0.00 & not stated \\
TPS Aug 2022 & 19 & 10.6 & 96 & 198,042 & +1.34 & not stated \\
TPS Feb 2021 & 14 & 7.8 & 100 & 78,030 & +0.16 & converged, within noise, plateau \\
TPS Feb 2022 & 16 & 8.9 & 89 & 137,229 & +0.28 & diminishing returns, within noise \\
TPS Jan 2022 & 31 & 17.2 & 97 & 151,857 & +1.87 & converged, within noise \\
TPS Jul 2021 & 18 & 10.0 & 90 & 126,509 & +37.08 & converged, plateau \\
TMDB Box Office & 24 & 13.3 & 85 & 143,823 & +4.68 & diminishing returns \\
\bottomrule
\end{tabular}
}
\end{table}

\paragraph{Equal budgets.} The single-agent control is matched in allowance, not in compute, so we also compare the two arms at equal budget. Each \method{} run is cut at the last agent, in iteration and population order, whose cumulative spend stays within what the single agent used on that competition, and the best elite of the truncated run is compared with the single agent's final development score. The comparison covers the \equalbudgetcompared{} competitions in which both arms assign the untouched default pipeline the same development score, so that their scores are directly comparable (all but \equalbudgetexcluded{}). This criterion is stricter than the shared scoring protocol used in the comparison above. Table~\ref{tab:equal-budget} shows that \method{}'s advantage is not efficiency per experiment. Cut at the single agent's scored experiments, it is better on \equalbudgetexpbetter{} competitions and worse on \equalbudgetexpworse{}, and cut at its tokens, it is better on \equalbudgettokbetter{} and worse on \equalbudgettokworse{}. \method{} first matches the single agent's final score on \equalbudgetreached{} of the \equalbudgetcompared{} competitions, after a median of \equalbudgetreachexp$\times$ its scored experiments and \equalbudgetreachtok$\times$ its tokens, and then continues to improve. Its advantage therefore comes from search that the single agent did not carry out when offered the allowance.

\begin{table}[H]
\centering
\small
\caption{\method{} against \single{} on development score at equal budget, over the \equalbudgetcompared{} competitions with directly comparable scores.}
\label{tab:equal-budget}
\begin{tabular}{lrrr}
\toprule
Where the \method{} run is cut & \method{} better & Equal & \method{} worse \\
\midrule
At the single agent's scored experiments & 3 & 2 & 8 \\
At the single agent's tokens & 1 & 0 & 12 \\
Not cut (full \method{} run) & 10 & 2 & 1 \\
\bottomrule
\end{tabular}

\end{table}

\paragraph{The conversation as a chain.} Section~\ref{sec:architecture} describes an agent's context as a chain that every model call re-reads. Figure~\ref{fig:chain} illustrates this on \casestudylabel{}: the single agent's chain is one ramp that ends when the agent stops, whereas the line of the reported \method{} member is a sequence of \casestudygenerations{} short chains, one per generation, each starting from its task prompt. Across all runs, the median single agent's chain grows from \chainonestart{} to \chainoneend{} tokens over \chainonecalls{} model calls, re-reading \chainonereread{}M tokens in total. Only \chainonecompressed{} of the \benchmarkcount{} single agents ever reached the harness's compression threshold, and at the stop the median chain held \chainoneendpct\% of the \chainlimit{}-token limit, so the agents did not stop because their context was full. The median \method{} agent's chain grows from \chaingastart{} to \chaingaend{} tokens over \chaingacalls{} calls, re-reading \chaingareread{}M tokens, after which the next generation starts a new one; \chaingacompressed{} of the \chaingaagents{} \method{} agents whose event logs record context size reached the compression threshold. The median run starts \chainsperrun{} such chains and makes \chaingacallsperrun{} model calls in total. AIDE, by contrast, has no chain: each of its \aidesolutions{} solutions is written from a prompt assembled by its scaffold, with \aidecallspersolution{} LLM calls per solution and \aidetokenspercall{} input tokens per call.

\begin{figure}[H]
\centering
\begin{tikzpicture}[x=1cm,y=1cm,font=\scriptsize]
\definecolor{seriesA}{HTML}{2A78D6}\definecolor{seriesB}{HTML}{EB6834}
\draw[black!12] (0,0.000) -- (12.40,0.000); \node[left,black!60] at (0,0.000) {0K};
\draw[black!12] (0,0.850) -- (12.40,0.850); \node[left,black!60] at (0,0.850) {50K};
\draw[black!12] (0,1.700) -- (12.40,1.700); \node[left,black!60] at (0,1.700) {100K};
\draw[black!12] (0,2.550) -- (12.40,2.550); \node[left,black!60] at (0,2.550) {150K};
\draw[black!12] (0,3.400) -- (12.40,3.400); \node[left,black!60] at (0,3.400) {200K};
\draw[black!45,dashed] (0,3.400) -- (12.40,3.400); \node[below right,black!60] at (0.05,3.400) {context limit};
\node[below,black!60] at (0.000,0) {0};
\node[below,black!60] at (2.298,0) {500};
\node[below,black!60] at (4.596,0) {1,000};
\node[below,black!60] at (6.894,0) {1,500};
\node[below,black!60] at (9.192,0) {2,000};
\node[below,black!60] at (11.490,0) {2,500};
\node[black!60] at (6.20,-0.5) {model calls since the start of the run};
\node[rotate=90,black!60] at (-1.0,1.70) {context re-read per call (tokens)};
\draw[seriesA,line width=1.0pt,line join=round] (0.000,0.094) -- (0.026,0.199) -- (0.052,0.325) -- (0.078,0.412) -- (0.104,0.472) -- (0.130,0.497) -- (0.156,0.511) -- (0.182,0.541) -- (0.208,0.583) -- (0.234,0.603) -- (0.260,0.632) -- (0.286,0.652) -- (0.312,0.677) -- (0.338,0.728) -- (0.364,0.765) -- (0.390,0.809) -- (0.415,0.821) -- (0.441,0.856) -- (0.467,0.886) -- (0.493,0.946) -- (0.519,0.980);
\draw[seriesA,line width=1.0pt,line join=round] (0.519,0.091) -- (0.545,0.176) -- (0.570,0.288) -- (0.596,0.358) -- (0.621,0.383) -- (0.647,0.445) -- (0.672,0.466) -- (0.698,0.482) -- (0.723,0.531) -- (0.749,0.562) -- (0.774,0.591) -- (0.800,0.649) -- (0.825,0.680) -- (0.851,0.737) -- (0.876,0.763) -- (0.902,0.825) -- (0.927,0.847) -- (0.953,0.907) -- (0.978,0.935) -- (1.004,0.977) -- (1.030,1.018);
\draw[seriesA,line width=1.0pt,line join=round] (1.030,0.096) -- (1.056,0.304) -- (1.083,0.368) -- (1.110,0.432) -- (1.137,0.446) -- (1.164,0.497) -- (1.191,0.512) -- (1.218,0.533) -- (1.245,0.607) -- (1.271,0.648) -- (1.298,0.676) -- (1.325,0.707) -- (1.352,0.735) -- (1.379,0.782) -- (1.406,0.843) -- (1.433,0.937) -- (1.460,1.011) -- (1.487,1.038) -- (1.513,1.100) -- (1.540,1.117) -- (1.567,1.200);
\draw[seriesA,line width=1.0pt,line join=round] (1.567,0.097) -- (1.597,0.215) -- (1.627,0.286) -- (1.656,0.426) -- (1.686,0.523) -- (1.715,0.622) -- (1.745,0.680) -- (1.775,0.700) -- (1.804,0.823) -- (1.834,0.880) -- (1.864,0.914) -- (1.893,0.936) -- (1.923,1.022) -- (1.953,1.079) -- (1.982,1.190) -- (2.012,1.244) -- (2.042,1.258) -- (2.071,1.300) -- (2.101,1.344) -- (2.130,1.434) -- (2.160,1.504);
\draw[seriesA,line width=1.0pt,line join=round] (2.160,0.103) -- (2.179,0.211) -- (2.199,0.320) -- (2.218,0.373) -- (2.237,0.387) -- (2.257,0.454) -- (2.276,0.506) -- (2.295,0.531) -- (2.315,0.544) -- (2.334,0.554) -- (2.353,0.644) -- (2.372,0.671) -- (2.392,0.678) -- (2.411,0.745) -- (2.430,0.774) -- (2.450,0.799) -- (2.469,0.811) -- (2.488,0.841) -- (2.508,0.858) -- (2.527,0.901) -- (2.546,0.938);
\draw[seriesA,line width=1.0pt,line join=round] (2.546,0.089) -- (2.577,0.243) -- (2.607,0.400) -- (2.637,0.463) -- (2.668,0.503) -- (2.698,0.557) -- (2.728,0.612) -- (2.759,0.664) -- (2.789,0.696) -- (2.819,0.789) -- (2.850,0.819) -- (2.880,0.847) -- (2.910,0.890) -- (2.941,0.913) -- (2.971,1.007) -- (3.001,1.072) -- (3.032,1.222) -- (3.062,1.276) -- (3.092,1.311) -- (3.123,1.385) -- (3.153,1.447);
\draw[seriesA,line width=1.0pt,line join=round] (3.153,0.096) -- (3.187,0.228) -- (3.221,0.341) -- (3.256,0.414) -- (3.290,0.584) -- (3.324,0.723) -- (3.358,0.761) -- (3.393,0.840) -- (3.427,0.944) -- (3.461,1.092) -- (3.495,1.210) -- (3.529,1.227) -- (3.564,1.255) -- (3.598,1.356) -- (3.632,1.455) -- (3.666,1.533) -- (3.701,1.642) -- (3.735,1.704) -- (3.769,1.730) -- (3.803,1.843) -- (3.838,1.888);
\draw[seriesA,line width=1.0pt,line join=round] (3.838,0.097) -- (3.875,0.289) -- (3.913,0.369) -- (3.951,0.457) -- (3.988,0.535) -- (4.026,0.604) -- (4.064,0.698) -- (4.101,0.744) -- (4.139,0.790) -- (4.177,0.872) -- (4.215,0.921) -- (4.252,0.951) -- (4.290,1.013) -- (4.328,1.062) -- (4.365,1.103) -- (4.403,1.155) -- (4.441,1.283) -- (4.478,1.316) -- (4.516,1.383) -- (4.554,1.490) -- (4.591,1.556);
\draw[seriesA,line width=1.0pt,line join=round] (4.591,0.095) -- (4.626,0.313) -- (4.660,0.339) -- (4.695,0.454) -- (4.729,0.515) -- (4.764,0.636) -- (4.798,0.696) -- (4.833,0.723) -- (4.867,0.783) -- (4.902,0.841) -- (4.936,0.914) -- (4.971,0.986) -- (5.005,1.038) -- (5.040,1.074) -- (5.074,1.195) -- (5.108,1.293) -- (5.143,1.349) -- (5.177,1.415) -- (5.212,1.444) -- (5.246,1.492) -- (5.281,1.600);
\draw[seriesA,line width=1.0pt,line join=round] (5.281,0.104) -- (5.331,0.177) -- (5.381,0.301) -- (5.431,0.399) -- (5.481,0.439) -- (5.531,0.503) -- (5.581,0.559) -- (5.631,0.601) -- (5.682,0.651) -- (5.732,0.687) -- (5.782,0.741) -- (5.832,0.784) -- (5.882,0.855) -- (5.932,0.954) -- (5.982,0.985) -- (6.032,1.051) -- (6.082,1.180) -- (6.132,1.249) -- (6.183,1.313) -- (6.233,1.386) -- (6.283,1.504);
\draw[seriesA,line width=1.0pt,line join=round] (6.283,0.101) -- (6.331,0.303) -- (6.379,0.440) -- (6.428,0.561) -- (6.476,0.636) -- (6.524,0.716) -- (6.572,0.866) -- (6.621,0.959) -- (6.669,1.021) -- (6.717,1.095) -- (6.765,1.199) -- (6.814,1.306) -- (6.862,1.484) -- (6.910,1.599) -- (6.958,1.748) -- (7.007,1.849) -- (7.055,1.978) -- (7.103,2.040) -- (7.151,2.121) -- (7.200,2.181) -- (7.248,2.296);
\draw[seriesA,line width=1.0pt,line join=round] (7.248,0.106) -- (7.295,0.210) -- (7.342,0.272) -- (7.389,0.309) -- (7.435,0.383) -- (7.482,0.432) -- (7.529,0.460) -- (7.576,0.509) -- (7.623,0.593) -- (7.670,0.666) -- (7.717,0.726) -- (7.764,0.773) -- (7.810,0.853) -- (7.857,0.899) -- (7.904,0.978) -- (7.951,1.034) -- (7.998,1.142) -- (8.045,1.204) -- (8.092,1.250) -- (8.139,1.333) -- (8.185,1.403);
\draw[seriesA,line width=1.0pt,line join=round] (8.185,0.094) -- (8.225,0.367) -- (8.265,0.427) -- (8.304,0.477) -- (8.344,0.629) -- (8.383,0.727) -- (8.423,0.842) -- (8.462,0.956) -- (8.502,1.028) -- (8.541,1.161) -- (8.581,1.251) -- (8.620,1.369) -- (8.660,1.497) -- (8.699,1.536) -- (8.739,1.603) -- (8.778,1.682) -- (8.818,1.793) -- (8.857,1.594) -- (8.897,1.660) -- (8.936,1.427) -- (8.976,1.497);
\draw[seriesA,line width=1.0pt,line join=round] (8.976,0.098) -- (9.017,0.248) -- (9.058,0.362) -- (9.099,0.439) -- (9.140,0.552) -- (9.181,0.665) -- (9.221,0.768) -- (9.262,0.720) -- (9.303,0.807) -- (9.344,0.868) -- (9.385,0.893) -- (9.426,0.936) -- (9.467,1.005) -- (9.508,1.040) -- (9.549,1.085) -- (9.590,1.131) -- (9.630,1.157) -- (9.671,1.183) -- (9.712,1.225) -- (9.753,1.275) -- (9.794,1.332);
\draw[seriesA,line width=1.0pt,line join=round] (9.794,0.091) -- (9.821,0.249) -- (9.847,0.314) -- (9.874,0.347) -- (9.901,0.376) -- (9.927,0.405) -- (9.954,0.429) -- (9.981,0.447) -- (10.007,0.463) -- (10.034,0.495) -- (10.061,0.608) -- (10.087,0.656) -- (10.114,0.678) -- (10.141,0.706) -- (10.167,0.723) -- (10.194,0.767) -- (10.221,0.797) -- (10.247,0.825) -- (10.274,0.798) -- (10.301,0.810) -- (10.327,0.849);
\draw[seriesA,line width=1.0pt,line join=round] (10.327,0.099) -- (10.378,0.258) -- (10.428,0.462) -- (10.479,0.571) -- (10.529,0.855) -- (10.580,0.988) -- (10.631,1.073) -- (10.681,1.142) -- (10.732,1.233) -- (10.782,1.311) -- (10.833,1.114) -- (10.883,1.215) -- (10.934,1.301) -- (10.984,1.368) -- (11.035,1.394) -- (11.086,1.454) -- (11.136,1.495) -- (11.187,1.580) -- (11.237,1.629) -- (11.288,1.691) -- (11.338,1.810);
\draw[seriesA,line width=1.0pt,line join=round] (11.338,0.098) -- (11.391,0.239) -- (11.444,0.446) -- (11.498,0.550) -- (11.551,0.735) -- (11.604,0.990) -- (11.657,1.038) -- (11.710,1.110) -- (11.763,1.233) -- (11.816,1.338) -- (11.869,1.315) -- (11.922,1.360) -- (11.975,1.455) -- (12.028,1.489) -- (12.081,1.572) -- (12.135,1.623) -- (12.188,1.700) -- (12.241,1.761) -- (12.294,1.824) -- (12.347,1.903) -- (12.400,2.017);
\draw[seriesB,line width=1.5pt,line join=round] (0.000,0.093) -- (0.050,0.288) -- (0.101,0.375) -- (0.151,0.441) -- (0.201,0.489) -- (0.252,0.565) -- (0.302,0.617) -- (0.352,0.674) -- (0.403,0.725) -- (0.453,0.805) -- (0.503,0.877) -- (0.554,0.969) -- (0.604,1.023) -- (0.654,1.102) -- (0.705,1.161) -- (0.755,1.228) -- (0.805,1.282) -- (0.856,1.321) -- (0.906,1.388) -- (0.956,1.475) -- (1.007,1.622);
\fill[seriesB] (1.007,1.622) circle (2pt);
\node[above right,align=left] at (1.007,1.622) {\single{}: one chain, stops here};
\node[above left,align=right] at (12.40,2.346) {\method{}: 17 generations of the reported member, each a fresh chain};
\end{tikzpicture}
\caption{How the conversation grows, on \casestudylabel{}. Every model call receives the retained working context, so a chain grows until the harness compresses it. \single{} keeps a single chain until it stops; \method{} starts every generation from a fresh chain and passes state on through files.}
\label{fig:chain}
\end{figure}

\subsection{The Always-Initial Control in Depth}
\label{app:always-initial}
Section~\ref{sec:always-initial} compares \method{} with a control in which every agent is a fresh \textbf{Initial} run. The control uses the configuration of \method{} (Appendix~\ref{app:hyperparam}) with three changes: the base probability of \textbf{Initial} is 1 and that of every other operator 0, \hedge{} is off, and the stopping rule is disabled so that every run completes \agentgamaxiterations{} iterations. Because an \textbf{Initial} agent depends on no earlier agent, cutting such a run at the iteration at which the stopping rule of Section~\ref{sec:stopping} would have fired gives the run that rule would have produced; this is the same-stopping-rule variant. The control therefore removes inheritance together with everything that depends on it: there is no \textbf{Continue}, \textbf{Ablation}, \textbf{EDA}, or \textbf{Merge} child and no operator allocation. What remains is \agentgapopsize{} parallel slots of repeated de novo runs, each keeping its best solution. Each agent sees only the task and the data: it is compared with the elite of its slot after it finishes, and never sees that solution or its score. One further difference is confined to a single competition: on \initonlysplitcompetition{} the control was scored with a time-based validation split, where \method{} and \single{} used the random split. Without that competition, \method{} ranks higher than the same-stopping-rule variant on \initonlysplitwinsstop{} competitions and lower on \initonlysplitlossesstop{} (mean difference +\initonlysplitdeltastop{} points) and higher than the full-budget variant on \initonlysplitwinsfull{} and lower on \initonlysplitlossesfull{} (+\initonlysplitdeltafull{} points), and the trend test of Section~\ref{sec:always-initial} gives $p=\initonlysplittrendpstop{}$ and $p=\initonlysplittrendpfull{}$. Table~\ref{tab:always-initial} compares both variants with \method{}.

\begin{table}[H]
\centering
\small
\caption{The always-Initial control against \method{}. The mean difference is \method{} minus the control, in points of \emph{Exceeds \% of Human}, with a 95\% bootstrap confidence interval. The last two columns are totals over the \benchmarkcount{} competitions.}
\label{tab:always-initial}
\resizebox{\textwidth}{!}{
\begin{tabular}{lrccrr}
\toprule
Arm & Exceeds (\%) & \method{} higher / lower / equal & Mean difference [95\% CI] & Iterations & Scored exp. \\
\midrule
\method{} & 71.13 & -- & -- & 272 & 3,286 \\
Always-Initial, same stopping rule & 65.76 & 11 / 4 / 1 & +5.37 [$-0.80$, $12.29$] & 106 & 1,271 \\
Always-Initial, full budget & 67.25 & 11 / 5 / 0 & +3.88 [$-2.11$, $9.79$] & 320 & 4,007 \\
\bottomrule
\end{tabular}
}
\end{table}

\paragraph{The ladder.} Section~\ref{sec:always-initial} tests whether leaderboard results rise from \single{} to the control to \method{}. Page's trend test is the several-arm form of the one-sided sign test used for the pairwise comparisons in this paper: the arms are ranked within each competition, and the weighted sum of their rank totals is compared with its exact permutation distribution. With the same-stopping-rule variant, the mean within-competition ranks are \initonlyranksingle{}, \initonlyrankcontrol{}, and \initonlyrankmethod{}, and the one-sided $p$ is \initonlytrendpstop{}; it is \initonlytrendpfull{} with the full budget and \initonlytrendpequal{} with the control cut at the scored experiments of the corresponding \method{} run. Placing AIDE (DS) below the three arms gives $p=\initonlytrendpaide{}$. The ordering does not depend on any single competition: when any one of them is omitted, the largest $p$ is \initonlytrendloostop{} (\initonlytrendloofull{} with the full budget). The test concerns the ordering as a whole. For the last step alone, the one-sided sign test gives $p=\initonlysignpstop{}$ against the same-stopping-rule variant and $p=\initonlysignpfull{}$ against the full budget.

\paragraph{Search.} With the stopping rule disabled, the control still finds occasional improvements late in a run, which is why the full budget scores above the same-stopping-rule variant. A new from-scratch child beats the elite of its slot in \initonlychildwins{} of \initonlychildpairs{} tournaments (\initonlychildwinrate\%): in \initonlyearlywinrate\% of those in iterations 2--5 but in only \initonlylatewinrate\% of those in iterations 16--20. This is the pattern of \textbf{Initial} children inside \method{} (\initialearlywinrate\% and \initiallatewinrate\%, Section~\ref{sec:lineage}), so the within-run comparison was representative of de novo search. Under \method{}'s stopping rule the control ends after between \initonlystopmin{} and \initonlystopmax{} iterations.

\paragraph{Scoring.} The control's solutions were selected and scored on Kaggle in the same way as those of the other arms, without rerunning any model, and ranked on the same fixed leaderboards.

\paragraph{Resources.} The \initonlyagents{} agents of the full budget ran \initonlyexperimentsfull{} scored experiments and recorded \initonlytokensfull{}B tokens, at an estimated cost of \$\initonlycostfull{}. The same-stopping-rule variant comprises \initonlyagentsstop{} agents, \initonlyexperimentsstop{} scored experiments, and \initonlytokensstop{}B tokens (\$\initonlycoststop{}). Token totals are those recorded: \initonlyagentswithtokensfull{} of the \initonlyagents{} control agents and \initonlygaagentswithtokens{} of the \agentgatotalinvocations{} \method{} agents have a token record. A from-scratch agent recorded fewer tokens on average than an agent of \method{}.

\paragraph{Other comparisons.} Cut at the last whole iteration within the scored experiments of the corresponding \method{} run, the control averages \initonlymeanequal\%; \method{} ranks higher on \initonlywinsequal{} competitions and lower on \initonlylossesequal{}, with a mean difference of +\initonlydeltaequal{} points (CI [$\initonlycilowequal{}$, $\initonlycihighequal{}$]). Against AIDE (DS), the same-stopping-rule variant ranks higher on \initonlyaidewinsstop{} competitions and lower on \initonlyaidelossesstop{} (+\initonlyaidedeltastop{} points, CI [\initonlyaidecilowstop{}, \initonlyaidecihighstop{}]), and the full budget on \initonlyaidewinsfull{} and \initonlyaidelossesfull{}. Against \single{}, the mean differences are +\initonlysingledeltastop{} and +\initonlysingledeltafull{} points, and both intervals include zero. Table~\ref{tab:always-initial-competitions} lists every competition.

\begin{table}[H]
\centering
\scriptsize
\caption{The always-Initial control, competition by competition. Exceeds is \emph{Exceeds \% of Human}; the best of the three values in a row is in bold. Iterations are those the run completed, or for the same-stopping-rule variant the iteration at which \method{}'s stopping rule would have ended it.}
\label{tab:always-initial-competitions}
\resizebox{\textwidth}{!}{
\begin{tabular}{lrrrrr}
\toprule
 & \multicolumn{2}{c}{\method{}} & \multicolumn{2}{c}{Always-Initial, same stopping rule} & \multicolumn{1}{c}{Always-Initial, full budget} \\
\cmidrule(lr){2-3}\cmidrule(lr){4-5}\cmidrule(lr){6-6}
Competition & Iterations & Exceeds (\%) & Iterations & Exceeds (\%) & Exceeds (\%) \\
\midrule
Bike Sharing Demand & 20 & \textbf{93.34} & 7 & 93.25 & 88.81 \\
Cat in the Dat & 18 & \textbf{58.69} & 4 & 51.45 & 56.38 \\
House Prices & 11 & \textbf{96.50} & 4 & 81.12 & 79.03 \\
NYC Taxi Fare & 14 & \textbf{89.76} & 4 & 76.36 & 77.44 \\
Playground S3E14 & 20 & \textbf{84.82} & 8 & 80.87 & 80.87 \\
Playground S3E16 & 20 & 76.17 & 6 & \textbf{90.43} & \textbf{90.43} \\
Playground S3E19 & 20 & 43.10 & 8 & \textbf{43.61} & \textbf{43.61} \\
Playground S3E22 & 5 & 80.04 & 6 & \textbf{96.24} & 90.54 \\
Playground S3E24 & 20 & \textbf{99.06} & 6 & 80.47 & 80.47 \\
Playground S3E25 & 4 & 63.99 & 5 & 63.99 & \textbf{84.94} \\
TPS Aug 2022 & 20 & \textbf{92.17} & 9 & 49.23 & 63.53 \\
TPS Feb 2021 & 20 & \textbf{82.29} & 10 & 77.06 & 77.06 \\
TPS Feb 2022 & 20 & \textbf{39.38} & 13 & 38.90 & 35.40 \\
TPS Jan 2022 & 20 & \textbf{44.35} & 6 & 38.51 & 34.42 \\
TPS Jul 2021 & 20 & 25.97 & 4 & 27.36 & \textbf{29.68} \\
TMDB Box Office & 20 & \textbf{68.46} & 6 & 63.37 & 63.44 \\
\midrule
Mean & & 71.13 & & 65.76 & 67.25 \\
\bottomrule
\end{tabular}
}
\end{table}

\section{Lineage Tracing and Verifier Alignment}
\label{app:verifier}
This appendix details the lineage tracing and retrospective scoring used to assess verifier--leaderboard alignment in Section~\ref{sec:verifier}. Among reported runs that completed all \agentgamaxiterations{} iterations, we examine the two with the highest and the two with the lowest \emph{Exceeds \% of Human}. For each run, we trace the population member that produced the reported submission back to its iteration-1 \textbf{Initial} ancestor. For a Merge child, we follow the parent from the child's own population slot. Each subsequent member in the traced lineage replaced its slot's elite, so development performance improves strictly along the lineage.

For each lineage member, we select its saved submission using the same rule as the reported results: order its attempts from best to worst development score and take the first with a submission file. We submit these files unchanged as late submissions to Kaggle and obtain their scores and ranks on the fixed private leaderboard, following the protocol of Table~\ref{tab:completed-results}. No models are retrained or predictions altered, and each lineage's final submission reproduces the reported result exactly. In total, \verfilesscored{} distinct files were scored, covering the traced lineages and the two leading elites of each run after iterations \veriterations{}. These additional checkpoints are included in the rank correlations in Table~\ref{tab:verifier}.

\begin{figure}[H]
\centering
\resizebox{\textwidth}{!}{\input{generated/fig_verifier}}
\caption{\textbf{Development-score improvement does not guarantee leaderboard improvement.} Trajectories of the reported member's lineage in the two highest-ranked and two lowest-ranked completed \agentgamaxiterations{}-iteration runs. Each competition has two aligned traces over successive members of its selected lineage: direction-aware relative development-score improvement above (blue), and change in \emph{Exceeds \% of Human} below (orange), both relative to the initial ancestor. Development panels use task-specific scales; leaderboard panels share a percentage-point scale. Although development scores improve along all four lineages, leaderboard performance improves in two and remains approximately flat or declines in the other two. Endpoint labels give the total change in each measure.}
\label{fig:verifier}
\end{figure}

Table~\ref{tab:verifier} summarizes each line, and Table~\ref{tab:verifier-checkpoints} lists every generation. The training-data column of Table~\ref{tab:verifier} is measured on the competition files: whether the test set lies after the training period, and how much of a random holdout has an exact copy among the training rows.

\begin{table}[H]
\centering
\small
\caption{The reported member's line of descent in each competition. Gen.: number of generations in the line, all of which were scored. Verifier: relative change of the development score from the first to the last generation (positive is better). Exceeds: \emph{Exceeds \% of Human} of the first and the last generation. Steps up / down: generation steps in which \emph{Exceeds} rose or fell. $\rho$: Spearman correlation between development score and private score over every scored file of the run.}
\label{tab:verifier}
\resizebox{\textwidth}{!}{
\begin{tabular}{llp{0.28\textwidth}rrrrr}
\toprule
Competition & Group & Training data against the test set & Gen. & Verifier & Exceeds & Steps up / down & $\rho$ \\
\midrule
Bike Sharing Demand & highest & test set inside the training period & 16 & +8.39\% & 83.2 $\rightarrow$ 93.3 & 10 / 3 & +0.46 \\
Playground S3E24 & highest & no time order, no duplicate rows & 17 & +0.98\% & 70.8 $\rightarrow$ 99.1 & 10 / 3 & +0.94 \\
TPS Feb 2022 & lowest & 51\% of a holdout has an exact copy in training & 9 & +0.26\% & 37.7 $\rightarrow$ 39.4 & 4 / 4 & -0.18 \\
TPS Jul 2021 & lowest & test set after the training period & 14 & +10.03\% & 31.5 $\rightarrow$ 26.0 & 5 / 4 & -0.65 \\
\bottomrule
\end{tabular}
}
\end{table}

\begin{table}[H]
\centering
\scriptsize
\caption{Every generation of the reported member's line of descent: the iteration in which it was produced, its population slot, its operator, its development score, and its Kaggle private score, fixed-leaderboard rank, and \emph{Exceeds \% of Human}. The last row of each competition is the reported result. Consecutive development scores that print the same differ beyond the fifth decimal.}
\label{tab:verifier-checkpoints}
\begin{tabular}{lrrlrrrr}
\toprule
Competition & It.\ & Pop.\ & Operator & Development & Private & Rank & Exceeds \\
\midrule
Bike Sharing Demand & 1 & 1 & Initial & 0.27081 & 0.42880 & 545 & 83.19 \\
Bike Sharing Demand & 2 & 1 & EDA & 0.27024 & 0.42798 & 537 & 83.44 \\
Bike Sharing Demand & 3 & 1 & Continue & 0.26917 & 0.42418 & 502 & 84.52 \\
Bike Sharing Demand & 4 & 1 & EDA & 0.26607 & 0.41720 & 414 & 87.23 \\
Bike Sharing Demand & 6 & 1 & EDA & 0.26283 & 0.41539 & 386 & 88.10 \\
Bike Sharing Demand & 7 & 1 & EDA & 0.26260 & 0.41338 & 372 & 88.53 \\
Bike Sharing Demand & 8 & 1 & Continue & 0.26143 & 0.40502 & 265 & 91.83 \\
Bike Sharing Demand & 9 & 1 & Continue & 0.26123 & 0.40510 & 265 & 91.83 \\
Bike Sharing Demand & 10 & 1 & EDA & 0.25307 & 0.40853 & 324 & 90.01 \\
Bike Sharing Demand & 11 & 1 & Continue & 0.25296 & 0.41034 & 341 & 89.49 \\
Bike Sharing Demand & 12 & 1 & EDA & 0.25173 & 0.41228 & 364 & 88.78 \\
Bike Sharing Demand & 13 & 1 & EDA & 0.25167 & 0.41215 & 363 & 88.81 \\
Bike Sharing Demand & 14 & 1 & EDA & 0.25142 & 0.41205 & 363 & 88.81 \\
Bike Sharing Demand & 15 & 1 & EDA & 0.25093 & 0.40983 & 336 & 89.64 \\
Bike Sharing Demand & 17 & 1 & EDA & 0.25038 & 0.40192 & 248 & 92.35 \\
Bike Sharing Demand & 19 & 1 & EDA & 0.24808 & 0.39848 & 216 & 93.34 \\
\midrule
Playground S3E24 & 1 & 1 & Initial & 0.86952 & 0.87139 & 557 & 70.84 \\
Playground S3E24 & 2 & 1 & EDA & 0.87199 & 0.87401 & 363 & 80.99 \\
Playground S3E24 & 3 & 1 & EDA & 0.87240 & 0.87417 & 356 & 81.36 \\
Playground S3E24 & 4 & 1 & EDA & 0.87241 & 0.87414 & 358 & 81.26 \\
Playground S3E24 & 5 & 1 & Continue & 0.87242 & 0.87415 & 358 & 81.26 \\
Playground S3E24 & 7 & 1 & Continue & 0.87242 & 0.87415 & 358 & 81.26 \\
Playground S3E24 & 8 & 1 & EDA & 0.87643 & 0.87744 & 211 & 88.95 \\
Playground S3E24 & 9 & 1 & Merge & 0.87650 & 0.87782 & 206 & 89.21 \\
Playground S3E24 & 10 & 1 & Continue & 0.87730 & 0.87867 & 123 & 93.56 \\
Playground S3E24 & 11 & 1 & EDA & 0.87734 & 0.87853 & 136 & 92.88 \\
Playground S3E24 & 12 & 1 & EDA & 0.87742 & 0.87863 & 123 & 93.56 \\
Playground S3E24 & 13 & 1 & EDA & 0.87749 & 0.87829 & 161 & 91.57 \\
Playground S3E24 & 14 & 1 & EDA & 0.87751 & 0.87841 & 145 & 92.41 \\
Playground S3E24 & 15 & 1 & EDA & 0.87774 & 0.87854 & 135 & 92.93 \\
Playground S3E24 & 18 & 1 & EDA & 0.87791 & 0.87862 & 123 & 93.56 \\
Playground S3E24 & 19 & 1 & EDA & 0.87816 & 0.87898 & 18 & 99.06 \\
Playground S3E24 & 20 & 1 & Ablation & 0.87816 & 0.87899 & 18 & 99.06 \\
\midrule
TPS Feb 2022 & 1 & 3 & Initial & 0.99317 & 0.94888 & 783 & 37.71 \\
TPS Feb 2022 & 2 & 3 & Merge & 0.99321 & 0.95008 & 766 & 39.06 \\
TPS Feb 2022 & 3 & 3 & EDA & 0.99325 & 0.94996 & 767 & 38.98 \\
TPS Feb 2022 & 13 & 3 & Continue & 0.99345 & 0.94707 & 801 & 36.28 \\
TPS Feb 2022 & 16 & 3 & EDA & 0.99349 & 0.95238 & 742 & 40.97 \\
TPS Feb 2022 & 17 & 3 & EDA & 0.99350 & 0.95044 & 761 & 39.46 \\
TPS Feb 2022 & 18 & 3 & EDA & 0.99359 & 0.94795 & 793 & 36.91 \\
TPS Feb 2022 & 19 & 3 & Merge & 0.99572 & 0.95004 & 766 & 39.06 \\
TPS Feb 2022 & 20 & 3 & EDA & 0.99576 & 0.95023 & 762 & 39.38 \\
\midrule
TPS Jul 2021 & 1 & 1 & Initial & 0.12417 & 0.25170 & 886 & 31.53 \\
TPS Jul 2021 & 2 & 1 & Continue & 0.12331 & 0.25937 & 906 & 29.98 \\
TPS Jul 2021 & 3 & 1 & Ablation & 0.12315 & 0.24845 & 880 & 31.99 \\
TPS Jul 2021 & 4 & 1 & Continue & 0.12305 & 0.25016 & 880 & 31.99 \\
TPS Jul 2021 & 5 & 1 & EDA & 0.12305 & 0.25063 & 880 & 31.99 \\
TPS Jul 2021 & 7 & 1 & EDA & 0.12305 & 0.24968 & 880 & 31.99 \\
TPS Jul 2021 & 9 & 1 & EDA & 0.12289 & 0.24775 & 880 & 31.99 \\
TPS Jul 2021 & 10 & 1 & Merge & 0.11739 & 0.34397 & 1,105 & 14.61 \\
TPS Jul 2021 & 11 & 1 & Continue & 0.11720 & 0.32317 & 1,019 & 21.25 \\
TPS Jul 2021 & 13 & 1 & EDA & 0.11698 & 0.31892 & 1,013 & 21.72 \\
TPS Jul 2021 & 14 & 1 & EDA & 0.11319 & 0.31024 & 996 & 23.03 \\
TPS Jul 2021 & 15 & 1 & EDA & 0.11229 & 0.28274 & 939 & 27.43 \\
TPS Jul 2021 & 19 & 1 & Merge & 0.11174 & 0.28398 & 943 & 27.13 \\
TPS Jul 2021 & 20 & 1 & EDA & 0.11172 & 0.29207 & 958 & 25.97 \\
\bottomrule
\end{tabular}

\end{table}

\section{Agent Trajectory Showcase: Bike Sharing Demand}
\label{app:trajectory}
This appendix follows the reported Bike Sharing Demand solution from its initial ancestor to its final descendant. We select it as a successful illustration with substantial relative reductions in both development and leaderboard error and a measured increase in \emph{Exceeds \% of Human}; it is not a representative or randomly selected run. The gains below accumulate across the lineage, from iteration \bikestartiteration{} to iteration \bikeenditeration{}, rather than arising from a single invocation.

\subsection{Search Progress}
Table~\ref{tab:bike-summary} compares the initial ancestor of the reported solution with its final descendant, not the globally best population member at the first iteration. Both scores are RMSLE, for which lower is better. Leaderboard values come from the saved submissions scored retrospectively in Appendix~\ref{app:verifier}; they were not available to the agent during search.

\begin{table}[H]
\centering\small
\caption{Bike Sharing Demand: measured ancestor-to-descendant gains. Exceeds improvement is in percentage points; error reductions are relative to the initial ancestor.}
\label{tab:bike-summary}
\begin{tabular}{lrrr}
\toprule Metric & Initial ancestor & Final descendant & Improvement \\
\midrule
Development RMSLE & 0.27081 & 0.24808 & 8.39\% reduction \\
Leaderboard RMSLE & 0.42880 & 0.39848 & 7.07\% reduction \\
Exceeds \% of Human & 83.19\% & 93.34\% & +10.15 points \\
Leaderboard rank & 545 & 216 & 329 places \\
\bottomrule\end{tabular}

\end{table}

\begin{figure}[H]
\centering
\begin{tikzpicture}[x=0.55cm,y=0.65cm,font=\scriptsize]
\definecolor{devblue}{HTML}{2563A6}
\definecolor{exorange}{HTML}{C76B20}
\node[anchor=west,font=\small\bfseries] at (0,8.3) {Development RMSLE (lower is better)};
\draw[gray!45] (0,4.8) -- (20,4.8);
\node[below] at (1,4.8) {1};
\node[below] at (5,4.8) {5};
\node[below] at (10,4.8) {10};
\node[below] at (15,4.8) {15};
\node[below] at (19,4.8) {19};
\draw[devblue,thick] (1,7.6500) -- (2,7.5869) -- (3,7.4699) -- (4,7.1287) -- (6,6.7726) -- (7,6.7470) -- (8,6.6180) -- (9,6.5966) -- (10,5.6991) -- (11,5.6868) -- (12,5.5507) -- (13,5.5448) -- (14,5.5169) -- (15,5.4629) -- (17,5.4031) -- (19,5.1500);
\fill[devblue] (1,7.6500) circle (1.5pt);
\fill[devblue] (2,7.5869) circle (1.5pt);
\fill[devblue] (3,7.4699) circle (1.5pt);
\fill[devblue] (4,7.1287) circle (1.5pt);
\fill[devblue] (6,6.7726) circle (1.5pt);
\fill[devblue] (7,6.7470) circle (1.5pt);
\fill[devblue] (8,6.6180) circle (1.5pt);
\fill[devblue] (9,6.5966) circle (1.5pt);
\fill[devblue] (10,5.6991) circle (1.5pt);
\fill[devblue] (11,5.6868) circle (1.5pt);
\fill[devblue] (12,5.5507) circle (1.5pt);
\fill[devblue] (13,5.5448) circle (1.5pt);
\fill[devblue] (14,5.5169) circle (1.5pt);
\fill[devblue] (15,5.4629) circle (1.5pt);
\fill[devblue] (17,5.4031) circle (1.5pt);
\fill[devblue] (19,5.1500) circle (1.5pt);
\node[devblue,anchor=west,above] at (1,7.6500) {0.27081};
\node[devblue,anchor=east,above] at (19,5.1500) {0.24808};
\node[anchor=west,font=\small\bfseries] at (0,3.5) {Exceeds \% of Human (higher is better)};
\draw[gray!45] (0,0) -- (20,0);
\node[below] at (1,0) {1};
\node[below] at (5,0) {5};
\node[below] at (10,0) {10};
\node[below] at (15,0) {15};
\node[below] at (19,0) {19};
\draw[exorange,thick] (1,0.3500) -- (2,0.4116) -- (3,0.6776) -- (4,1.3451) -- (6,1.5594) -- (7,1.6653) -- (8,2.4781) -- (9,2.4781) -- (10,2.0298) -- (11,1.9017) -- (12,1.7268) -- (13,1.7342) -- (14,1.7342) -- (15,1.9387) -- (17,2.6062) -- (19,2.8500);
\fill[exorange] (1,0.3500) circle (1.5pt);
\fill[exorange] (2,0.4116) circle (1.5pt);
\fill[exorange] (3,0.6776) circle (1.5pt);
\fill[exorange] (4,1.3451) circle (1.5pt);
\fill[exorange] (6,1.5594) circle (1.5pt);
\fill[exorange] (7,1.6653) circle (1.5pt);
\fill[exorange] (8,2.4781) circle (1.5pt);
\fill[exorange] (9,2.4781) circle (1.5pt);
\fill[exorange] (10,2.0298) circle (1.5pt);
\fill[exorange] (11,1.9017) circle (1.5pt);
\fill[exorange] (12,1.7268) circle (1.5pt);
\fill[exorange] (13,1.7342) circle (1.5pt);
\fill[exorange] (14,1.7342) circle (1.5pt);
\fill[exorange] (15,1.9387) circle (1.5pt);
\fill[exorange] (17,2.6062) circle (1.5pt);
\fill[exorange] (19,2.8500) circle (1.5pt);
\node[exorange,anchor=west,above] at (1,0.3500) {83.19\%};
\node[exorange,anchor=east,above] at (19,2.8500) {93.34\%};
\node[below] at (10,-0.5) {Search iteration};
\end{tikzpicture}
\caption{The reported Bike Sharing lineage. Every measured generation is retained, including leaderboard setbacks. Development RMSLE decreases monotonically, but leaderboard performance does not. The panels have separate vertical scales.}
\label{fig:bike-showcase}
\end{figure}

The complete checkpoints for this lineage appear in Table~\ref{tab:verifier-checkpoints}.

\subsection{How the Solution Evolved}
The initial solution already used a log-target model with hour-aware features. Subsequent descendants explored ensemble composition, refitting and calibration, and additional model members. Table~\ref{tab:bike-actions} shows selected recorded attempt notes. They describe the agents' motivations for changes, rather than isolating each change's causal contribution.

\begin{table}[H]
\centering\scriptsize
\caption{Selected agent-recorded actions along the Bike Sharing lineage. Notes are quoted, shortened for space.}
\label{tab:bike-actions}
\begin{tabular}{rp{0.23\textwidth}p{0.64\textwidth}}
\toprule Iter. & Attempt & Recorded note (truncated) \\
\midrule
4 & \path{gbm_ensemble_4member} & 3-member ensemble cut rmsle to 0.26656 (seed-noise lever); add a 4th structurally distinct GBM (low subsample/high colsample) to decorrelate errors further \\
8 & \path{ens6b} & model: ens5 + 2nd low-variance regularized specialist G (slow/shallow, low colsample); continues winning direction after deep member F regressed \\
10 & \path{ens_refit_scale17} & Refine iteration-scale lever: raise REFIT\_ITER\_SCALE 1.5-\textgreater{}1.7. Observation: offline faithful replica shows the 'more trees on the larger refit set' scaling plateau is flat 1.5-1.8 with shallow optimum ~1.6-1.7 (1.7 best across ... \\
17 & \path{a3_mlp2} & obs: a2 showed algorithmic-diversity axis delivers with strong non-tree MLP; add a 2nd orthogonal-sized (128,64) MLP — offline 5-seed holdout a2 0.25063-\textgreater{}0.25038 (5/5). 12 trees + 2 MLPs. \\
19 & \path{b3_cat14} & obs: CatBoost new-family axis near-saturated at ~14 members — offline on 8-cat bag, +6 more cats 0.248274-\textgreater{}0.248083 (4/5 seeds), +8 regresses. Take the 14-cat sweet spot. \\
\bottomrule\end{tabular}
\end{table}

\subsection{What a Long Agent Trajectory Looks Like}
\label{app:request-anatomy}
We now zoom into the invocation that produced the final solution. Figures~\ref{fig:request-anatomy} and~\ref{fig:conversation-actions} show what its trajectory looks like in practice, using \longmessages{} messages retained in a saved LLM request from Bike Sharing Demand (iteration \longiteration{}, population \longpopulation{}, EDA operator). The first view shows the message sequence and tool interactions; the second groups it into activities, from inspecting inherited work to developing and testing candidates. Table~\ref{tab:bike-invocation} lists the scored attempts in this invocation; these gains are distinct from the full-lineage gains above.

\begin{table}[H]
\centering\small
\caption{Recorded development RMSLE in the showcased Bike Sharing invocation.}
\label{tab:bike-invocation}
\begin{tabular}{llr}
\toprule Stage & Recorded attempt & RMSLE \\
\midrule
Inherited reproduction & \path{repro} & 0.250384 \\
Candidate 1: LightGBM diversity & \path{b1_lgb} & 0.250148 \\
Candidate 2: eight CatBoost members & \path{b2_cat8} & 0.248274 \\
Candidate 3: fourteen CatBoost members & \path{b3_cat14} & 0.248083 \\
\bottomrule\end{tabular}

\end{table}

\clearpage
\begin{figure}[p]
\centering
\input{generated/fig_request_anatomy}
\caption{\textbf{A recorded agent conversation: Bike Sharing Demand, iteration \longiteration{}, population \longpopulation{} (EDA).} All \longmessages{} messages in the saved request appear in order. Harness messages use the API's \texttt{user} role, not human input. Outlines keep tool calls and matching results together; dots count calls. Close-ups show the task and a two-call exchange, with IDs aliased as A/B and values abbreviated. The figure shows one retained request, not the entire invocation.}
\label{fig:request-anatomy}
\end{figure}
\clearpage

\begin{figure}[p]
\centering
\input{generated/fig_conversation_actions}
\caption{\textbf{The same conversation grouped by agent activity.} The agent reproduces the inherited solution, evaluates three candidates, and records the final winning submission (Table~\ref{tab:bike-invocation}). Bands are author-reviewed annotations of the recorded commands, outputs, and attempts; each describes the dominant activity, including interleaved monitoring and harness handoffs.}
\label{fig:conversation-actions}
\end{figure}
\clearpage

\section{Limitations and Threats to Validity}
\label{app:threats}

This appendix expands on the limitations of the system comparisons, leaderboard metric, and public benchmark.

\subsection{End-to-End System Comparison vs. Controlled Causal Claim}
\label{app:threats-causal}

The comparison with AIDE (DS) in Table~\ref{tab:completed-results} matches the model but neither the harness nor the compute. Both systems use \agentgamodelname{} for every LLM call, which we verified from the AIDE run configurations. AIDE, however, searches over code artifacts with a tree-search outer loop and localized code generation, whereas \method{} launches autonomous agents that manage their own context, follow task instructions, and complete a task over a long-horizon trajectory. We ran AIDE at its default budget of \aidesteps{} steps, at which it used three orders of magnitude fewer tokens than \method{}. Because we did not run AIDE with a larger step budget, Table~\ref{tab:completed-results} does not show how AIDE would scale. The published AIDE (GPT-4T) column is a reference obtained with a different model (GPT-4 Turbo), harness, and date, not a controlled arm.

The \single{} control removes the outer loop while holding the model, agent harness, tools, and workflow layer fixed. It is matched in the attempt allowance offered to the agent, not in the compute consumed. Because the single agent stopped early on most competitions (Section~\ref{sec:single}), the comparison shows that a larger allowance for one agent is not converted into search; it does not show that the two arms are equally expensive. We did not run a fixed-prior control for \hedge{}, a population-size ablation, or a variant that continues a single workspace in place, so the marginal contributions of adaptive allocation, population size, and the fresh-workspace reset are not isolated.

The tournament analysis compares descendants conditioned on inherited parent archives against de novo \textbf{Initial} proposals \emph{within} \method{} runs (i.e., with the model, scaffold, and nominal attempt allowance held constant, with children challenging their respective slot elites). Two caveats apply. First, a child keeps the best of its scored attempts and is compared with a fixed parent score, so absolute win rates include selection on the development splits; both kinds of child are subject to this rule, but its differential effects are not isolated. Second, operator sampling probabilities are adapted by \hedge{} rather than held fixed, which is why Section~\ref{sec:lineage} also reports the comparison restricted to iterations that launched both kinds of child. Table~\ref{tab:completed-results} asks whether the system is competitive on a public benchmark, while the tournament analysis asks whether the inheritance mechanism contributes within the system. We do not claim that Table~\ref{tab:completed-results} alone establishes the causal effect of agent-seed evolution.

The always-Initial control (Section~\ref{sec:always-initial}) is the controlled counterpart of the tournament analysis: it removes inheritance while holding the model, harness, population, tournament, and attempt allowance fixed. Three caveats apply. First, the comparison uses one run per arm and competition, so run-to-run variation is not measured, and the confidence interval of the leaderboard difference between the arms includes zero. Second, the control has no inheriting operator at all, so it also has no recombination and no operator allocation, and it does not separate these from inheritance itself. Third, on \initonlysplitcompetition{} the control was scored with a time-based validation split; Appendix~\ref{app:always-initial} repeats the comparison without that competition.

\subsection{Leaderboard-Rank Metric Limitations}
\label{app:threats-metric}

\emph{Exceeds \% of Human} is defined in Section~\ref{sec:experiments} as $100(1-r/N)$ for official rank $r$ among $N$ teams on the fixed Kaggle leaderboard. Two consequences follow from this definition that are relevant to interpreting the reported numbers.

First, because the metric is rank-based, small absolute differences in the underlying competition score can produce large rank changes when many teams are clustered near the same score, and conversely large absolute score differences can produce small rank changes when teams are sparse. \emph{Exceeds \% of Human} should therefore be read as a measure of external leaderboard position relative to the historical Kaggle field, not as a calibrated effect size in the underlying competition metric. We report private-leaderboard \emph{ranks} alongside \emph{Exceeds \% of Human} for the traced lineages (Section~\ref{sec:verifier}).

Second, ``\emph{Exceeds \% of Human}'' refers to the fraction of the Kaggle leaderboard surpassed by the submission, not to a measurement of true human ability on the underlying task. The leaderboard population includes a mixture of serious competitors, casual entries, and ensembled solutions that are themselves the product of automated tooling, and its composition varies across competitions. Cross-paper comparisons that use different reference populations (e.g., different submission cutoffs or filtered subsets of teams) should therefore be interpreted cautiously.

\subsection{Benchmark Contamination and Public Kaggle Risk}
\label{app:threats-contamination}

Weco-Kaggle Lite is built from public Kaggle competitions, several of which have public leaderboards, public discussion forums, and publicly archived top solutions, in some cases predating the training cutoffs of the LLMs used in this work and in baselines. The base models used by \method{} and by AIDE may have encountered task descriptions, leakage discussions, feature-engineering recipes, or full solution code for some of these competitions during pretraining. This is a property of LLM-based Kaggle-style benchmarks generally, not of \method{} specifically: it affects the AIDE baselines and any other agentic system evaluated on the same suite under the same protocol.

Concretely, this means that the \emph{Exceeds \% of Human} numbers reported in Table~\ref{tab:completed-results} should be read as system performance under the benchmark's stated protocol, given whatever public prior knowledge the underlying models bring, rather than as a measurement of solving each task from scratch. The shared-model comparisons provide a common source of potential prior exposure. Both the model-matched benchmark comparison and the within-run inheritance analysis use the same base model within each comparison. However, systems and operators may exploit prior knowledge differently, and its effect on relative performance and inheritance comparisons is unmeasured.

We do not attempt to fully decontaminate the benchmark or to construct a held-out competition set, both of which are outside the scope of this paper. A benchmark of more recent or harder-to-search competitions, paired with the same evaluation protocol, would be a useful complement to the present results.

\section{Code and Reproducibility}
\label{app:codeorg}

The implementation separates the evolution layer (\texttt{agentga-ga}) from the tabular domain package (\texttt{agentga-kaggle\_tabular\_ml}). The former coordinates agent seeds, selection, and lineage recording; the latter supplies competition manifests, workspace instructions, and the evaluation interface. An agent backend receives a task brief and a prepared workspace. The reported runs use our harness image; a goal-directed agent such as Codex can instead be integrated through an appropriate launcher. Package layouts, execution commands, and output formats are documented in the repository rather than repeated here.

\paragraph{Release plan.}
\label{app:codeorg-release}
Upon acceptance, we will release the evolution layer and domain package at the version used for the reported experiments, including run scripts, configuration templates, operator prompts and workspace instructions, competition manifests and data checksums, lineage and summary records with recorded Kaggle scores, and the analysis and plot-generation scripts. The release will also include adaptations for other agent harnesses. Provider credentials, model weights, pre-built images, and the full archive of intermediate experimental artifacts will not be redistributed; users supply their chosen agent backend and competition data.

The recorded artifacts support inspection of the reported experiments, while rerunning the search produces a new stochastic realization rather than an exact replay: neither the evolution layer's random generator nor LLM sampling is seeded. Using another backend likewise changes the execution configuration rather than reproducing the reported harness.

\end{document}